\pdfoutput=1
\documentclass[a4paper,10pt]{article}

\usepackage[utf8x]{inputenc}
\usepackage[english]{babel}
\usepackage{amsmath,amssymb,nicefrac}
\usepackage[mathscr]{eucal}
\usepackage{mathrsfs}
\usepackage{subfigure}
\usepackage{tikz,pgfplots}
\usepgfplotslibrary{patchplots,colormaps}
\usepackage{tabu}
\usepackage{graphicx}
\usepackage{ifpdf}
\ifpdf
  \DeclareGraphicsExtensions{.pdf,.png,.jpg}
\else
  \DeclareGraphicsExtensions{.eps}
\fi
\usepackage[colorlinks=true]{hyperref}
\usepackage{authblk}
\usepackage{color}
\hypersetup{urlcolor=blue, citecolor=red}

\usepackage{fancyhdr}
\def\Ldic{\mathscr{L}_{\mathrm{dic}}}
\def\Lrec{\mathscr{L}_{\mathrm{rec}}}
\def\Pdic{\Phi_{\mathrm{dic}}}
\def\Prep{\Phi_{\mathrm{rep}}}
\def\Psp{\Phi_{\textsc{sp}}}
\def\Pip{\Phi_{\textsc{ip}}}
\def\half{{\textstyle\frac{1}{2}}}
\def\ex{^{\mathrm{exact}}}

\title{Tomographic Image Reconstruction using Training Images}

\author{Sara Soltani \thanks{ssol@dtu.dk}}
\author{Martin S. Andersen \thanks{mskan@dtu.dk}}
\author{Per Christian Hansen \thanks{pcha@dtu.dk}}
\affil{Department of Applied Mathematics and Computer Science,
Technical University of Denmark, DK-2800 Kgs. Lyngby, Denmark.}

\begin{document}

\maketitle
\let\thefootnote\relax\footnote{This work is part of the project HD-Tomo funded by Advanced Grant
No.\ 291405 from the European Research Council}
\let\thefootnote\relax 
\let\thefootnote\relax
  
  \begin{abstract}
We describe and examine an algorithm for tomographic image reconstruction
where prior knowledge about the solution is available in the form of training images.
We first construct a nonnegative dictionary based on prototype elements from
the training images; this problem is formulated as a regularized non-negative
matrix factorization.
Incorporating the dictionary as a prior in a convex reconstruction problem,
we then find an approximate solution with a sparse representation in the dictionary.
The dictionary is applied to non-overlapping patches of the image,
which reduces the computational complexity compared to other algorithms.
Computational experiments clarify the choice and interplay of the model parameters
and the regularization parameters, and we show that in few-projection low-dose
settings our algorithm is competitive with total variation regularization and
tends to include more texture and more correct edges.
\end{abstract}

\section{Introduction}
\label{sec:Intro}
Computed tomography (CT) is a technique to compute an image of the interior
of an object from measurements obtained by sending X-rays through the object
and recording the damping of each ray.
CT is used routinely in medical imaging, materials science, nondestructive testing
and many other applications.

CT is an inverse problem \cite{MuSi12} and it is challenging to obtain sharp
and reliable reconstructions in low-dose measurements where we face
underdetermined systems of equations, because
we must limit the accumulated amount of X-rays for health reasons
or because measurement time is limited.
In these circumstances the
classic methods of CT, such as filtered back projection \cite{Kuch14} and
algebraic reconstruction techniques \cite{HansenHansen},
are often incapable of producing satisfactory reconstructions because they fail
to incorporate adequate prior information \cite{Bian}.
To overcome these difficulties it is necessary to incorporate a prior about
the solution that can compensate for the lack of data.

A popular prior is that the image is piecewise constant, leading to
total variation (TV) regularization schemes \cite{LaRoque}, \cite{Velikina}.
These methods can be very powerful when the solution is
approximately composed of homogeneous regions separated by sharp boundaries.

An completely different approach is to use prior information in the form of
``training images'' that characterize the geometrical or visual features
of interest, e.g., from high-accuracy reconstructions (the typical case) or
from pictures of specimen slices.
The goal of this work is to elaborate on this approach.
In particular we consider the two-stage framework
where the most important features of the training data are first extracted and then
integrated in the reconstruction problem.

A natural way to extract and represent prior information from training images is to form a
\textit{dictionary} that sparsely encodes the information \cite{Olshausen}.
Learning the dictionary from given training data appears to be very suited for
incorporating priors that are otherwise difficult to formulate in a closed form,
such as image texture.
Dictionary learning\,---\,combined with sparse representation
\cite{Bruckstein,Eladbook,Tropp}\,---\,is now used in many image processing
areas including denoising \cite{Chen}, \cite{Li}, inpainting \cite{Mairal0},
and deblurring \cite{Liu}.
Elad and Ahron \cite{Elad} address the image denoising problem using a
process that combines dictionary learning and reconstruction.
They use a dictionary trained from a noise-free image
using the K-SVD algorithm \cite{Aharon} combined with an adaptive dictionary
trained on patches of the noisy image.

The use of dictionary learning in tomographic imaging has also emerged recently,
e.g., in X-ray CT \cite{Etter,Xu,Zhao},
magnetic resonance imaging \cite{Huang,Ravishankar},
electron tomography \cite{LiuB}, positron emission tomography \cite{ChenS},
and phase-contrast tomography \cite{Mirone}.
Two different approaches have emerged\,---\,either one constructs the dictionary
from the given data in a joint learning-reconstruction algorithm
\cite{ChenS,Huang,LiuB,Ravishankar}, or one constructs the dictionary
from training images in a separate step before the reconstruction
\cite{Etter,Mirone,Xu,Zhao}.
Most of these works use K-SVD to learn the dictionary
(except \cite{Etter} that uses an ``online dictionary learning method'' \cite{Mairal1}),
and all the methods regularize the reconstruction by means of a
penalty that is applied to a patch around every pixel in the image. In
other words, all patches in the reconstruction should be close to the
subspace spanned by the dictionary images.
While all these methods
perform better than classical reconstruction methods, they show no significant
improvement over the TV-regularized approach.

In simultaneous learning and reconstruction, where the dictionary is
learned from the given data, the prior is purely data-driven.
Hence, one can argue that it violates a fundamental principle of inverse problems
where a data-independent prior
is incorporated to eliminate unreasonable models that fit the data.
For this reason we prefer to separate the two steps (which requires that
reliable training images are available).
We describe and examine a two-stage framework where we first construct a dictionary
that contains prototype elements from these images, and
then we use the dictionary as a prior to regularize the reconstruction problem
via computing a solution that has a sparse representation in the dictionary.

Our two-stage algorithm is inspired by the work in \cite{Etter} and, to some extent, \cite{Xu}.
The algorithm in \cite{Etter} is tested on a simple tomography setup with no noise in the data and in \cite{Xu} the dictionary
is trained from an image reconstructed by a high-dose X-ray exposure and then used to reconstruct the same image with fewer X-ray projections.
We utilize the dictionary in a different way using blocks of the image
(to be discussed later) which reduces the number of unknowns.
We seek to use more realistic simulations with noisy data, we avoid committing
``inverse crime,'' and
we perform a careful study of the sensitivity of the reconstruction to
the different parameters in the reconstruction model and in the algorithm.
Finally we compare our algorithm with both classical methods and with TV.
We are not aware of comprehensive studies of the influence of the learned
dictionary structure and dictionary parameters in CT.

Our paper is organized as follows.
In section \ref{sec:framework} we briefly discuss dictionary learning methods and
present a framework for solving the image reconstruction problem using dictionaries,
and in Section \ref{sec:Methodology} we describe the implementation
details of algorithm.
Section \ref{sec:Results} presents careful numerical experiments where we
study the influence of the algorithm and design parameters.
Section \ref{sec:Final} summarizes our work.
We use the following notation, where $A$ is an arbitrary matrix:
  \[
    \textstyle
    \| A \|_{\mathrm{F}} = \left( \sum_{ij} A_{ij}^2 \right)^{1/2}, \quad
    \| A \|_{\mathrm{sum}} = \sum_{ij} | A_{ij} | , \quad
    \| A \|_{\max} = \max_{ij} | A_{ij}| .
  \]

\section{The Reconstruction Framework}
\label{sec:framework}
X-ray CT is based on the principle that if we send X-rays through an object and
measure the damping of each ray then, with infinitely many rays, we can perfectly
reconstruct the object.
The attenuation of an X-ray is proportional to the object's attenuation
coefficient, as described by Lambert-Beer's law \cite[\S 2.3.1]{Buzug}.
We divide the domain onto pixels whose unknown nonnegative attenuation coefficients are
organized in the vector $x\in\mathbb{R}^n$.
Similarly we organize the measured damping of the rays into the vector $b\in\mathbb{R}^m$.
Then we obtain a linear system of equations $A\, x = b$ with a large sparse
\emph{system matrix}
governed solely by the geometry of the measurements:\ element $a_{ij}$ is
the length of the $i$th ray passing through pixel~$j$, and the matrix is sparse because
each ray only hits a small number of pixels \cite{MuSi12}.

The matrix $A$ is ill-conditioned, and often rank deficient, due to the
ill-posedness of the underlying inverse problem and therefore the solution
is very sensitive to noise in the data~$b$.
For this reason, a simple least squares approach with nonnegativity constraints
fails to produce a meaningful solution, and we must use regularization to incorporate
prior information about the solution \cite{Hansen}.

This work is concerned with underdetermined problems where $m<n$, and
the need for regularization is even more pronounced.
Classical reconstruction methods such as filtered back projection and
algebraic iterative methods are not suited for these problems because
they fail to incorporate enough prior information.
TV regularization, which is suited for edge-preserving reconstructions,
takes the form
  \begin{equation}
  \label{eq:TV}
    \min_{x} \quad
   \half\, \|A\, x-b\|_2^2 + \lambda_{\mathrm{TV}}
   \sum_{1\leq i \leq n} \left\| D^{\mathrm{fd}}_i x \right\|_2
   \qquad \hbox{subject to} \qquad x \geq 0 ,
  \end{equation}
where we have included a nonnegativity constraint; $D^{\mathrm{fd}}_i x$ is
a finite-difference approximation of the gradient at pixel~$i$, and
$\lambda_{\mathrm{TV}} > 0$ is a regularization parameter.
TV methods produce images whose pixel values are clustered into regions with somewhat
constant intensity \cite{Strong}, with the result that textural images tend to be
over-smoothed (except for the sharp edges).
Another drawback is that the TV problem \eqref{eq:TV} tends to produce reconstructions whose
intensities are incorrect \cite{Strong}.

Our goal is to incorporate prior information\,---\,e.g., about texture\,---\,from
a set of training images.
We focus on formulating and finding a learned dictionary $W$
from the training images and solving the tomography problem such that $x=W\alpha$ is
a sparse linear combination of the dictionary elements (the columns of~$W$).
We build on ideas from sparse approximation \cite{Bruckstein,Eladbook,Tropp}
which seeks an approximate representation of a signal/image using a linear combination
of a few known basis elements.

As mentioned in the Introduction, some works use a joint formulation that
combines the dictionary learning problem and the reconstruction problem into one
optimization problem, i.e., the dictionary is learned from the given noisy data.
This corresponds to a ``bootstrap'' situation where one creates the prior as part
of the solution process.
Our work is different:\ we use a prior that is already available in the form of
a set of training images, and we use this prior to regularize the reconstruction problem.
To do this, we use a two-stage algorithm where we first compute the
dictionary from the given training images, and then we use the dictionary
to compute the reconstruction.

The dictionary $W$ should comprise all the important features of the desired solution.
A learned dictionary\,---\,while computationally more expensive than a
fixed dictionary\,---\,has the advantage that it
is tailored to the characteristics of the desired
solution and optimized for the training images.
Dictionary learning is a way to summarize and represent
a large number of training images into fewer elements and, at the same time,
compensate for noise or other errors in these images.
The learned dictionary should be robust to irrelevant features,
and the number of training images should be large enough to ensure that all
image features are represented; hence dictionaries are typically overcomplete.

Using training images of the same size as the image to be reconstructed
would require a huge number of training images and lead to an enormous dictionary.
All algorithms therefore use a \emph{patch dictionary} $D$ learned from
patches of the training images.
But contrary to previous algorithms that apply a
  dictionary-based regularization based on
overlapping patches around every pixel in the image,
we divide the reconstruction into nonoverlapping blocks of the same size as the patches and
use the dictionary $D$ within each block (ensuring that we limit blocking effects);
conceptually this corresponds to building a \textit{global dictionary} $W$ from $D$.

Let the patches be of size $P \times Q$, and
let the matrix $Y \in \mathbb{R}^{p\times t}$
consist of $t$ training image patches arranged as vectors of length $p=P Q$.
Then the dictionary learning problem can be viewed as the problem of
approximating the training  matrix as a product of two matrices,
$Y \approx DH$, where $D \in \mathbb{R}^{p\times s}$ is
the dictionary of $s$ dictionary image patches (the columns of $D$),
and $H \in \mathbb{R}^{s\times t}$ contains information about the
approximation of each of the training image patches.
Such a decomposition is clearly not unique, so we
must incorporate further requirements to ``shape'' the patch dictionary $D$ and the
representation matrix $H$.

Imposing norm and/or non-negativity constraints on the elements of $D$ and $H$
or imposing sparsity constraint on matrix $H$ are widely used
in unsupervised learning. We take the same approach, and thus our generic dictionary learning problem
takes the form:
  \begin{equation}
  \label{eq:diclergen}
    \min_{D,H} \quad
    \Ldic(Y,DH) + 
    \Pdic(D) + 
    \Prep(H) .
  \end{equation}
Here, the misfit of the factorization approximation is measured by the
loss function $\Ldic$, while the priors on the patch dictionary $D$
and the representation matrix $H$ are taken into account by the regularization functions
$\Pdic$ and $\Prep$.

The dictionary learning problem \eqref{eq:diclergen}
is a non-convex optimization problem.
If we choose the functions $\Ldic$, $\Pdic$ and $\Prep$ to be convex,
then the optimization problem in \eqref{eq:diclergen} is
not jointly convex in $(D,H)$, but it is convex with respect to each variable $D$ or $H$
when the other is fixed.
A natural way to find a local minimum is therefore to use an alternating approach,
first minimizing over $H$ with $D$ fixed, and then minimizing over $D$ with $H$ fixed.

Various dictionary learning methods proposed in the literature
share the same overall structure but they consider different priors
when formulating the dictionary learning problem.
Examples of such methods include, but are not limited to,
non-negative matrix factorization \cite{LeeSeung},
the method of optimal directions \cite{Engan},
K-means clustering \cite{kmeans} and its generalization
K-SVD \cite{Elad}, and the online dictionary learning method \cite{Mairal1}.
The methods in \cite{Kreutz} and \cite{Lewicki} are designed for training data
corrupted by additive noise; but this it is not relevant for our work.

Having computed the patch dictionary $D$ and formed the corresponding global
dictionary $W$, the second step is to solve the reconstruction problem.
Using ideas from sparse approximation, we compute a solution $x = W\alpha$ where $\alpha$ solves the problem
  \begin{equation}
  \label{eq:recgen}
    \min_{\alpha} \quad
    \Lrec(AW\alpha,b) + 
    \Psp(\alpha) + 
    \Pip(W\alpha) ,
  \end{equation}
in which the data fidelity is measured by the loss function $\Lrec$
and regularization is imposed via penalty functions.
Specifically, the function $\Psp$ enforces the Sparsity Prior on $\alpha$,
often formulated in terms of a sparsity inducing norm,
while the function $\Pip$ enforces the Image Prior.
If we choose the three functions $\Lrec$, $\Psp$ and $\Pip$ to be convex, then the
problem formulation \eqref{eq:recgen} can be solved by means of convex optimization methods.
Given a solution $\alpha^\star$ to (\ref{eq:recgen})
we compute the solution as $x^\star = W{\alpha}^\star$.
In Section \ref{sec:Results} we illustrate with numerical examples that the
sparsity penalties in \eqref{eq:diclergen} and \eqref{eq:recgen} tend to have a
regularizing effect on the reconstruction.

\section{Details of Formulation and Implementation}
\label{sec:Methodology}
Recall that the proposed framework for dictionary-based tomographic
reconstruction consists of two conceptual steps: (i) computing a
dictionary (using techniques from machine learning), and (ii) computing a
reconstruction composed of images from the dictionary.
In this section we describe one of many ways to efficiently implement such a scheme.
We pose the dictionary-learning problem as a so-called non-negative sparse coding
problem, and we use least squares optimization with non-negative variables and
1-norm regularization to compute a reconstruction.

\subsection{The Dictionary Learning Problem}
\label{sec:OurDic}

Dictionary learning problems of the form \eqref{eq:diclergen} are
generally non-convex optimization problems due to the bilinear term
$DH$ where both $D$ and $H$ are unknown. Applying a convergent
iterative optimization method therefore does not guarantee that we
find a global minimum (only a local stationary point). To obtain a
good dictionary, we must be careful when choosing the loss function $\Ldic$
and the penalties $\Pdic$ and $\Prep$ on $D$ and $H$,
and we must also pay attention to implementation issues such as the
starting point; see the Appendix for details.

A non-negative matrix factorization (NMF) has the ability to extract
meaningful factors \cite{LeeSeung}, and with non-negative elements in
$D$ its columns represent a basis of images. Similarly, having
non-negative elements in $H$ corresponds to each training image being
represented as a conic combination of dictionary images, and the
representation itself is therefore non-negative. NMF often
works well in combination with sparsity heuristics \cite{Hoyer2004}
which in our application translates to training image patches being
represented as a conic combination of a small number of dictionary elements
(basis images).

The dictionary learning problem that we will use henceforth takes the form
of non-negative sparse coding \cite{Hoyer2004}
of a non-negative data matrix $Y$:
  \begin{equation}
  \label{e-dictionary-learn}
    \min_{D,H} \quad
    \half\, \|Y-DH\|_{\mathrm{F}}^2 + \lambda\, \| H \|_{\mathrm{sum}}
    \qquad \mathrm{s.t.} \qquad D \in \mathcal{D}, \ H \in
    \mathbb{R}_+^{s \times t} ,
  \end{equation}
where the set $\mathcal D$ is compact and convex and $\lambda \geq 0$
is a regularization parameter that controls the sparsity-inducing
penalty $\| H \|_{\mathrm{sum}}$.
This problem is an instance of the more general formulation
(\ref{eq:diclergen}) if we define
  \[
    \Ldic(Y,DH) = \half \| Y - DH \|_{\mathrm{F}}^2
  \]
and
  \[
    \Pdic(D) = I_{\mathcal D}(D), \qquad
    \Prep(H) = I_{\mathbb{R}_+^{s \times t}}(H) + \lambda \| H \|_{\mathrm {sum}} \ ,
\]
where $I_{\mathcal Z}$ denotes the indicator function of a set
$\mathcal{Z}$.
Note that the loss function $\Ldic$ is invariant under a
scaling $D \mapsto \zeta D$ and $H \mapsto \zeta^{-1} H$ for $\zeta > 0$, and
letting $\zeta \rightarrow \infty$ implies that $\Prep(\zeta^{-1}H)
\rightarrow 0$ and $\|\zeta D\| \rightarrow \infty$ if $D$ is nonzero.
This means that $\mathcal D$ must be compact to ensure that the
problem has well-defined minima. Here we will consider two different
definitions of the set $\mathcal{D}$, namely
\begin{align*}
    \mathcal{D}_\infty  \equiv  \{ D \in \mathbb{R}_+^{p \times s}
    \,|\, \|d_j\|_{\infty} \leq 1\}  \quad  \text{and} \quad
    \mathcal{D}_2 \equiv  \{ D \in \mathbb{R}_+^{p \times s} \,|\,
    \|d_j\|_{2} \leq \sqrt{p}\} .
\end{align*}
The set $\mathcal{D}_\infty$ corresponds to box constraints, and
$\mathcal{D}_2$ is a spherical sector of the 2-norm ball with radius
$\sqrt{p}$. As we will see in the next section, the use of
$\mathcal{D}_\infty$ as a prior gives rise to binary-looking images
(corresponding to the vertices of $\mathcal{D}_\infty$) whereas
$\mathcal{D}_2$ gives rise to more ``natural looking'' images.

We emphasize an important difference between the classical K-SVD method and our method.
While K-SVD requires that we explicitly set the sparsity level, in our approach
we affect sparsity implicitly through 
$1$-norm regularization
and via the regularization parameter~$\lambda$.

We use the Alternating Direction Method of Multipliers (ADMM) \cite{Boyd} to
compute an approximate local minimizer of (\ref{e-dictionary-learn}).
Learning the dictionary with the ADMM method has the advantages that the updates
are cheap to compute, making the method suited for large-scale problems.
The implementation details are described in the Appendix.

\subsection{The Reconstruction Problem}

Recall that we formulate the CT problem
as $Ax\approx b$, where $b$ contains the noisy data and $A$ is the system matrix.
The vector $x$ represents an
$M \times N$ image of absorption coefficients, and
these coefficients must be nonnegative to have physical meaning.
Hence we must impose a nonnegativity constraint on the solution.

Let us turn to the reconstruction problem based on the patch
dictionary $D$ and the formulation \eqref{eq:recgen}.
For ease of our presentation we assume
that the image size $M\times N$ is a multiple of the patch size $P\times Q$,
and we partition the image into
an $(M/P) \times (N/Q)$ array of non-overlapping blocks or patches
represented by the vectors $x_j \in \mathbb{R}^p$ for $j=1,\ldots,q=(M/P)(N/Q)$.
The advantage of using non-overlapping blocks, compared to
overlapping blocks, is that we avoid over-smoothing the image textures
when averaging over the overlapping regions, and it requires less computing time.

Each block of $x$ is expressed as a conic combination of dictionary images,
and hence the dictionary prior is expressed as
  \begin{equation}
  \label{e-lasso-recon-sol}
    \Pi \, x = W \alpha , \quad W = (I\otimes D) , \qquad \alpha \geq 0,
  \end{equation}
where $\Pi$ is a permutation matrix that re-orders the vector
$x$ such that we reconstruct the image block by block,
$W$ is the global dictionary, and
  \[
    \alpha = \begin{pmatrix} \alpha_1 \\ \vdots \\ \alpha_q \end{pmatrix}
    \in \underbrace{\mathbb{R}^{s} \times \cdots \times \mathbb{R}^{s}}_{\hbox{$q$ times}}
  \]
is a vector of coefficients for each of a total of $q$ blocks. With this non-overlapping formulation,
it is straightforward to determine the numebr of unknowns in the problem \eqref{eq:mainRec}.
The length of $\alpha$ is $sq=n\, s/p$ which is equal to the product of
the over-representation factor $s/p$ and the number of pixels $n$ in the image.

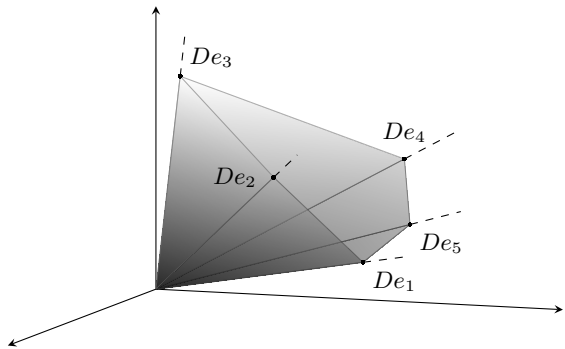
\begin{figure}
  \centering
 \begin{tikzpicture}
\begin{axis}[color=black,
  view={110}{12},
  width=3.5in,height=2.5in,
  axis lines=center,
  xmajorticks=false,ymajorticks=false,zmajorticks=false,
  xmin=0,xmax=2.5,ymin=0,ymax=2.5,zmin=0,zmax=3.2]

\addplot3[
	table/row sep=\\,
	patch,
	patch type=triangle,
	shader=faceted interp,
  	colormap/gray,
	draw opacity=0.6,
	fill opacity=0.6,		
	patch table={
		0 1 2\\
		0 2 3\\
		0 3 4\\
		0 4 5\\
		0 5 1\\
	}]
	table {
		x y z\\
		0 0 0\\
		2 2 1\\
		1.8741 1.4056 1.8741\\
		1.0954 0.5477 2.7386\\
		1.1717 1.9528 1.9528\\
		1.6100 2.1466 1.3416\\
	};

\addplot3 [mark=*,mark size=0.5] (2.0000,2.0000,1.0000);
\addplot3 [no marks,dashed] coordinates {(2.0000,2.0000,1.0000)(2.4000,2.4000,1.2000)};
\addplot3 [mark=*,mark size=0.5] (1.8741,1.4056,1.8741);
\addplot3 [no marks,dashed] coordinates {(1.8741,1.4056,1.8741)(2.2489,1.6867,2.2489)};
\addplot3 [mark=*,mark size=0.5] (1.0954,0.5477,2.7386);
\addplot3 [no marks,dashed] coordinates {(1.0954,0.5477,2.7386)(1.3145,0.6572,3.2863)};
\addplot3 [mark=*,mark size=0.5] (1.1717,1.9528,1.9528);
\addplot3 [no marks,dashed] coordinates {(1.1717,1.9528,1.9528)(1.4060,2.3434,2.3434)};
\addplot3 [mark=*,mark size=0.5] (1.6100,2.1466,1.3416);
\addplot3 [no marks,dashed] coordinates {(1.6100,2.1466,1.3416)(1.9320,2.5759,1.6099)};

\node [below right] at (axis cs:2.0000,2.0000,1.0000) {\small $De_1$};
\node [left=0.1cm] at (axis cs:1.8741,1.4056,1.8741) {\small $De_2$};
\node [above right] at (axis cs:1.0954,0.5477,2.7386) {\small $De_3$};
\node [above=0.1cm] at (axis cs:1.1717,1.9528,1.9528) {\small $De_4$};
\node [below right] at (axis cs:1.6100,2.1466,1.3416) {\small $De_5$};

\end{axis}
\end{tikzpicture}
  \caption{Polyhedral cone in $\mathbb{R}_+^p$ spanned by five nonnegative dictionary elements,
    where $e_i$ denotes the $i$th canonical unit vector in $\mathbb{R}^s$.}
  \label{fig:polyhedral_cone}
\end{figure}

In pursuit of a nonnegative image $x$, we impose the constraint that the vector
$\alpha$ should be nonnegative. This implies that each block $x_j$ of
$x$ lies inside a polyhedral cone
  \begin{align}
  \label{eq:cone}
    \mathcal C = \{ D z \,|\, z \in \mathbb{R}_+^s \} \subseteq \mathbb{R}_+^p
  \end{align}
as illustrated in Figure~\ref{fig:polyhedral_cone}.
Clearly, if the dictionary contains the standard basis of
$\mathbb{R}^p$ then $\mathcal{C}$ is equivalent to the entire nonnegative
orthant in $\mathbb{R}^p$. However, if the cone $\mathcal C$ is a
proper subset of $\mathbb{R}_+^p$, then not all nonnegative images
have an exact representation in $\mathcal C$, and hence the
constraints $x_j \in\mathcal{C}$ may have a regularizing effect even
without a sparsity prior on $\alpha$. This can also be motivated by
the fact that the faces of the cone $\mathcal C$ consist of images $x_j$
that can be represented as a conic combination of at most $p-1$
dictionary images.

Adding a sparsity prior on $\alpha$, in addition to nonnegativity
constraints, corresponds to the assumption that $x_j$ can be expressed as
a conic combination of a small number of dictionary images and hence
provides additional regularization. We include a 1-norm regularizer in
our reconstruction problem as the standard approximate sparsity prior on $\alpha$.

Reconstruction based on non-overlapping blocks
often gives rise to block artifacts in the reconstruction
because the objective in the reconstruction
problem does not penalize jumps across the boundaries of neighboring blocks.
To mitigate this type of artifact, we add a penalty term
that discourages such jumps.
We choose a penalty of the form
  \begin{equation}
  \label{eq:psi}
    \psi(z) =  \half \, \| L \, z \|_2^2 / \ell , \quad
    \ell = M(M/P-1)+N(N/Q-1)
  \end{equation}
where $L$ is a matrix such that $L\, z$ is a vector with finite-difference
approximations of the directional derivatives across the block boundaries.
The factor $\ell$ is the total number of pixels along the boundaries of the blocks
in the image.

The constrained least squares reconstruction problem is then given by
  \begin{align}
  \label{eq:mainRec}
    \begin{array}{ll}
    \mbox{minimize}_{\alpha}
      & \half \frac{1}{m} \| A \Pi^T (I\otimes D) \alpha - b \|_2^2
      + \mu\, \frac{1}{q} \| \alpha \|_1 + \delta^2 \, \psi(\Pi^T(I\otimes D) \alpha)\\
    \mbox{subject to}
      & \alpha \geq 0
    \end{array}
 \end{align}
with regularization parameters $\mu, \delta > 0$.
We normalize the problem formulation by
i) division of the squared residual norm by the number of measurement $m$,
ii) division of the 1-norm of $\alpha$ by the number of blocks $q$,
and iii) division by $\ell$ in the function $\psi$.
Problem \eqref{eq:mainRec} is convex and it is an instance of a
sparse approximation problem similar to formulations studied in \cite{Elad}.

\section{Numerical Experiments}
\label{sec:Results}
In this section we use numerical examples to demonstrate and quantify the
behavior of our two-stage algorithm and evaluate the computed reconstructions.
In particular we explore the influence of
the dictionary structure and its parameters (number of elements, patch sizes)
on the reconstruction, in order to illustrate the role of the learned dictionary.

The underlying idea is to compute a regularized least squares fit in which the
solution is expressed in terms of the dictionary, and hence it lies in the cone
$\mathcal{C}$ \eqref{eq:cone} defined by the dictionary elements.
Hence there are two types of errors in the reconstruction process.
Typically, the exact image does not lie in the cone ${\mathcal C}$,
leading to an \emph{approximation error}. 
Moreover, we encounter a \emph{regularization error} due to the combination
of the error present in the data and the regularization scheme.

In the learning stage we use a set of images which are similar to
the ones we wish to reconstruct.
The ground-truth or exact image $x\ex$ is not contained in the training set,
so that we avoid committing an inverse crime.
All images are gray-level and scaled in the interval $[0,1]$.

All experiments were run in MATLAB (R2011b) on a 64-bit Linux system.
The reconstruction problems are solved using the software package
TFOCS (Templates for First-Order Conic Solvers) \cite{Becker}.
We compare with TV reconstructions computed by means of
the MATLAB software \textsc{TVReg} \cite{Jensen}, with filtered back
projection solutions computed by means of MATLAB's ``iradon'' function,
and solutions computed by means of the algebraic reconstruction technique
(ART, also known as Kaczmarz's method) with nonnegativety constraints implemented
in the MATLAB package \textsc{AIR Tools} \cite{HansenHansen}.
(We did not compare with Krylov subspace methods because they are inferior to ART
for images with sharp edges.)

\subsection{The Test Image and the Tomographic Test Problem}

\begin{figure}[ht!]
\centering
\subfigure{ \includegraphics[width=0.3\linewidth]{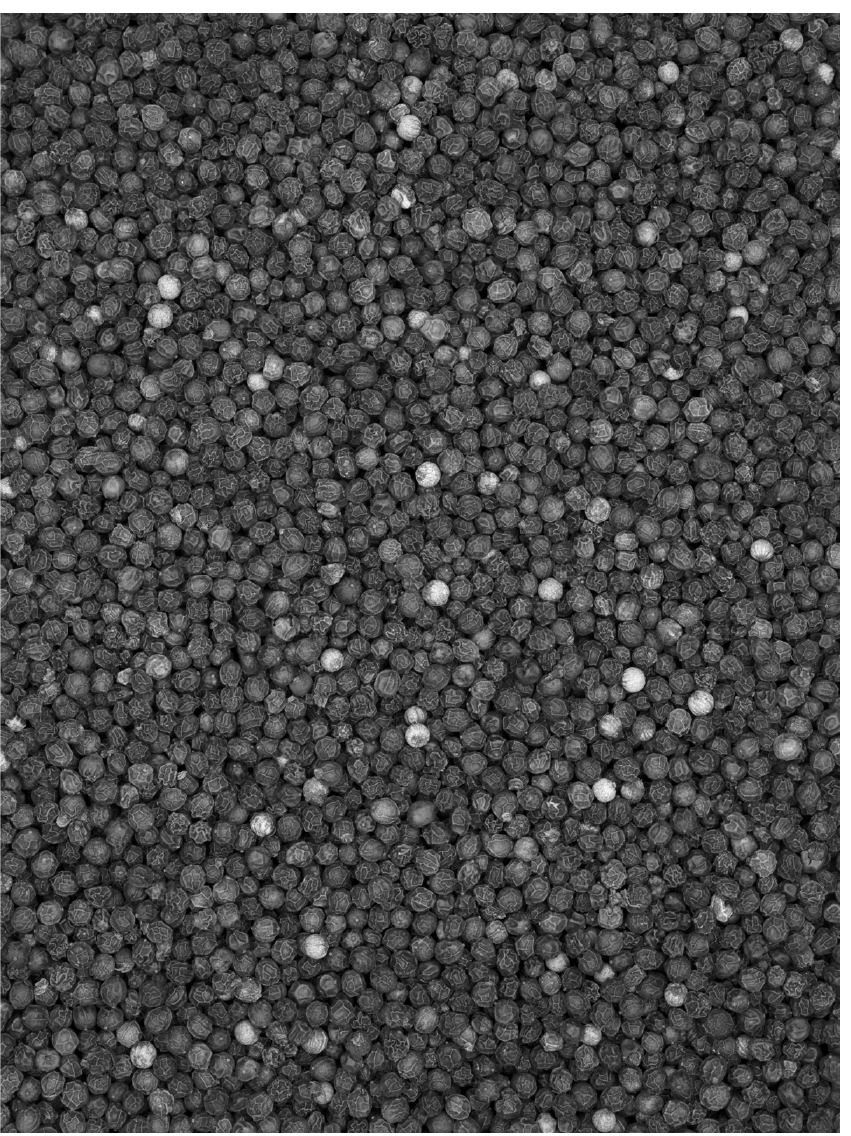}}
\subfigure{ \includegraphics[width=0.25\linewidth]{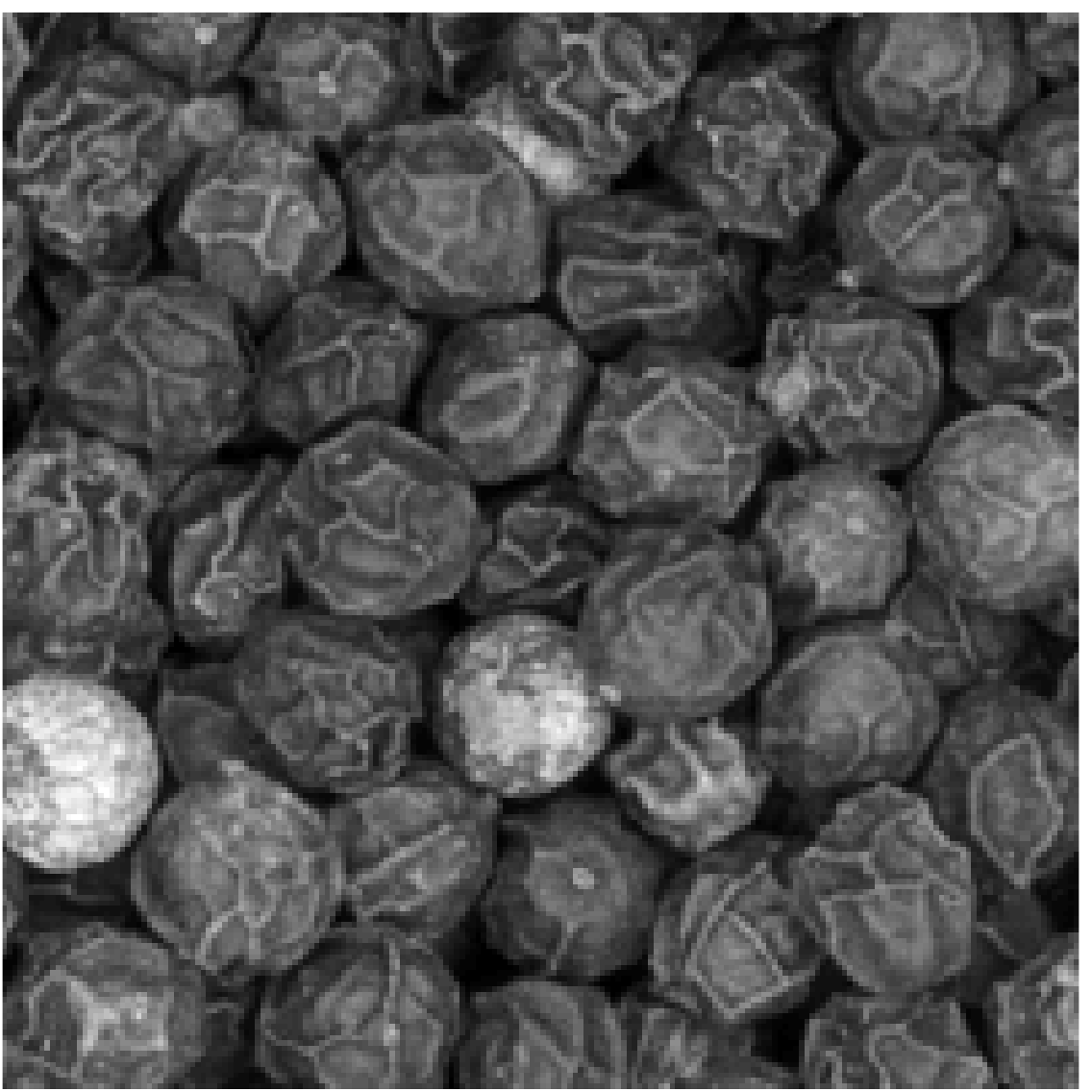}}
\caption{Left:\ the high-resolution image from which we obtain the
training image patches.
Right:\ the $200 \times 200$ exact image $x\ex$.}
\label{fig:Images}
\end{figure}

The test images used in Sections \ref{sec:DicComp}--\ref{sec:MoreTests}
are square patches from a high-resolution photo of peppers
with uneven surfaces resembling texture, making them interesting test images for studies of the
reconstruction of textures and structures with sharp boundaries.
Figure \ref{fig:Images} shows the $1600 \times 1200$ high-resolution image
and the exact image of dimensions $M \times N = 200\times 200$.
This size allows us to perform many numerical
experiments in a reasonable amount of time; we demonstrate the performance
of our algorithm on a larger test problem in Section \ref{sec:large}.

All test problems represent a parallel-beam tomographic measurement,
and we use the function \texttt{paralleltomo} from
\textsc{AIR Tools} \cite{HansenHansen} to compute the system matrix~$A$.
The data associated with a set of parallel rays is called a projection,
and number of rays in each projection is given by
$N_{\mathrm{r}} = \lfloor \sqrt{2}N \rfloor$.
If the total number of projections is $N_{\mathrm{p}}$ then the
number of rows in $A$ is $m = N_{\mathrm{r}}N_{\mathrm{p}}$
while the number of columns is $n = MN$.
Recall that we are interested in scenarios with a small number of projections.
The exact data is generated with the forward model
after which we add Gaussian white noise, i.e., $b = Ax\ex+e$.

\subsection{Studies of the Dictionary Learning Stage}
\label{sec:DicComp}

A good dictionary should preserve the structural information of the training images
as much as possible and, at the same time, admit a
sparse representation as well as a small representation error.
These requirements are related to the number of dictionary elements,
i.e., the number of columns $s$ in the matrix $D \in \mathbb{R}^{p\times s}$.
Since we want a compressed representation of the training images
we choose $s$ such that $p \leq s \ll t$, and the precise value will be investigated.
The optimal patch size $P\times Q$ is unclear and will also be studied;
without loss of generality we assume $P=Q$.

The regularization parameter $\lambda$ in \eqref{e-dictionary-learn}
balances the matrix factorization error and the sparsity constraint on
the elements of the matrix $H$.
The larger the $\lambda$, the more weight is given to minimization
of $\| H \|_{\mathrm{sum}}$, while for small $\lambda$ more weight is given
to minimization of the factorization error.
If $\lambda=0$ then \eqref{e-dictionary-learn} reduces to the
classical nonnegative matrix factorization problem.

From the analysis of the upper bound on the regularization parameter $\lambda$
in the Appendix, 
we know $\lambda \geq p$ implies $H=0$; so $\lambda$ can be varied in the interval $(0,p]$
to find dictionaries with different sparsity priors.
Note that the scaling of the training images affects the scaling of
the matrix $H$ as well as the regularization parameter~$\lambda$.

To evaluate the impact of the dictionary parameters,
we use three different patch sizes ($5\times 5$, $10\times 10$,
and $20 \times 20$) and the number of dictionary elements $s$ is chosen to be
2, 3, and 4 times the of the number of rows $p$ in the dictionary $D$.
We extract more than $50,000$ overlapping patches from the high-resolution image
in Figure~\ref{fig:Images}.
For different combinations of patch sizes and number of dictionary elements we
solve the dictionary learning problem~\eqref{e-dictionary-learn}.

\begin{figure}[ht!]%
 \centering
\subfigure[{$5 \times 5$, $s=100$}]{ \includegraphics[width=0.3\linewidth]{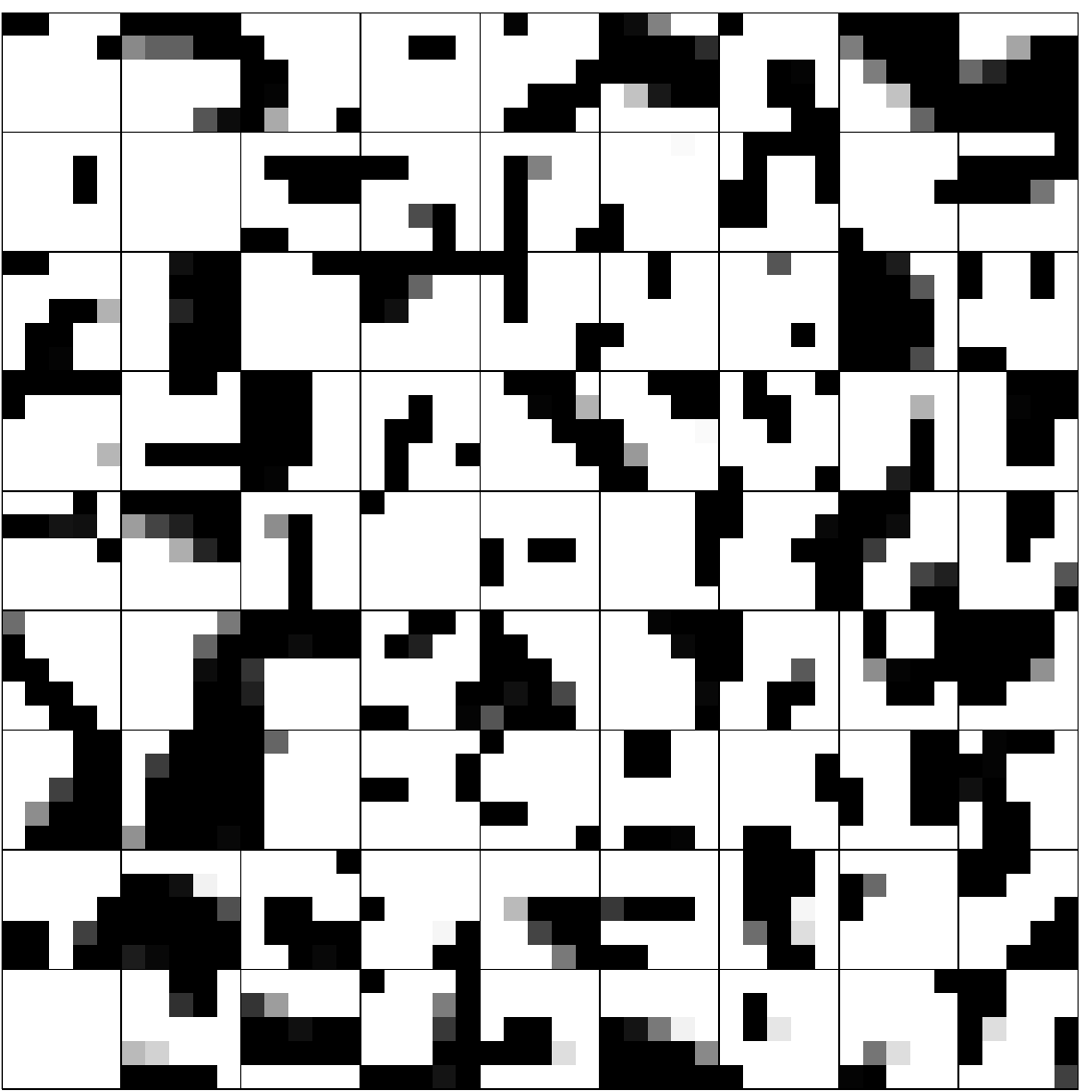}}
\subfigure[{$10 \times 10$, $s=300$}]{ \includegraphics[width=0.3\linewidth]{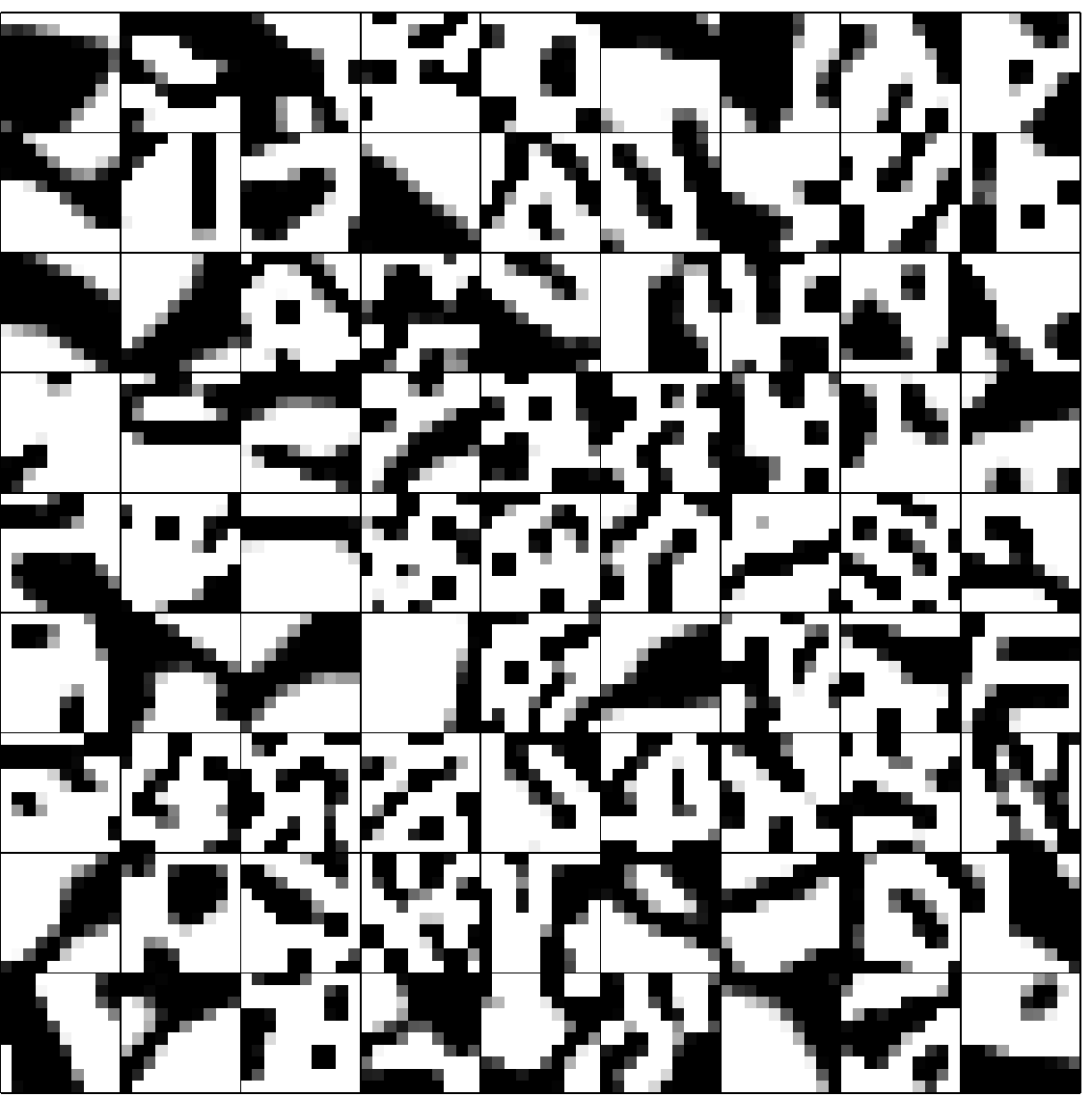}}
\subfigure[{$20 \times 20$, $s=800$}]{ \includegraphics[width=0.3\linewidth]{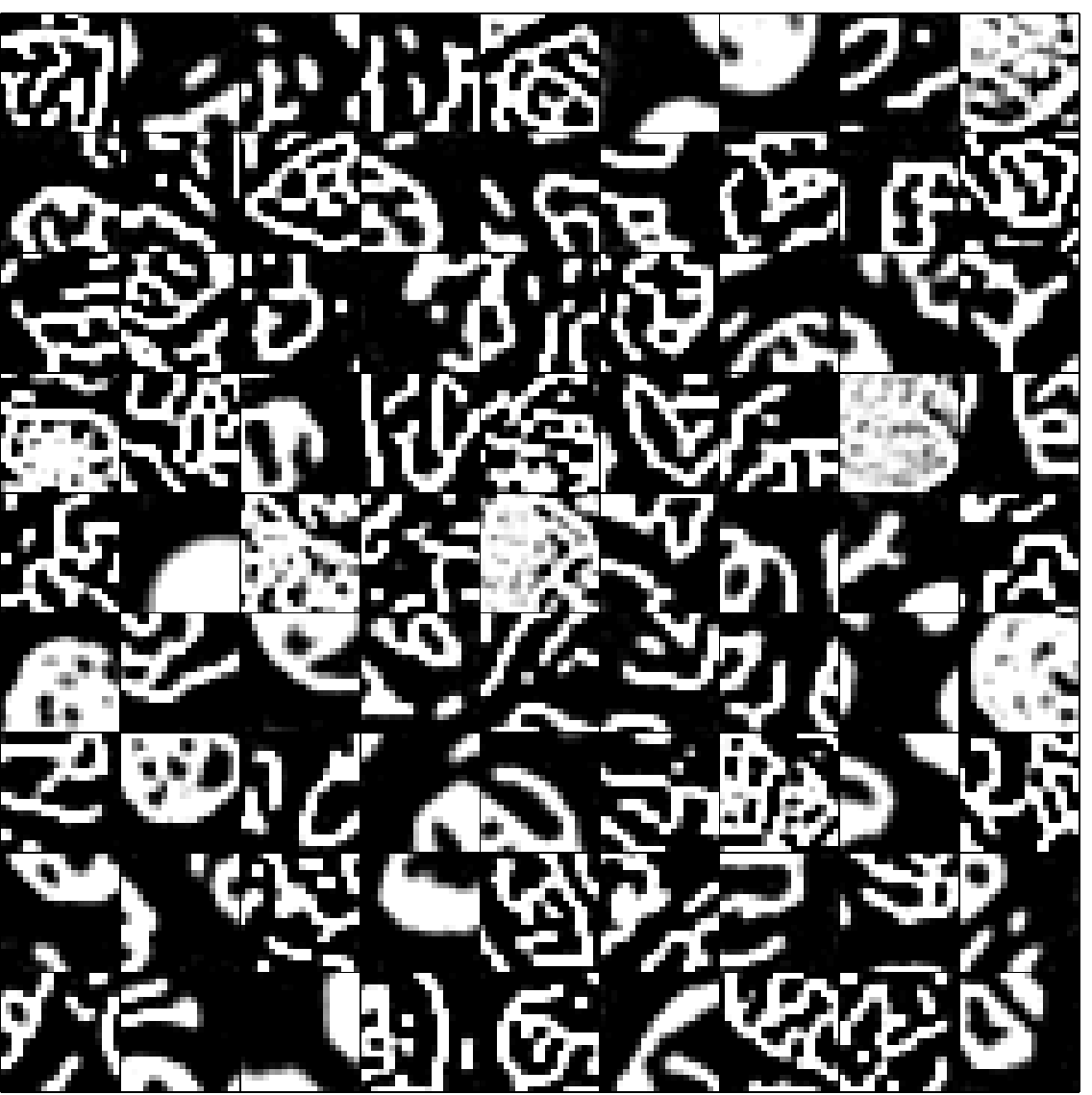}} \\
\subfigure[{$5 \times 5$, $s=100$}]{ \includegraphics[width=0.3\linewidth]{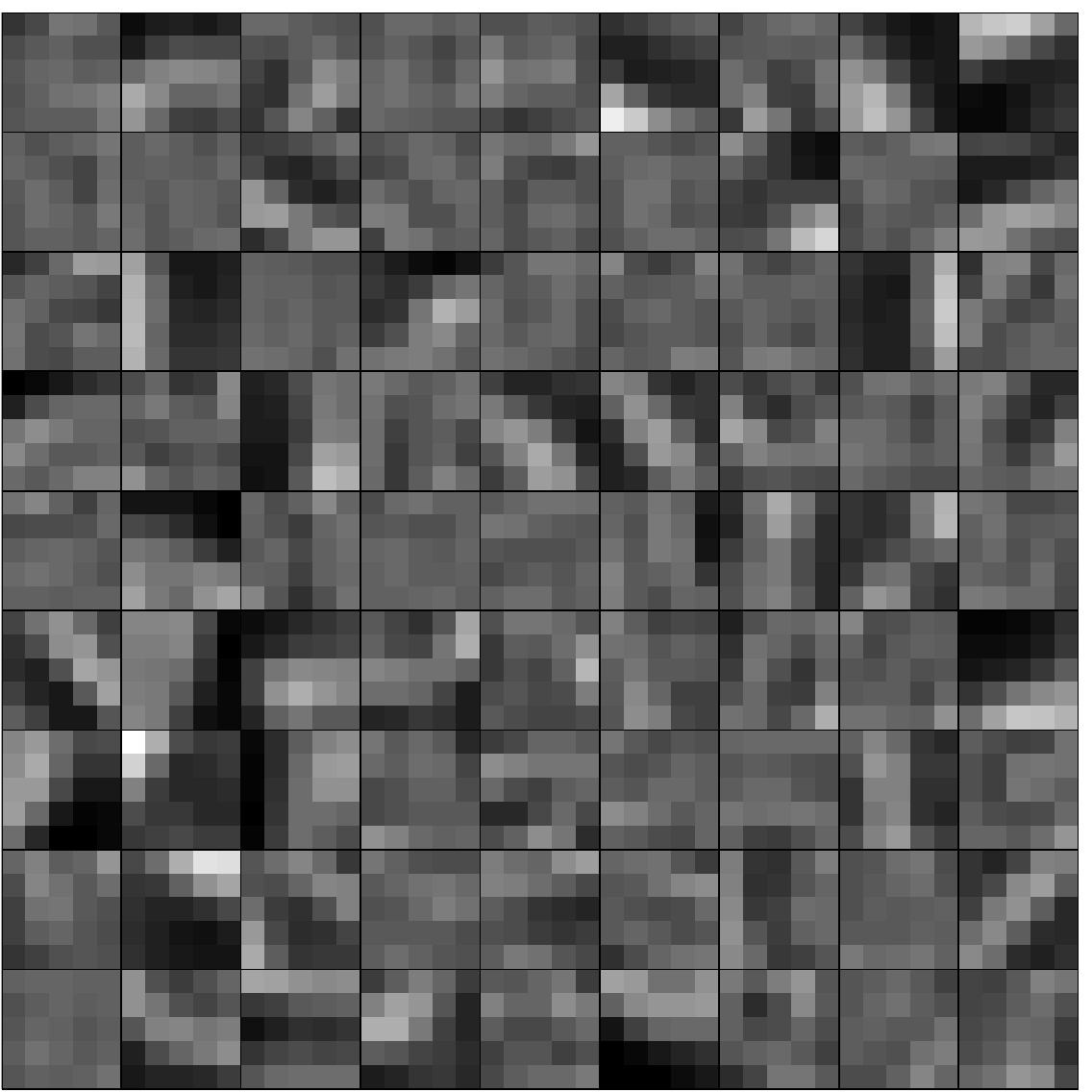}}
\subfigure[{$10 \times 10$, $s=300$}]{ \includegraphics[width=0.3\linewidth]{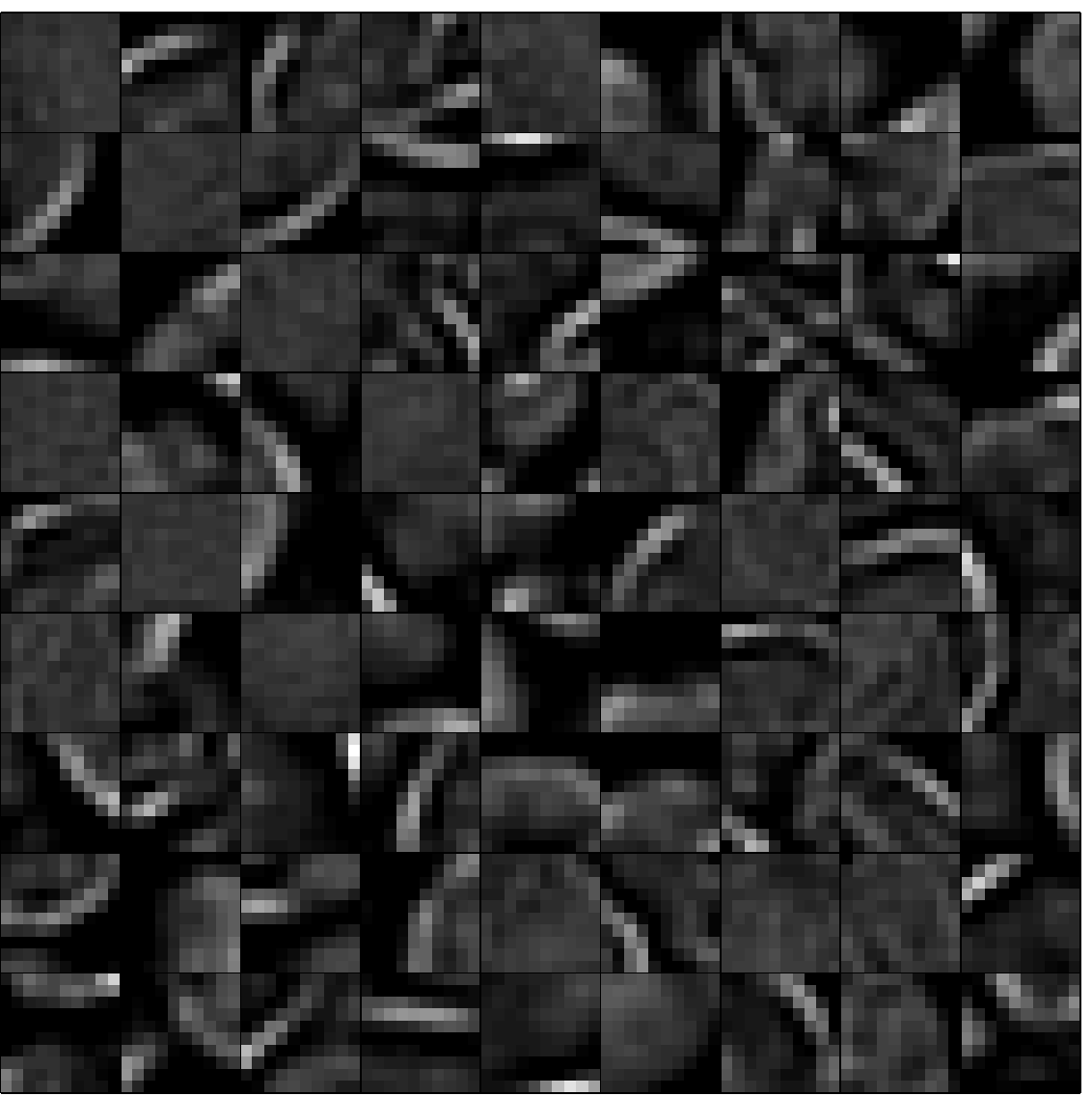}}
\subfigure[{$20 \times 20$, $s=800$}]{ \includegraphics[width=0.3\linewidth]{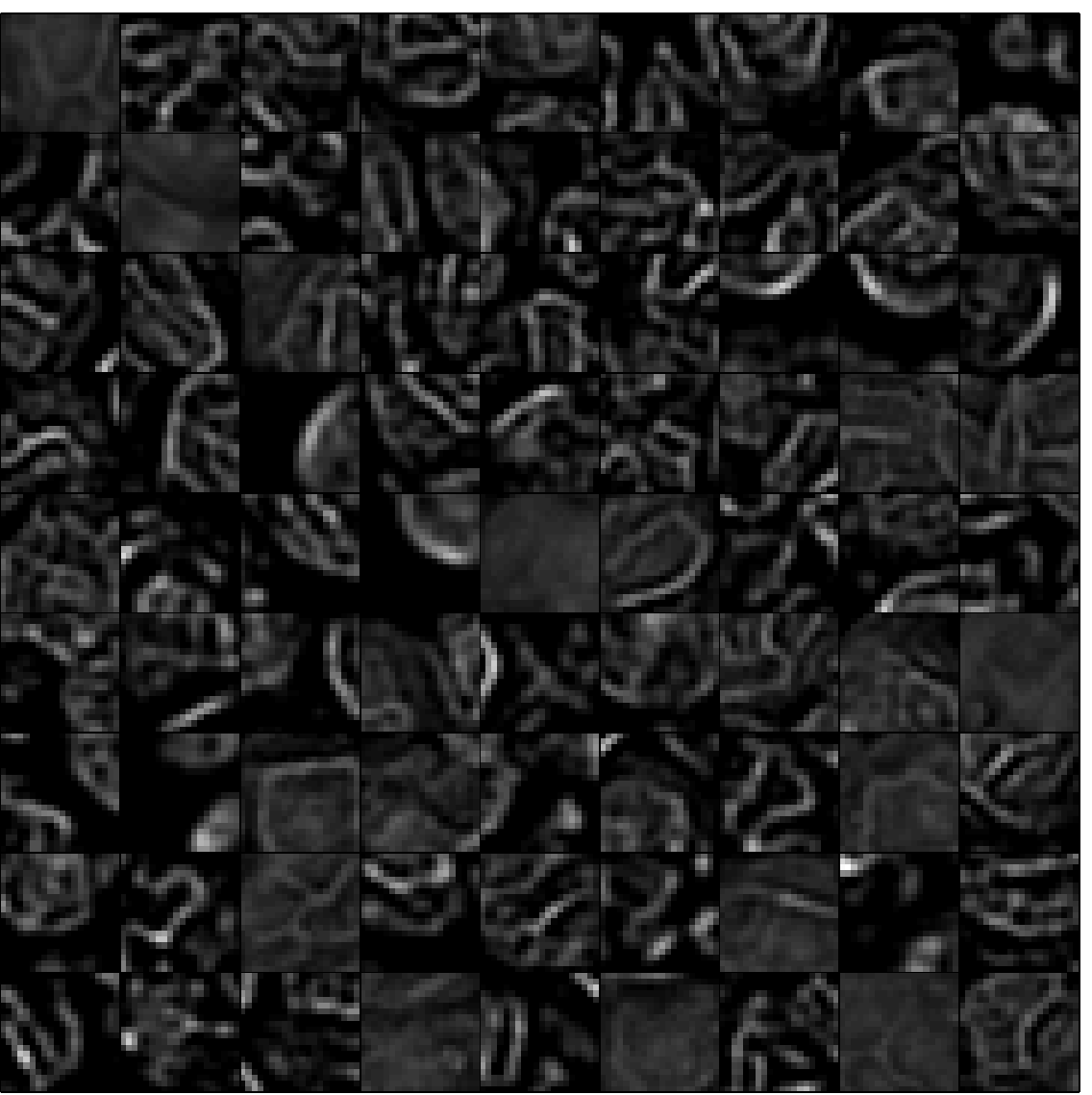}}
\caption{Examples of dictionary elements.
Top row:\ with the constraint $D \in \mathcal{D}_{\infty}$ the images appear as
``binary looking.''  Bottom row:\ with the constraint $D \in \mathcal{D}_2$ the images
appear to use the whole gray-scale range.}
 \label{fig:Dics}
\end{figure}

Figure \ref{fig:Dics} shows examples of such learned dictionaries,
where columns of $D$ are represented as images;
we see that the penalty constraint $D \in \mathcal{D}_{\infty}$ gives rise to
``binary looking'' dictionary elements while $D \in \mathcal{D}_2$ results
in dictionary elements that use the whole gray-scale range.

To evaluate the approximation error, i.e., the distance of the exact image $x\ex$ to
its projection on the cone $\mathcal{C}$ \eqref{eq:cone},
we compute the solutions $\alpha_j^\star$ to the
$q$ approximation problems for all blocks $j=1,2,\ldots,q$ in $x\ex$,
  \begin{equation}
  \label{eq:BestRep}
    \min_{\alpha_j} \half \bigl\| D \alpha_j - x\ex_j \bigr\|_2^2  \qquad
    \mathrm{s.t.} \qquad \alpha_j \geq 0 .
  \end{equation}
Then $P_{\mathcal{C}}(x_j\ex) = D \alpha_j^\star$ is the best
representation/approximation of the $j$th block in the cone.
The mean approximation error (MAE) is then computed as
  \begin{equation}
  \label{eq:MeanRep}
    \mathrm{MAE} = \frac{1}{q} \sum_{j=1}^q \frac{1}{\sqrt{p}} \bigl\|P_{\mathcal{C}}(x_j\ex)  - x_j\ex \bigr\|_2 .
  \end{equation}

\begin{figure}[ht!]%
 \centering
\subfigure{ \includegraphics[width=0.3\linewidth]{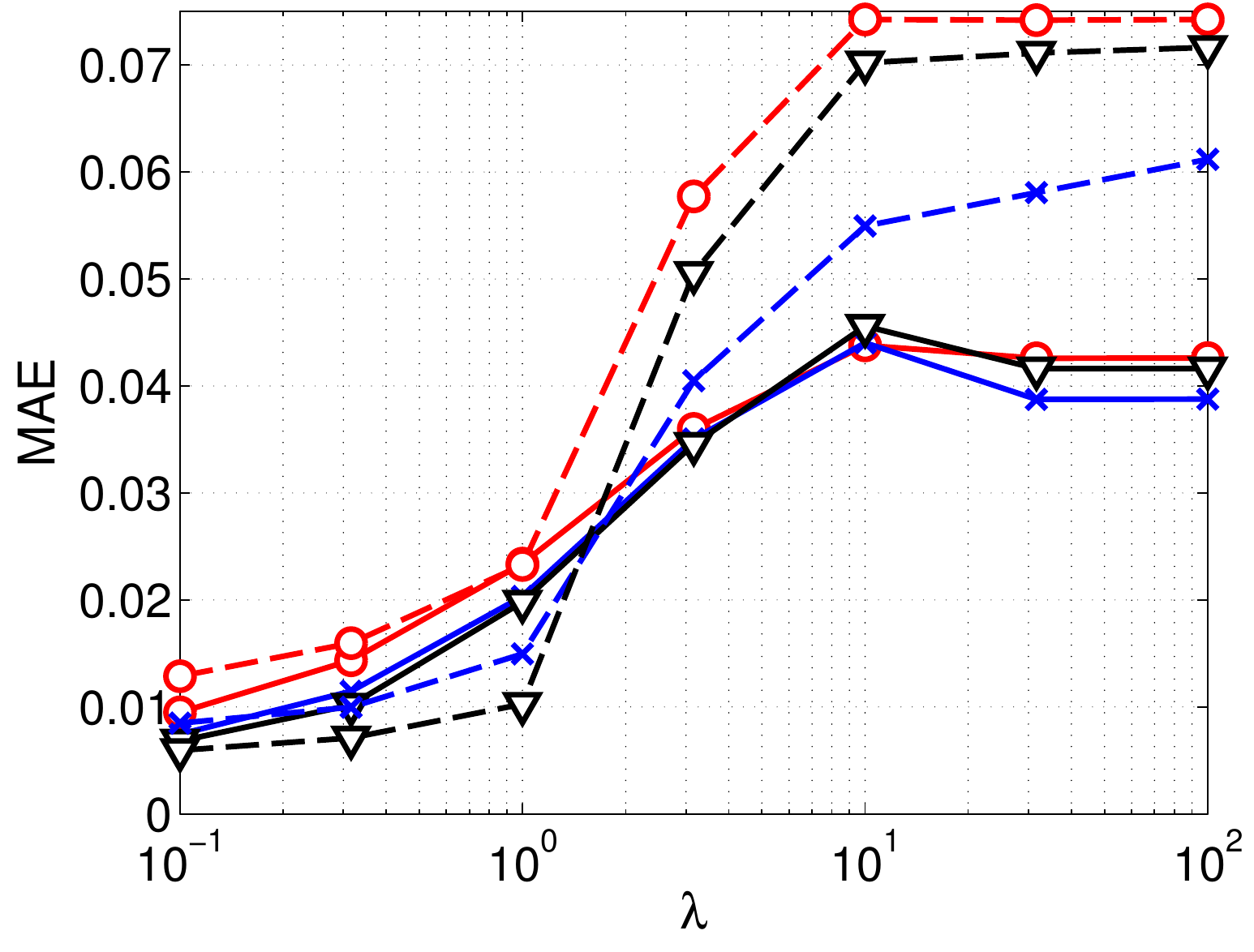}}~
\subfigure{ \includegraphics[width=0.3\linewidth]{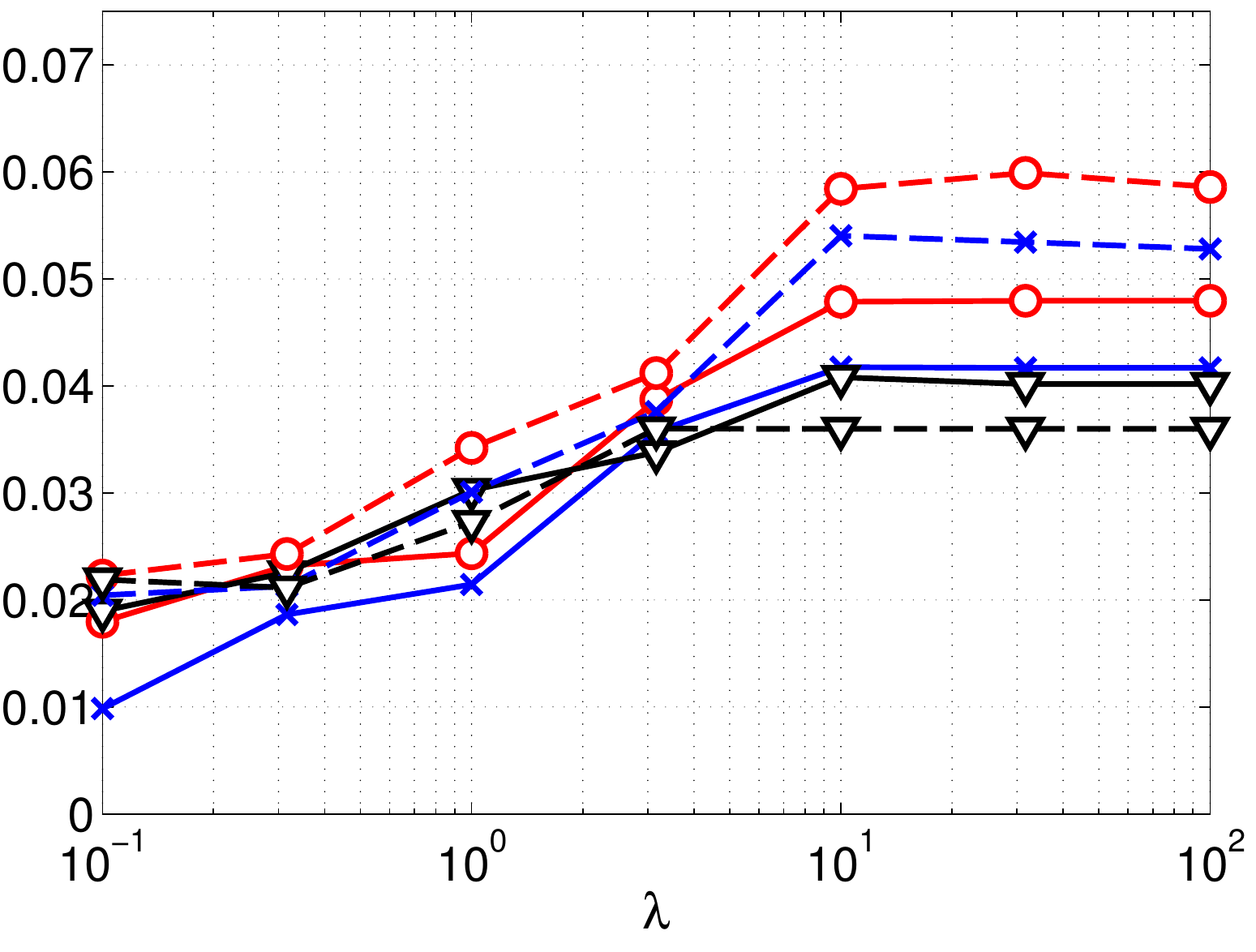}}~
\subfigure{ \includegraphics[width=0.3\linewidth]{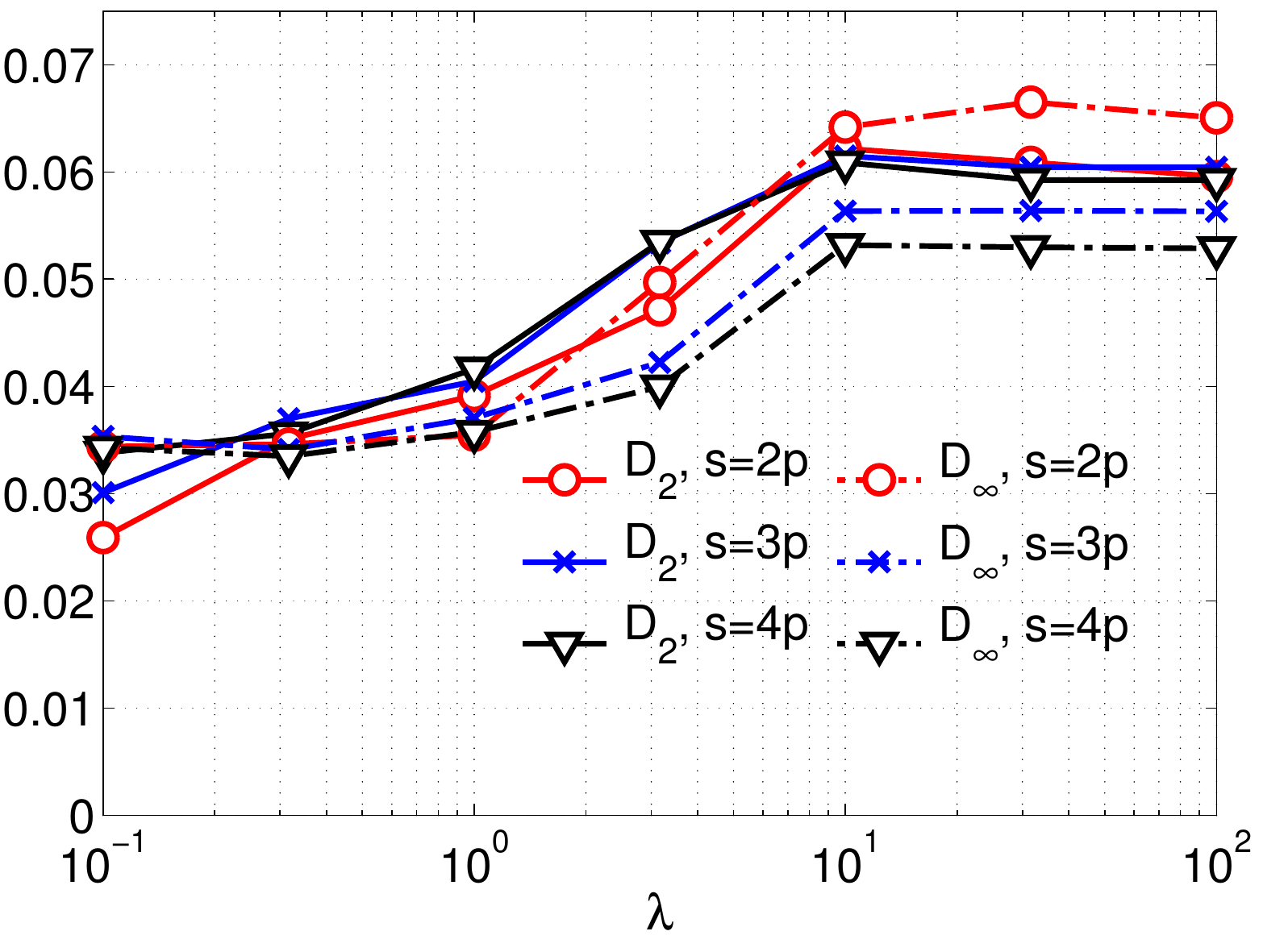}} \\
{\footnotesize \hspace{5mm} $5 \times 5$ patches \hspace{20mm} $10 \times 10$ patches
  \hspace{20mm} $20 \times 20$ patches}
\caption{Mean approximation errors \eqref{eq:MeanRep}
  for both $D \in \mathcal{D}_{\infty}$ and $D \in \mathcal{D}_2$
  with different patch sizes and different~$s$.}
\label{fig:ResultsDic}
\end{figure}

The ability of the dictionary to represent features and textures from the training images,
which determines how good reconstructions we are able to compute,
depends on the regularization parameter $\lambda$,
the patch size, and the number of dictionary elements.
Figure \ref{fig:ResultsDic} shows how the mean approximation error MAE \eqref{eq:MeanRep}
associated with the dictionary varies with patch size $p$, number of
dictionary elements $s$, and regularization parameter~$\lambda$.
An advantage of larger patch sizes is that the variation of MAE with
$s$ and $\lambda$ is less pronounced than for small patch sizes, so overall we tend to prefer larger patch sizes.
In particular, for a large patch size we can use a smaller over-representation
factor $s/p$ than for a small patch size.
As $\lambda$ approaches $p$ we have that $\|H\|_{\mathrm{sum}}$ approaches 0,
the dictionary $D$ takes arbitrary values, and the approximation errors level off
at a maximum value.
Regarding the two different constraints $D \in \mathcal{D}_{\infty}$ and $D \in \mathcal{D}_2$
we do not see any big difference in the approximation errors for $10\times 10$ and
$20\times 20$ patches; to limit the amount of results we now use $\mathcal{D}_2$.

The computational work depends on the patch size and
the number of dictionary elements which, in turn,
affects the approximation error: the larger the dictionary, the smaller the approximation
error, but at a higher computational cost. We have found that a good trade-off between the
computational work and the approximation error can be obtained by increasing the
number of dictionary elements until the approximation error levels off.

\subsection{Studies of the Reconstruction Stage}
\label{sec:RecsTests}

Here we evaluate the overall reconstruction framework
including the effect of the reconstruction parameters as well as their connection to the
dictionary learning parameter~$\lambda$ and the patch size.

We solve the reconstruction problem \eqref{eq:mainRec} using
projection data based on the exact image
given in Figure~\ref{fig:Images}.
We choose $N_{\mathrm{p}} = 25$ uniformly distributed projection
angles in $[0^\circ,180^\circ]$.
Hence the matrix $A$ has dimensions $m= 7,050$
and $n=40,000$, so the problem is highly underdetermined.
We use the relative noise level $\| e \|_2 / \| Ax\ex \|_2 = 0.01$.
Moreover, we use $5 \times 5$, $10 \times 10$ and $20 \times 20$ patches
and corresponding dictionary matrices $D^{(5)}$, $D^{(10)}$, and $D^{(20)}$
in $\mathcal{D}_2$
of size $25 \times 100$, $100 \times 300$, and $400 \times 800$, respectively.
Examples of the dictionary elements
are shown in the bottom row of Figure \ref{fig:Dics}.

We first investigate the reconstruction's sensitivity  to the choice
of $\lambda$ in the dictionary learning problem and
the parameters $\mu$ and $\delta$ in the reconstruction problem.
It follows from the optimality conditions of \eqref{eq:mainRec} that
$\alpha^\star=0$ is optimal when
$\mu \geq \bar{\mu} = \frac{q}{m} \| (I\otimes D^T)\Pi A^Tb \|_{\infty}$
and hence we choose $\mu \in [0,\bar{\mu}]$.
Large values of $\mu$ refer to the case where the sparsity prior is strong
and the solution is presented with too few dictionary elements.
On the other hand if $\mu$ is small and a sufficient number of dictionary
elements are included, the reconstruction error worsens only slightly
when $\mu$ decreases.
In the next subsection we show that we may, indeed, obtain reasonable
reconstructions for $\mu=0$.

\begin{figure}[ht!]%
\centering
\subfigure[$5 \times 5$ patches]{ \includegraphics[height=0.3\linewidth]{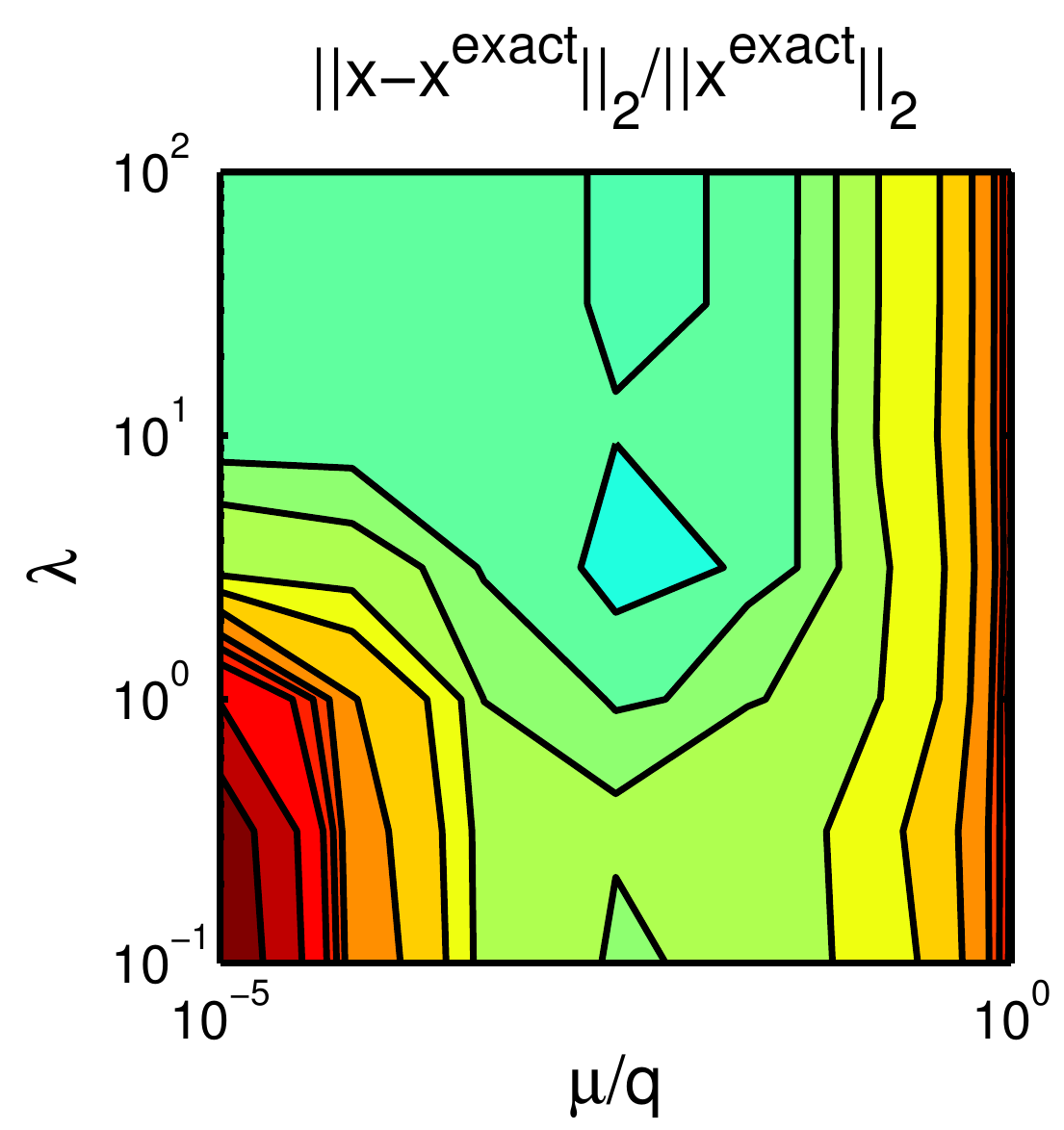}}~
\subfigure[$10 \times 10$ patches]{ \includegraphics[height=0.3\linewidth]{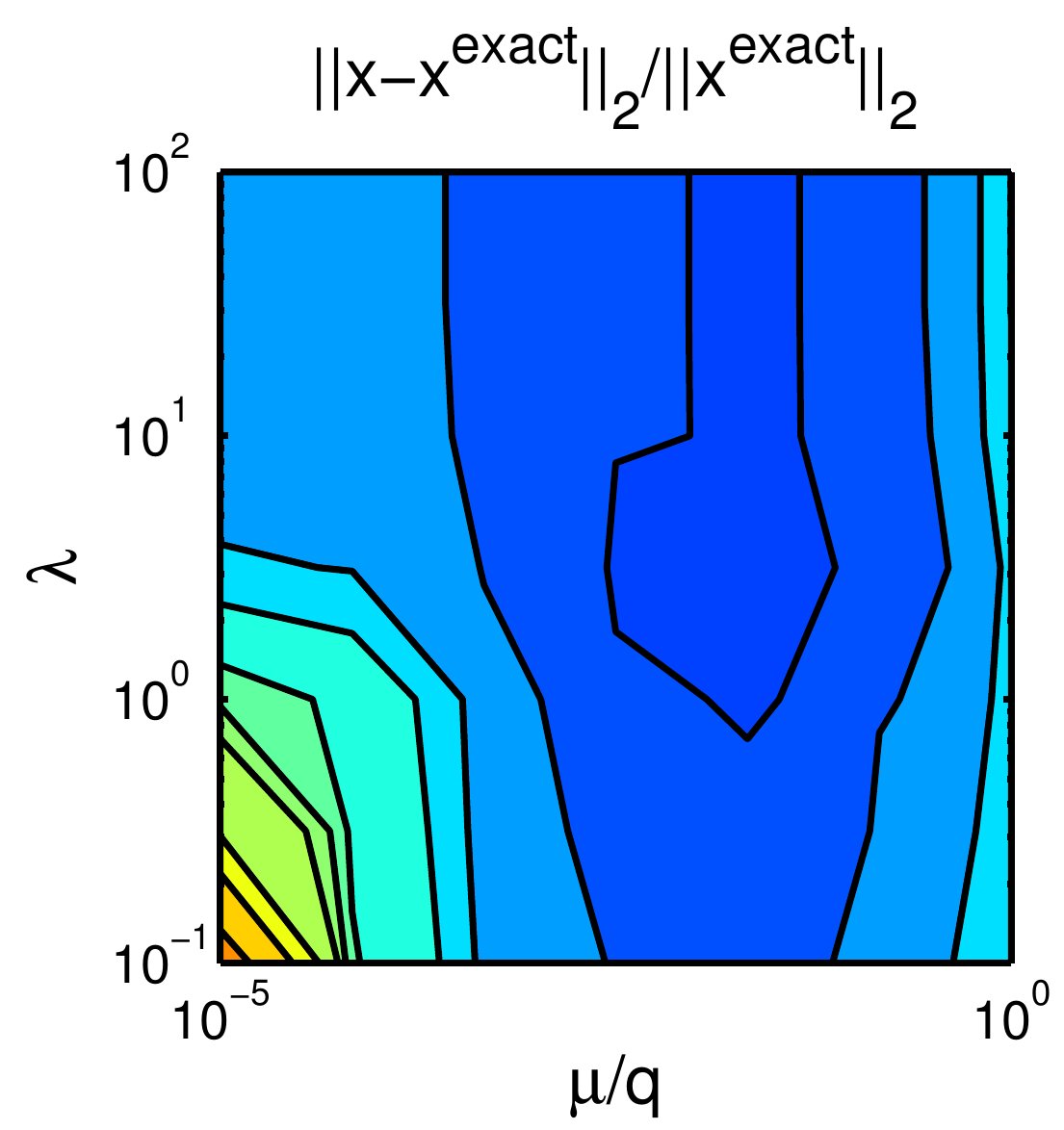}}~
\subfigure[$20 \times 20$ patches]{ \includegraphics[height=0.3\linewidth]{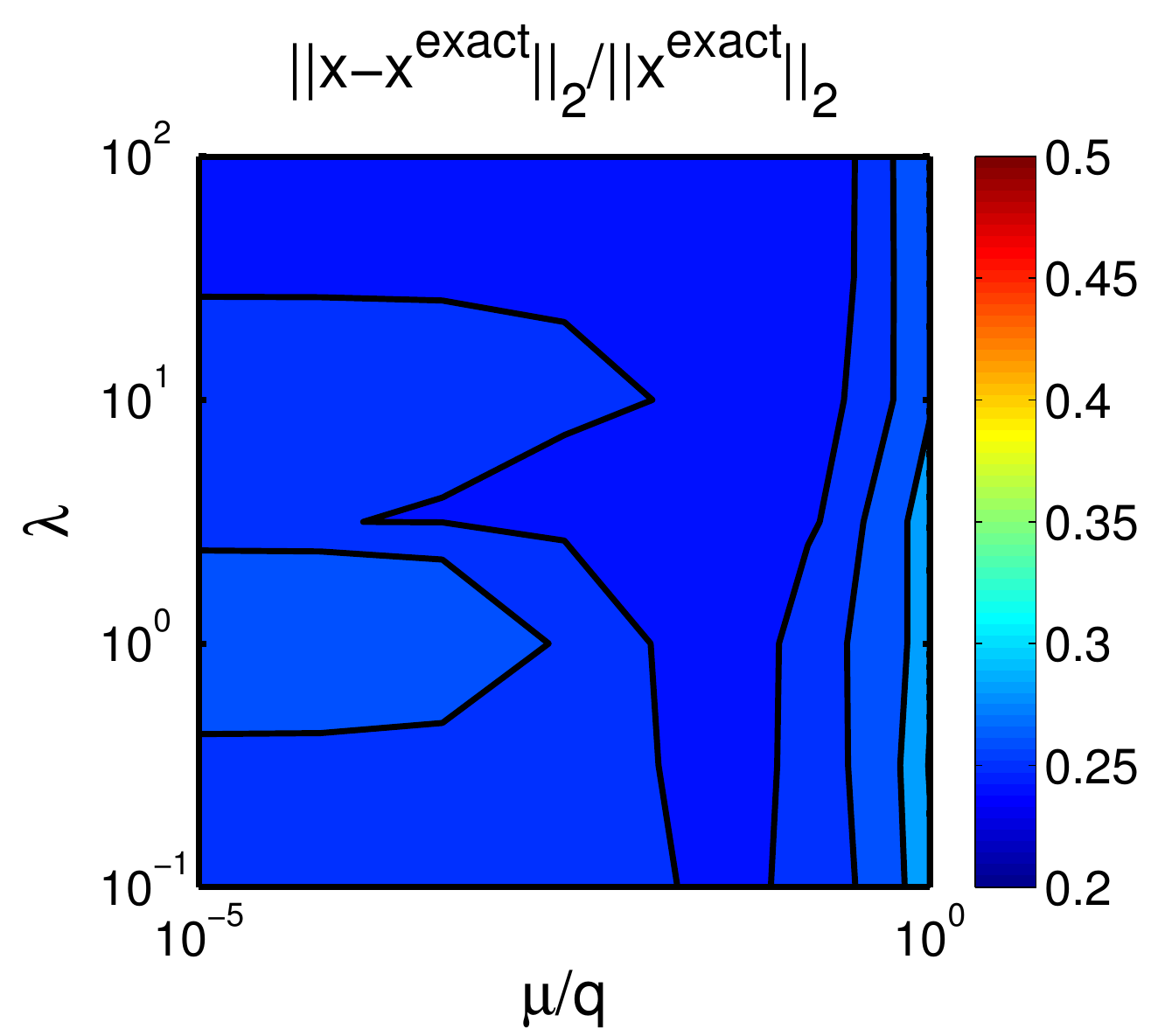}}
\caption{Contour plots of the reconstruction
  error RE \eqref{eq:RE} versus $\lambda$ and $\mu/q$.}
  \label{fig:ErrorTauLambda}
\end{figure}

To investigate the effect of regularization parameters $\lambda$ and $\mu$,
we first perform experiments with $\delta=0$ corresponding to no image prior.
The quality of a solution $x$ is evaluated by the \emph{reconstruction error}
  \begin{equation}
  \label{eq:RE}
    \mathrm{RE} = \| x-x\ex \|_2/\| x \ex \|_2
  \end{equation}
and Figure \ref{fig:ErrorTauLambda} shows contour plots of RE as a function
of $\lambda$ and $\mu/q$.
The reconstruction error is smaller for larger patch sizes, and also less
dependent on the regularization parameter $\lambda$ and the normalized regularization parameter $\mu/q$.
The smallest reconstruction errors are obtained in all dictionary sizes
for $\lambda \approx 3$.

\begin{figure}[ht!]%
\centering
\subfigure[$5 \times 5$ patches]{ \includegraphics[height=0.3\linewidth]{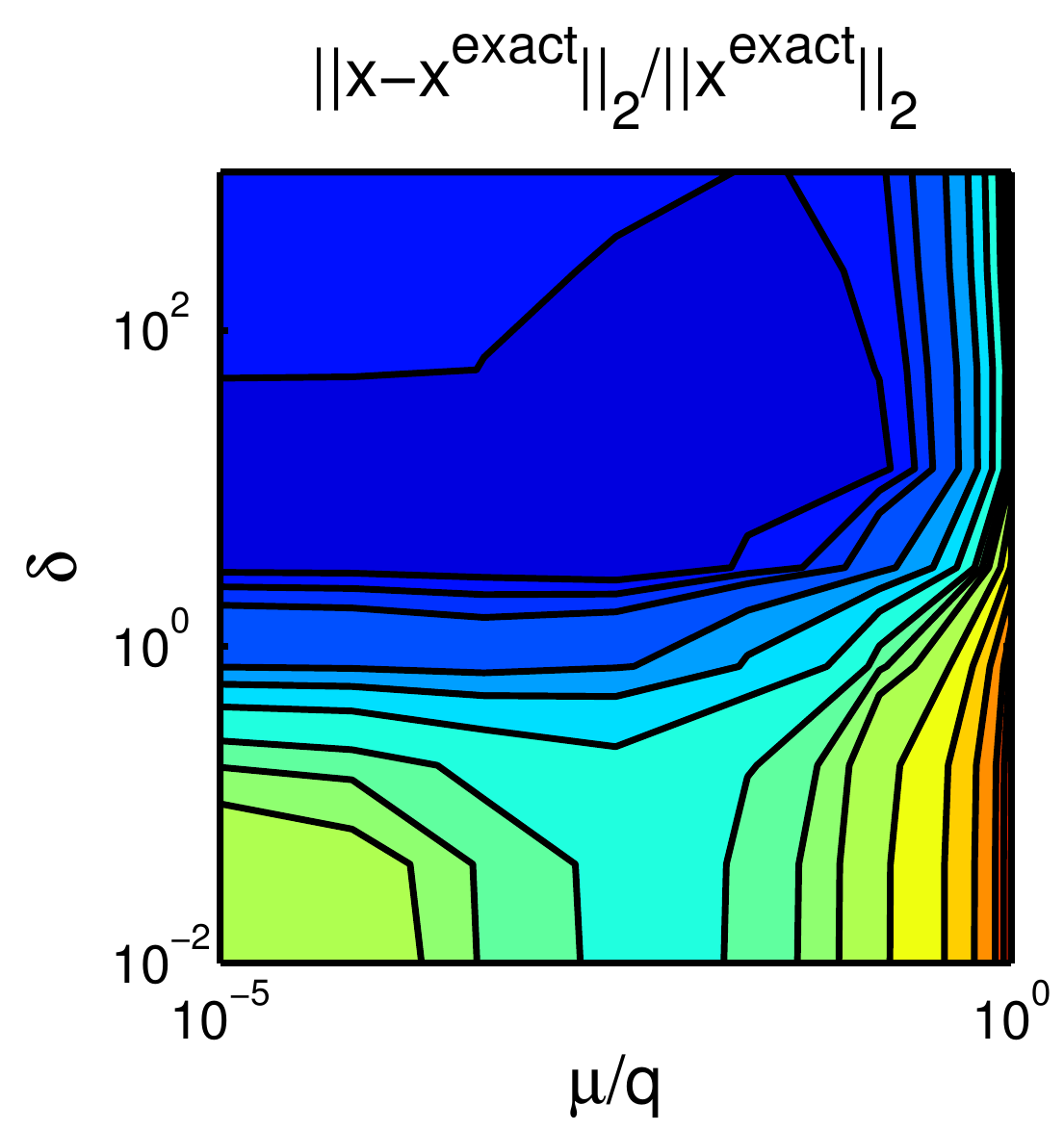}}~
\subfigure[$10 \times 10$ patches]{ \includegraphics[height=0.3\linewidth]{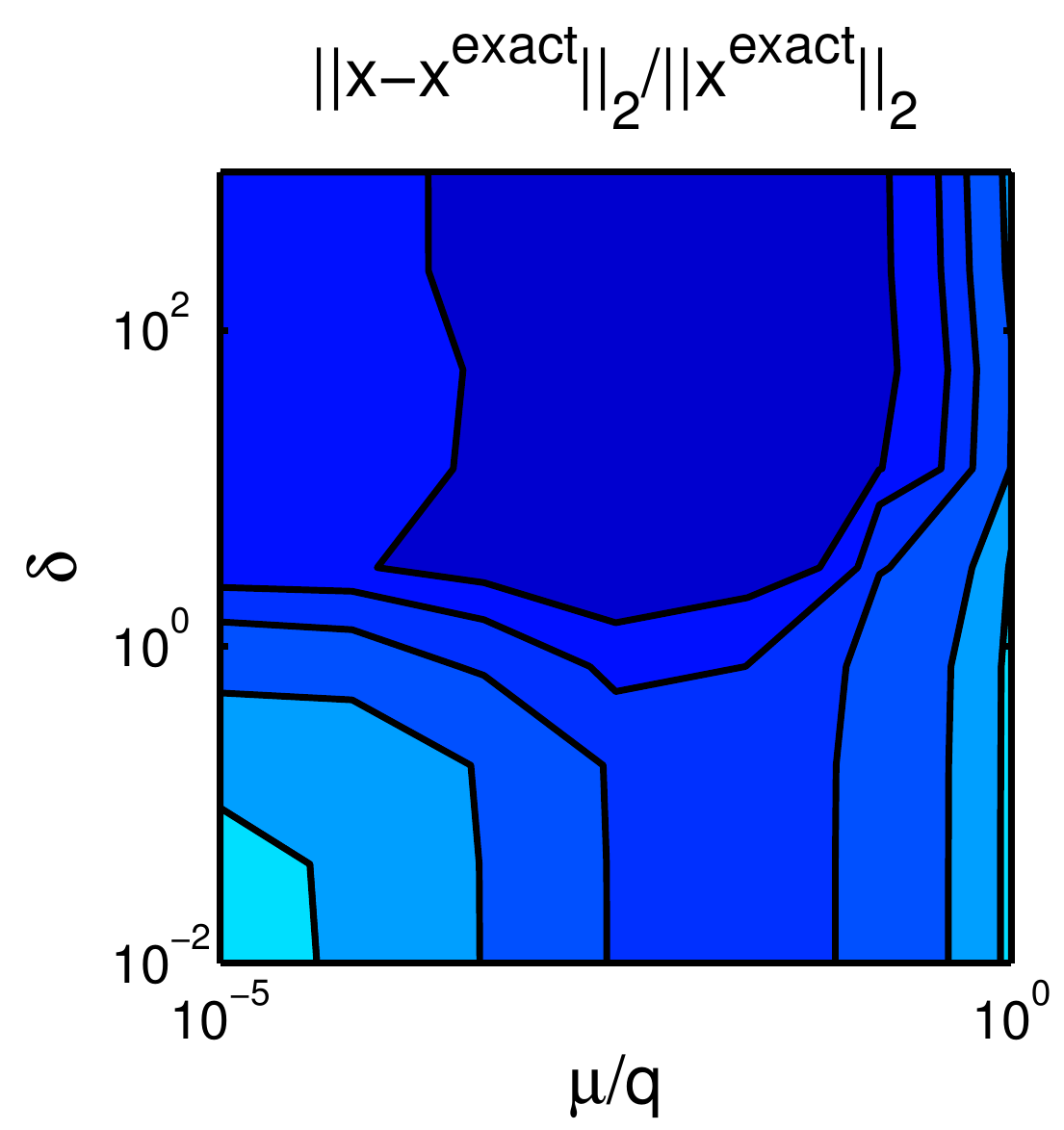}} ~
\subfigure[$20 \times 20$ patches]{ \includegraphics[height=0.3\linewidth]{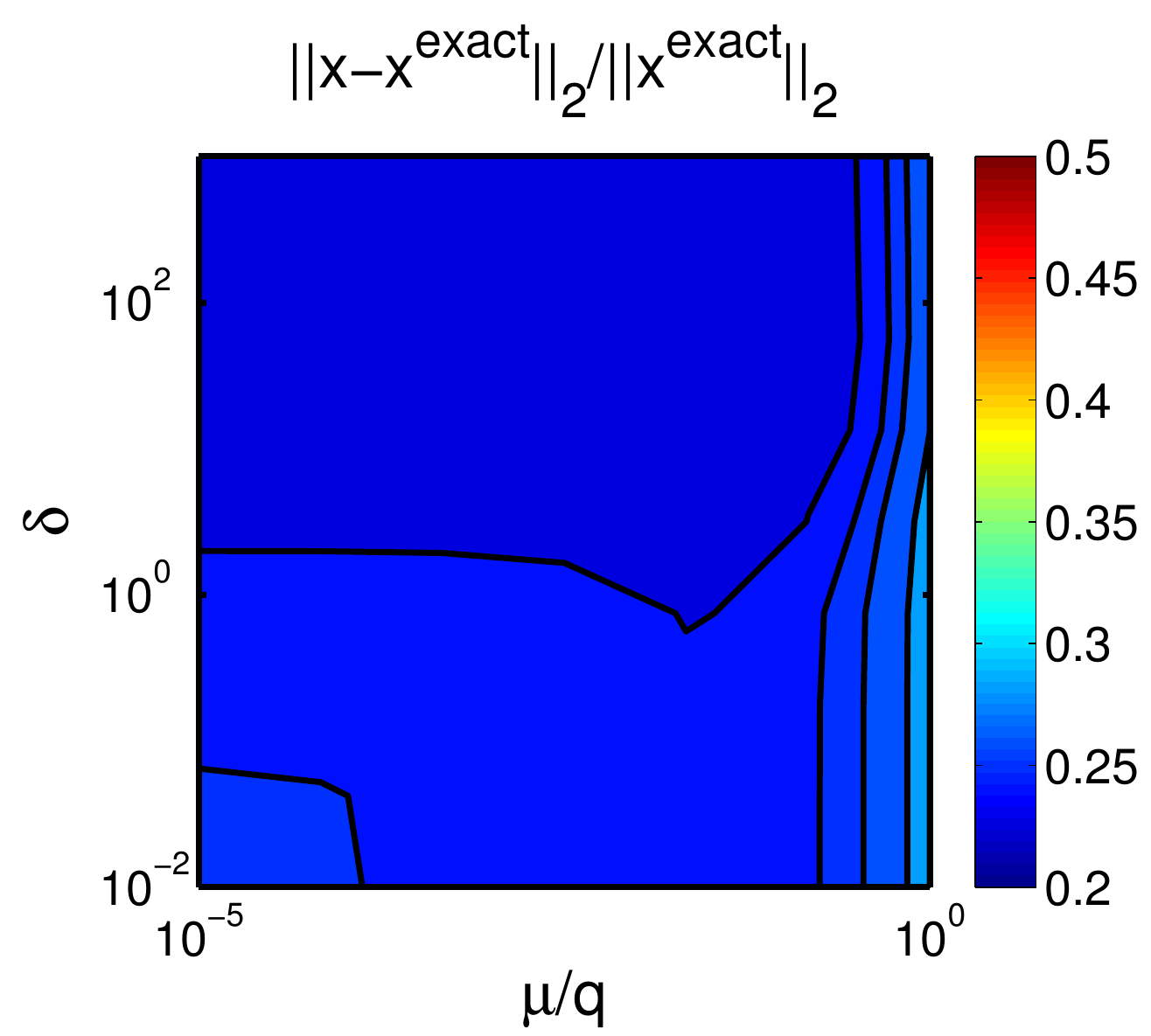}}
\caption{Contour plots of the reconstruction errors RE \eqref{eq:RE}
versus $\mu/q$ and $\delta$ for a fixed $\lambda=3.16$.}
 \label{fig:ErrorDeltaLambda}
 \end{figure}

Let us now consider the reconstructions when $\delta >0$
in order to reduce block artifacts.
Figure \ref{fig:ErrorDeltaLambda} shows contour plots of the reconstruction error
versus $\mu/q$ and $\delta$, using a fixed $\lambda=3.16$.
It is no surprise that introducing $\delta$
acts as a regularizer that can significantly reduce blocking artifacts and
thus improve the reconstruction.
Sufficiently large values of $\delta$ yield smaller reconstruction errors.
Consistent with the results from Figure \ref{fig:ErrorTauLambda},
the reconstruction errors
are smaller for $10 \times 10$ and $20 \times 20$ patch sizes than for $5 \times 5$ patches.
For larger patch sizes (which allow for capturing more structure
in the dictionary elements) the reconstruction error is quite
insensitive to the choice of $\delta$ and $\mu$.
The contour plots in Figure \ref{fig:ErrorDeltaLambda}
suggest that with our problem specification, we should choose $\delta \geq 1$.

\begin{figure}[ht!]%
\centering
\subfigure[FBP, RE = 0.481]{ \includegraphics[width=0.3\linewidth]{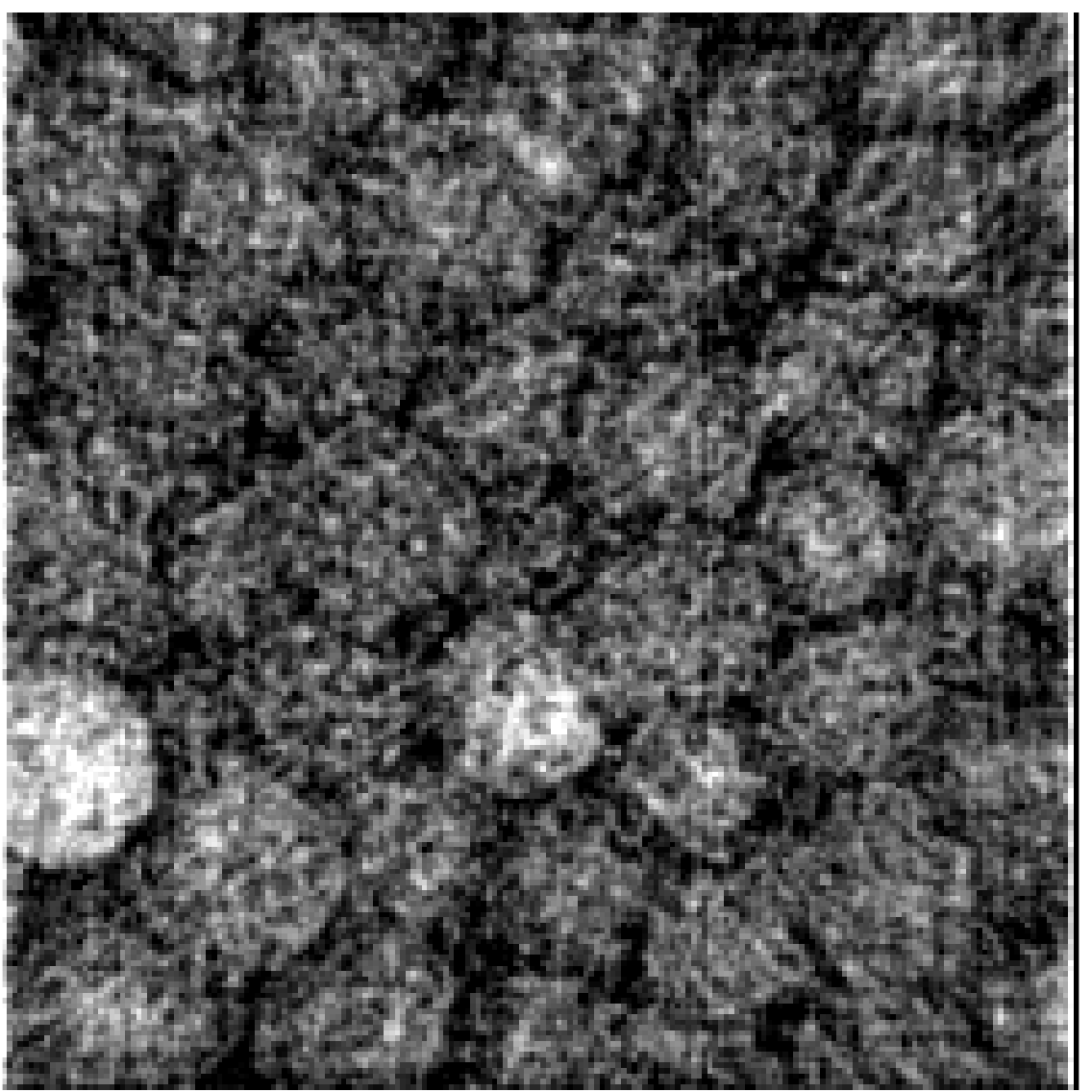}}~
\subfigure[ART, RE = 0.225]{ \includegraphics[width=0.3\linewidth]{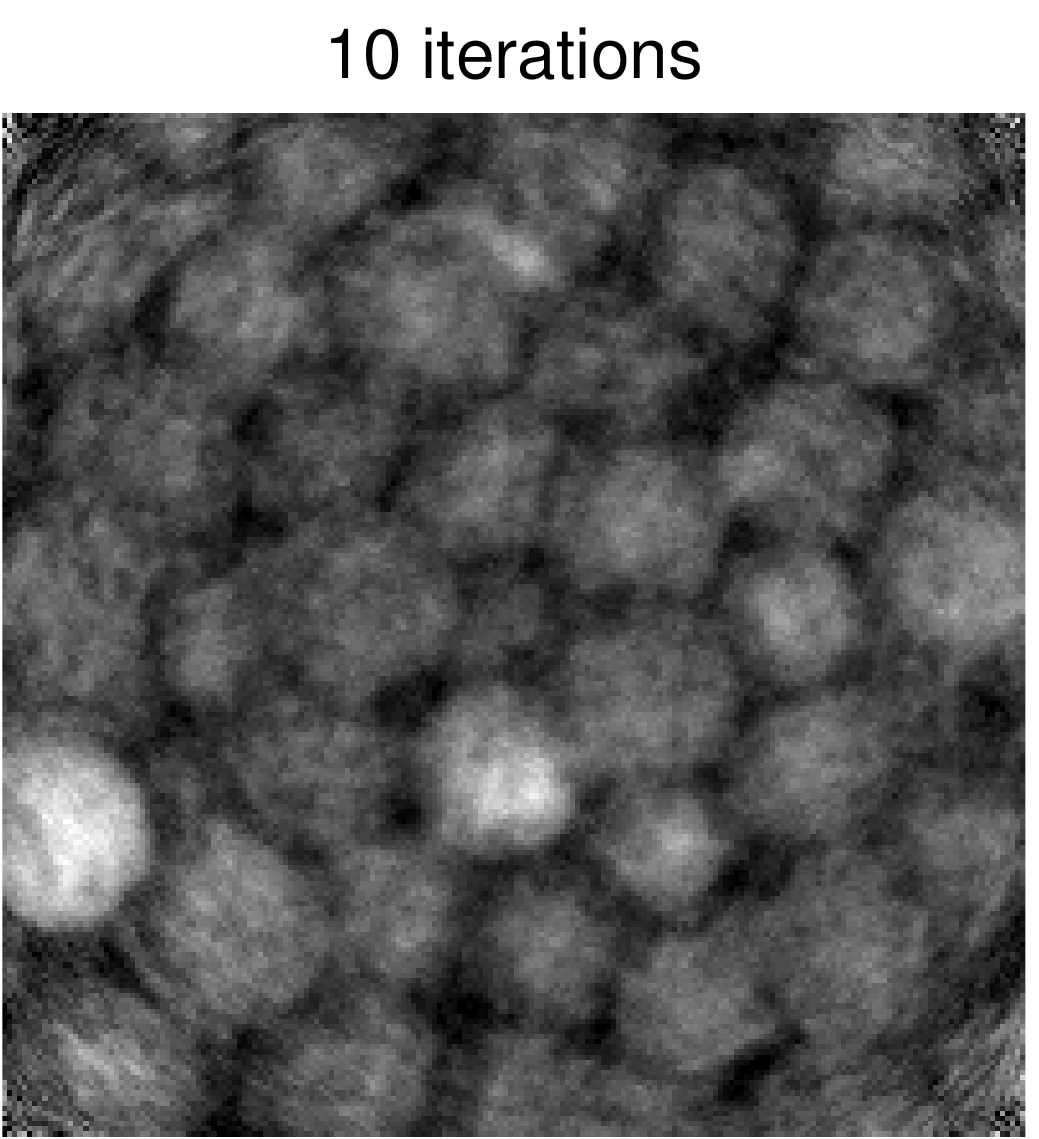}}~
\subfigure[{TV, RE = 0.214}]{ \includegraphics[width=0.3\linewidth]{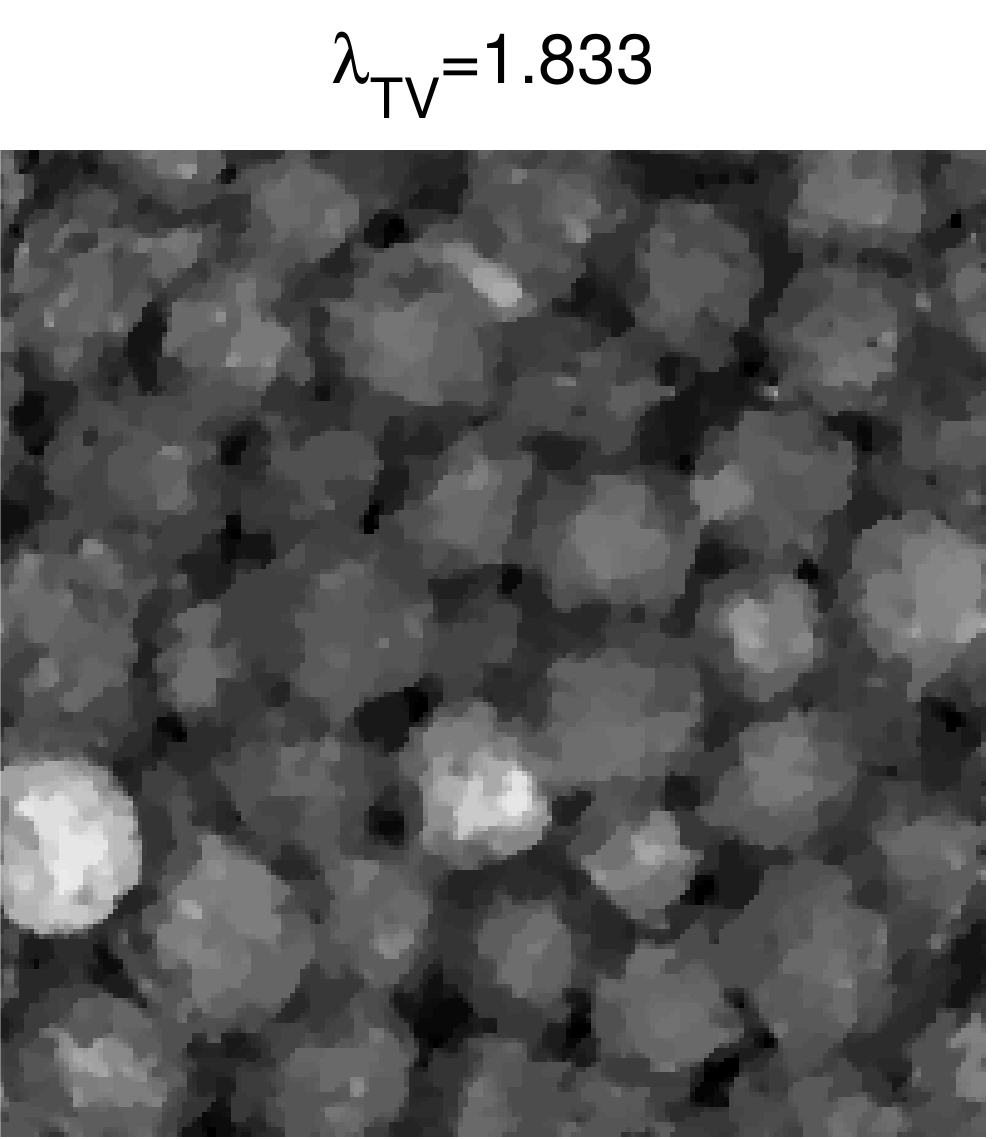}}\\
\subfigure[{$5 \times 5$, RE = 0.224}]{ \includegraphics[width=0.3\linewidth]{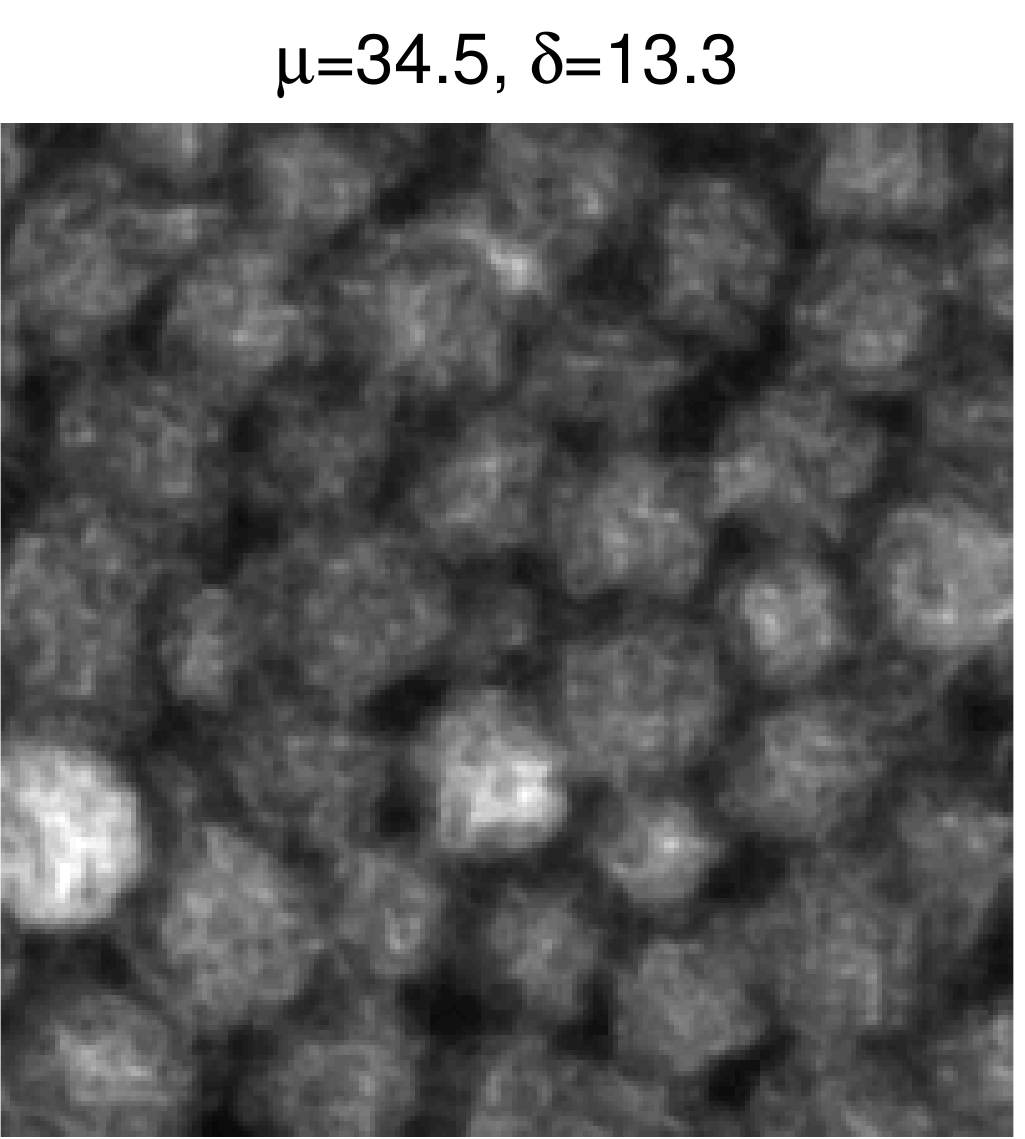}}~
\subfigure[{$10 \times 10$, RE = 0.220}]{ \includegraphics[width=0.3\linewidth]{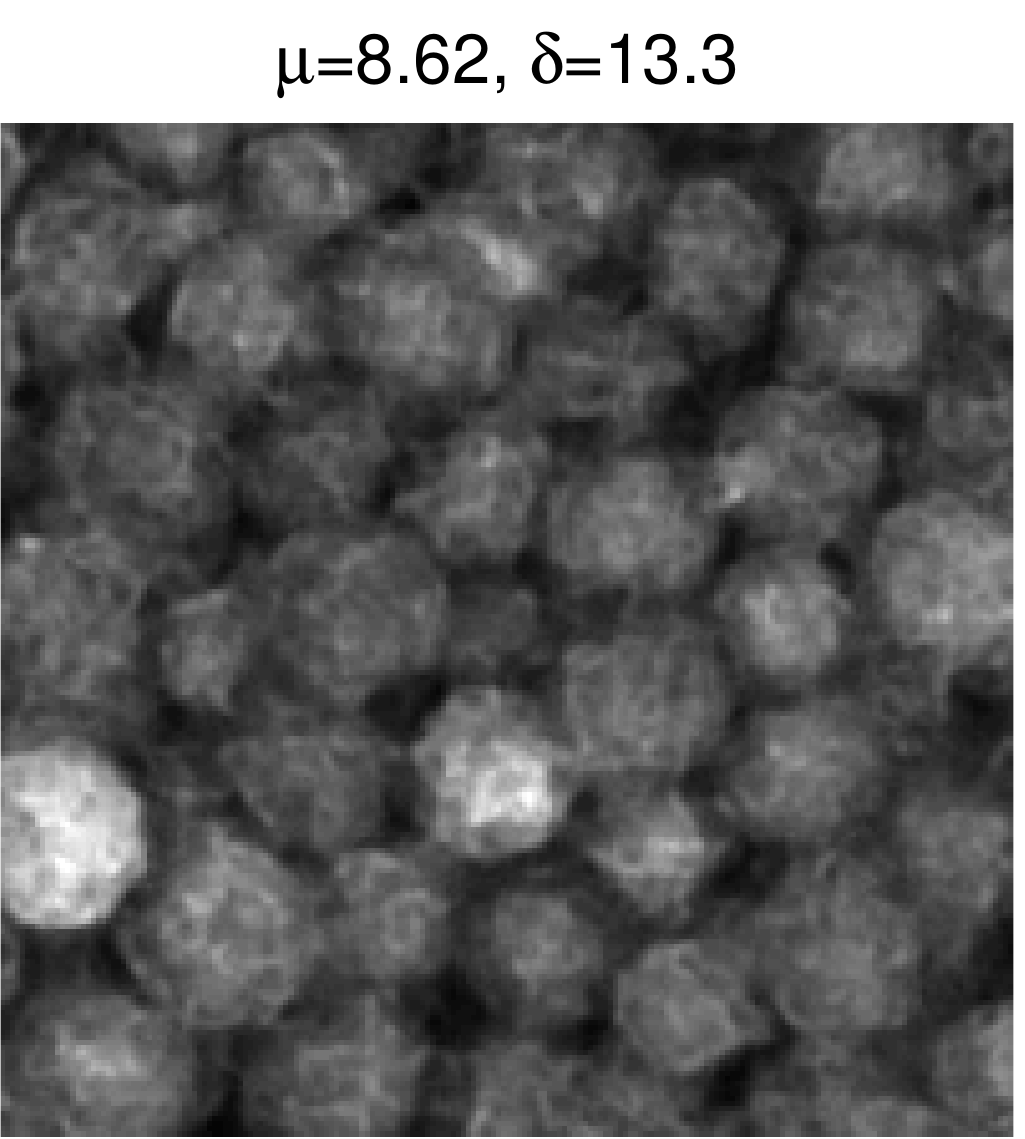}}~
\subfigure[{$20 \times 20$, RE = 0.226}]{ \includegraphics[width=0.3\linewidth]{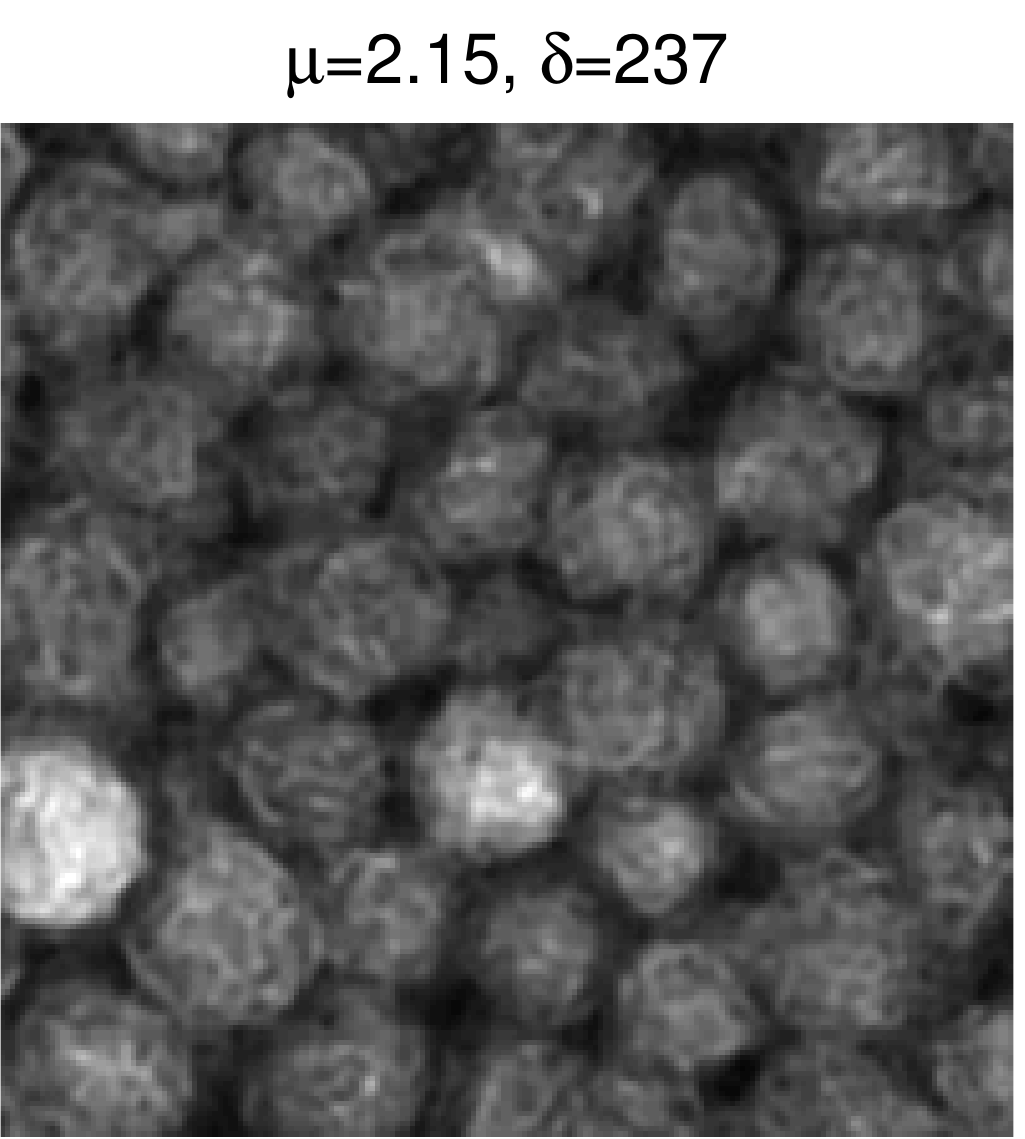}} \\
\caption{Reconstructions for different patch sizes, with $D \in \mathcal{D}_2$ and $\lambda=3.16$,
 compared with the FBP, ART and TV solutions. Note that in all our three reconstructions $\mu/q=0.022$.
 RE denotes the reconstruction error \eqref{eq:RE}.}
\label{fig:Recs}
\end{figure}

Finally, in Figure \ref{fig:Recs} we compare our reconstructions with
those computed by means of filtered back projection (FBP),
the algebraic reconstruction technique (ART), and TV regularization.
We used the Shepp-Logan filter in ``iradon.''
To be fair, the TV regularization parameter and the number of ART iterations were
chosen to yield an optimal reconstruction.
\begin{itemize}
\item
The FBP reconstruction contains the typical artifacts associated with this method for
underdetermined problems, such as line structures.
\item
The ART reconstruction\,--\,although having about the same RE as our
reconstruction\,--\,is blurry and contains artifacts such
as circle structures and errors in the corners.
\item
The TV reconstruction has the typical ``cartoonish'' appearance of TV
solutions and hence it fails to include most of the details associated
with the texture; the edges of the pepper grains are distinct but
geometrically somewhat un-smooth.
\item
Our reconstructions, while having about the same RE as the TV reconstruction,
include more texture and some of the details from the
exact image (but not all) are recovered, especially with $D^{(20)}$.
Also the pepper grain edges resemble more the smooth edges from
the exact image.
\end{itemize}
We conclude that our dictionary-based reconstruction method appears to have an
edge over the other three methods.

Our formulation in \eqref{eq:mainRec} enforces that the solution is an exact
representation in the dictionary, and searching for solutions in the cone spanned
by the dictionary elements is a strong assumption in the reconstruction formulation.
In \cite{Soltani} we investigated this requirement experimentally and showed that
relaxing the equality $\Pi x=(I\otimes D) $ does not give an
advantage, i.e., approximating a solution by $\Pi x \approx (I\otimes D) \alpha$
and minimizing $\|\Pi x-(I\otimes D) \alpha \|_2$ does not
improve the reconstruction quality, and one can compute a good
reconstruction as a conic combination of the dictionary elements.

\subsection{Simplifying the Computational Problem}
\label{sec:simple}

We have been working under the assumption that $\alpha \geq 0$ and that it is sparse.
Imposing both non-negativity and a 1-norm constraint on the representation vector $\alpha$
are strong assumptions in the reconstruction formulation.
If we drop the non-negativity constraint in the image reconstruction problem,
then \eqref{eq:mainRec} takes the form of a constrained least squares problem:
  \begin{equation}
   \label{eq:Lasso}
     \min_{\alpha} \frac{1}{2} \left\| \begin{pmatrix}
       \frac{1}{\sqrt{m}} A \,\Pi^{\mathrm{T}} (I\otimes D) \\[2mm]
       \frac{\delta}{\sqrt{\ell}} L \, \Pi^{\mathrm{T}}(I\otimes D)
     \end{pmatrix} \alpha - \begin{pmatrix} b \\ 0 \end{pmatrix} \right\|_2^2
    \qquad \text{s.t.}  \quad \|\alpha \|_1 \leq \gamma ,
\end{equation}
where $\gamma>0$.
Alternatively we can neglect the parameter $\mu$.
This is motivated by the plots in Figures \ref{fig:ErrorTauLambda}
and \ref{fig:ErrorDeltaLambda} which
suggest that for sufficiently large $\lambda$, $\delta$ and patch sizes,
the reconstruction error is almost independent of $\mu$ as long as it is small.
When $\mu=0$ \eqref{eq:mainRec} reduces to a nonnegatively
constrained least square problem:
  \begin{equation}
  \label{eq:NNLSQ}
  \min_{\alpha} \frac{1}{2}\left\|
    \begin{pmatrix} \frac{1}{\sqrt{m}} A  \, \Pi^{\mathrm{T}} (I\otimes D)\\[2mm]
    \frac{\delta}{\sqrt{\ell}} L  \, \Pi^{\mathrm{T}} (I\otimes D) \end{pmatrix}
     \alpha - \begin{pmatrix} b \\ 0 \end{pmatrix} \right\|_2^2
  \qquad \text{s.t.}  \qquad \alpha \geq 0.
  \end{equation}

\begin{figure}[ht!]%
 \centering
\includegraphics[height=0.3\linewidth]{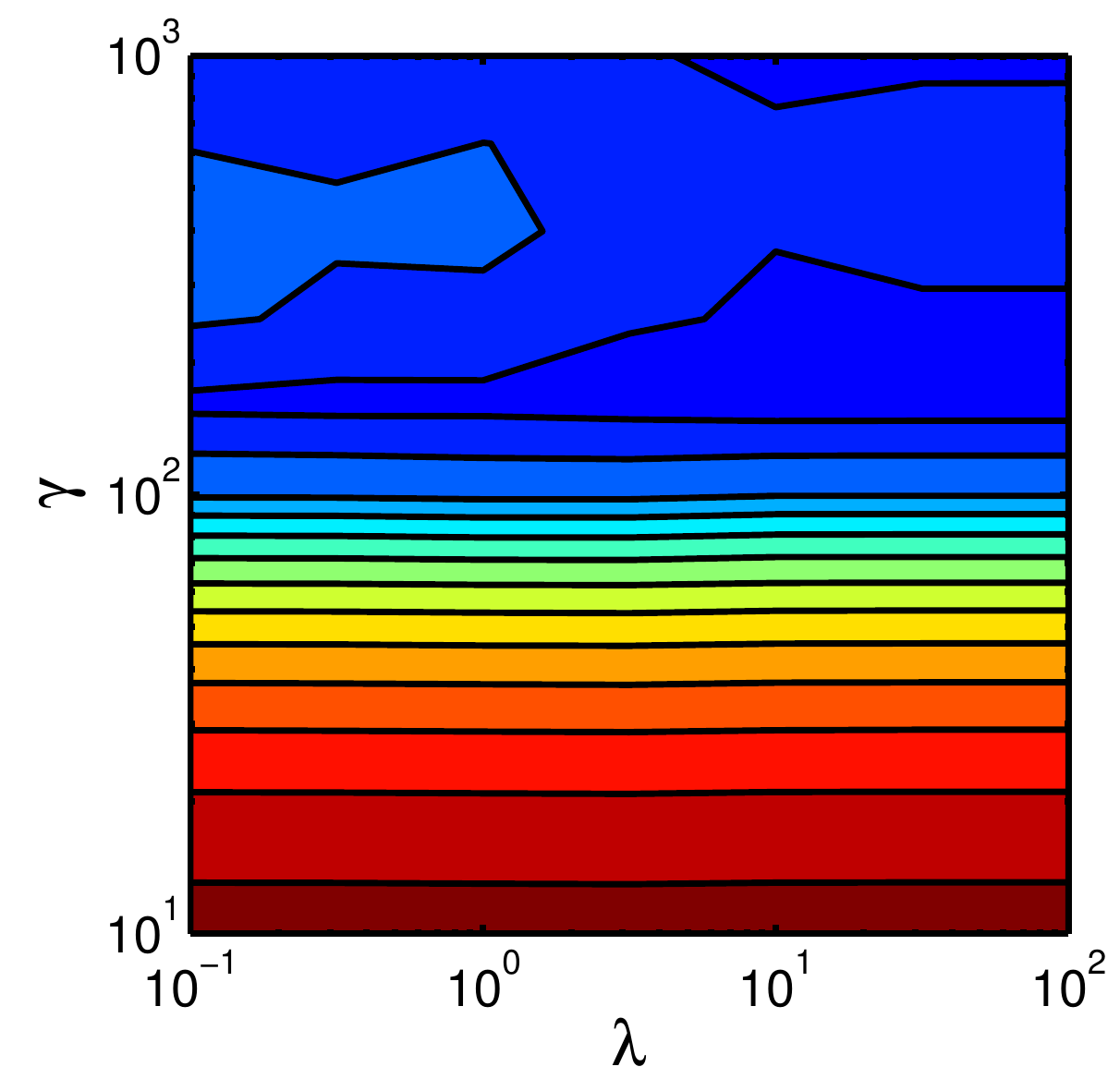}
\includegraphics[height=0.3\linewidth]{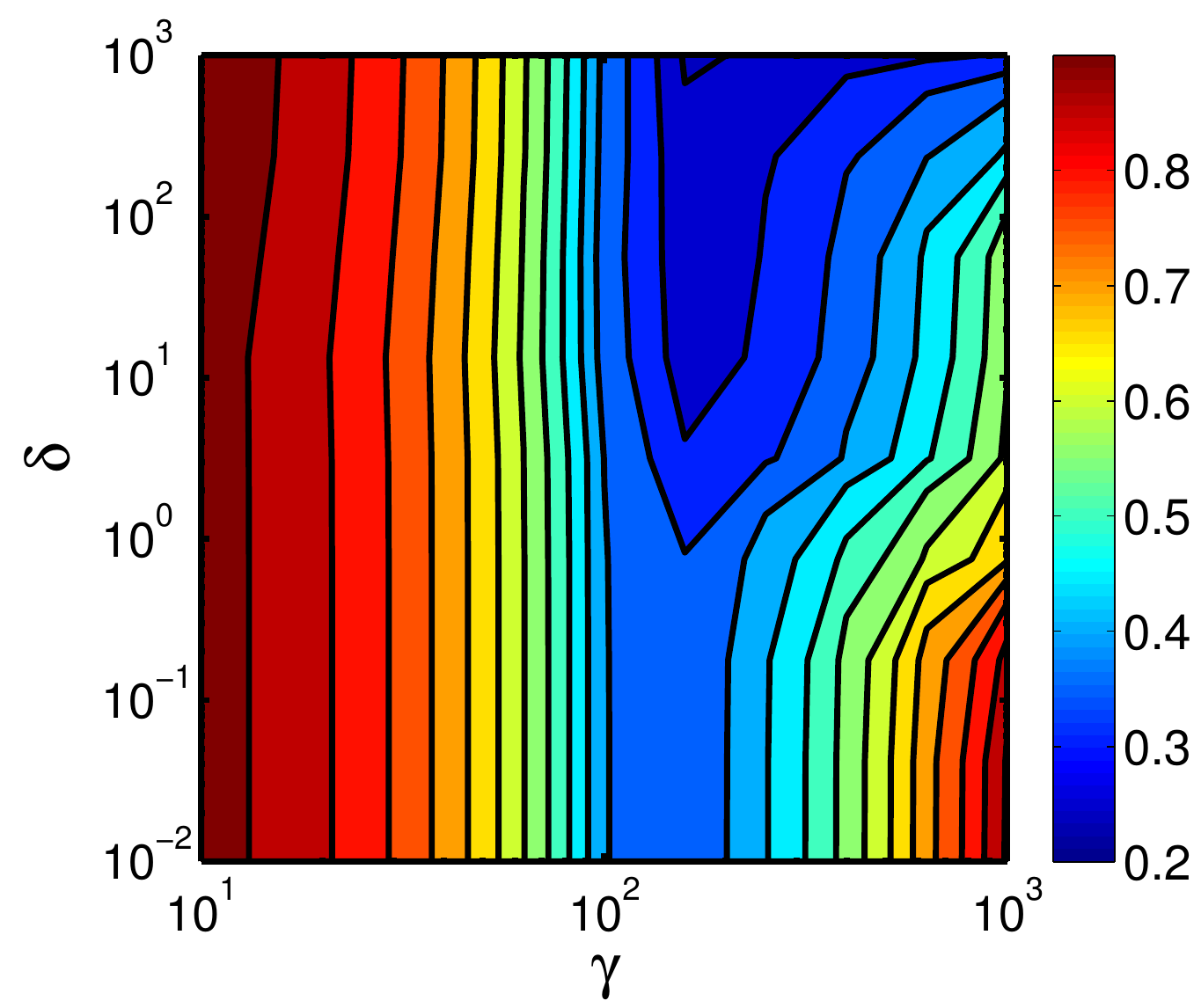}
\hspace*{2mm}
\includegraphics[trim=0mm -12mm 0mm 0mm, width=0.27\linewidth]{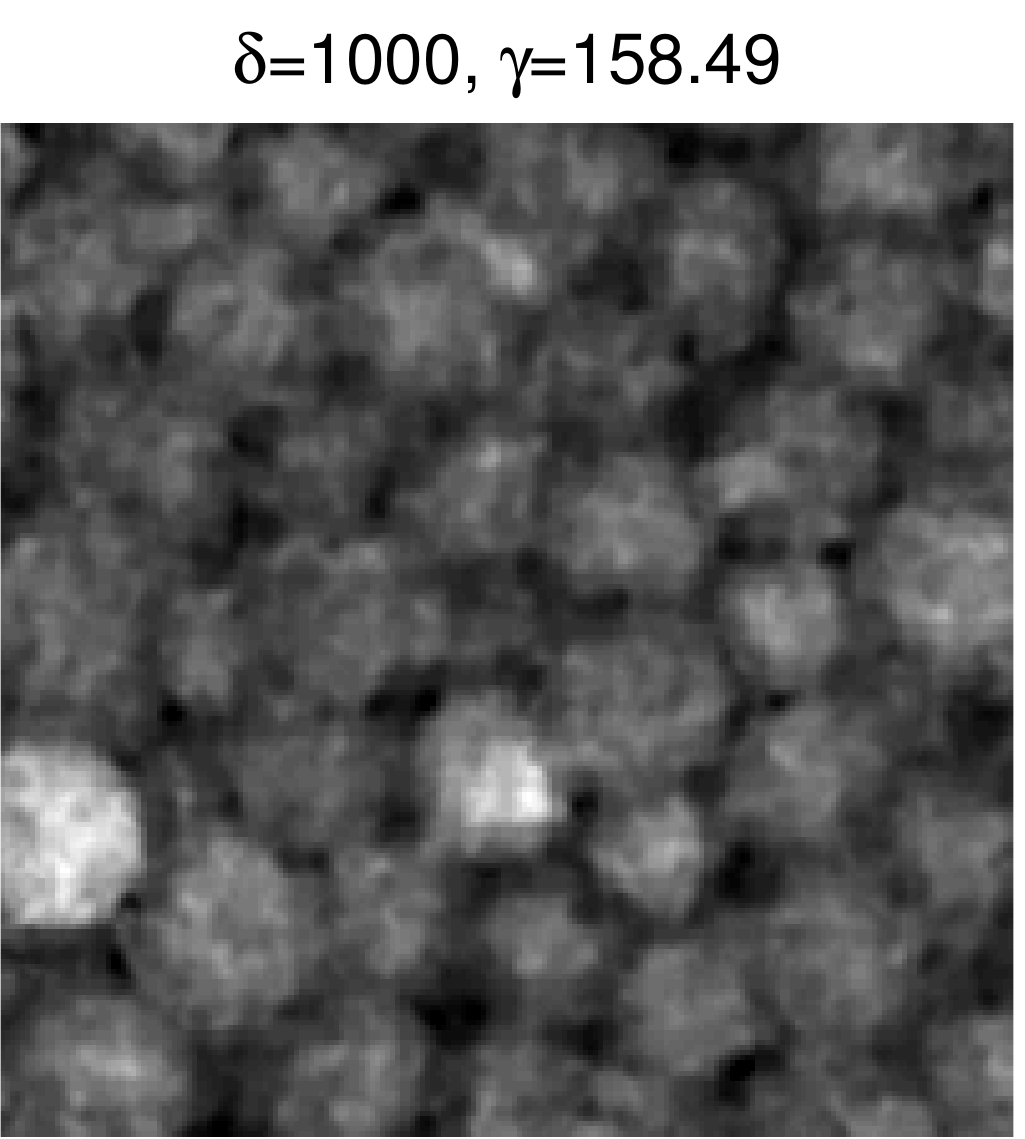}
\caption{Contour plots of the reconstruction error RE for problem \eqref{eq:Lasso},
similar to Figures \ref{fig:ErrorTauLambda} and \ref{fig:ErrorDeltaLambda}.
Left:\ RE versus $\lambda$ and $\gamma$ when $\delta=0$.
Middle:\ RE versus $\gamma$ and $\delta$ with fixed $\lambda=10$.
Right:\ The best reconstruction with $\mathrm{RE} = 0.243$.}
 \label{fig:lasso}
\end{figure}

\begin{figure}[ht!]%
 \centering
\subfigure{\includegraphics[scale=0.3]{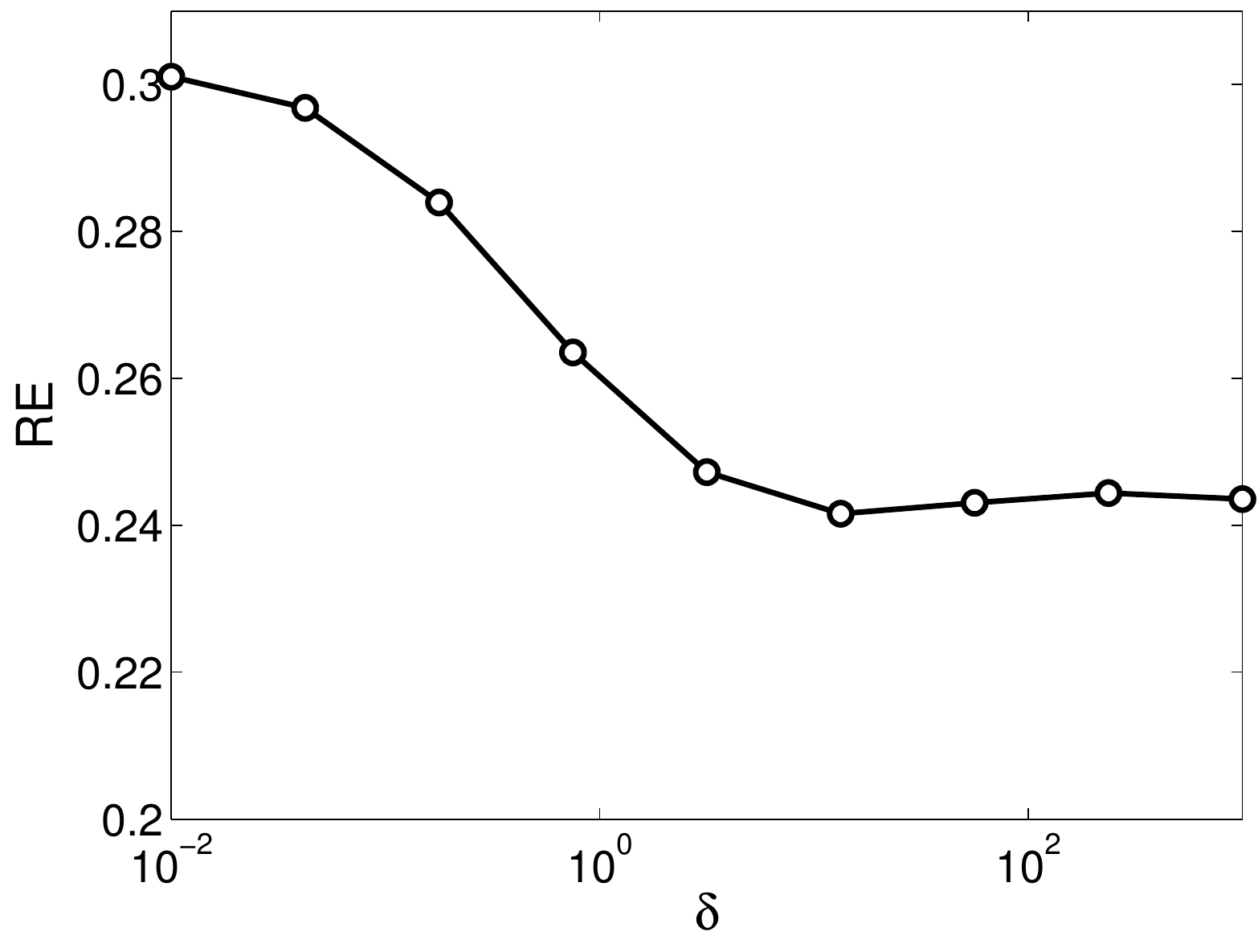}} \hspace{2mm}
\subfigure{\includegraphics[scale=0.34]{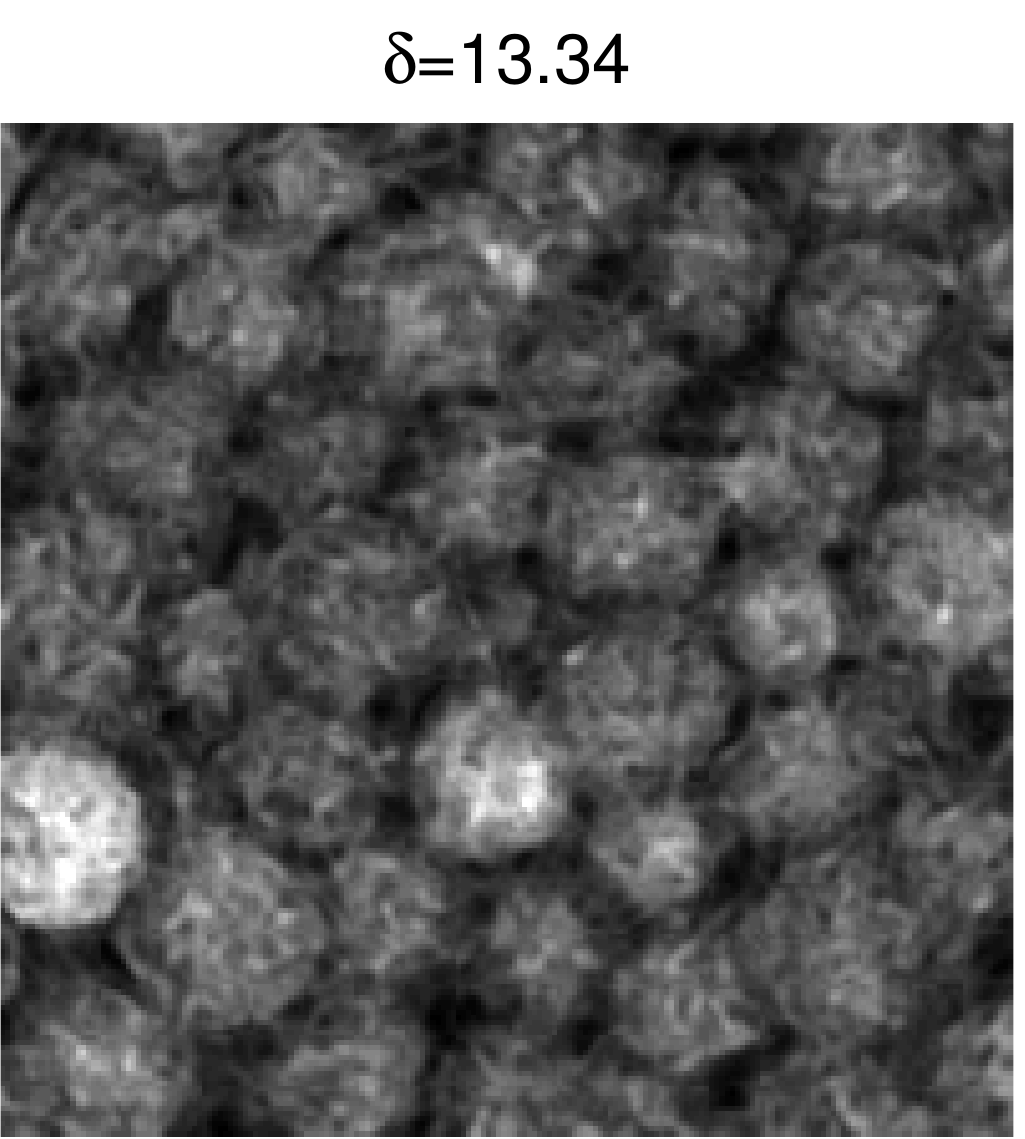}} \hspace{2mm} \\
\subfigure{\includegraphics[scale=0.3]{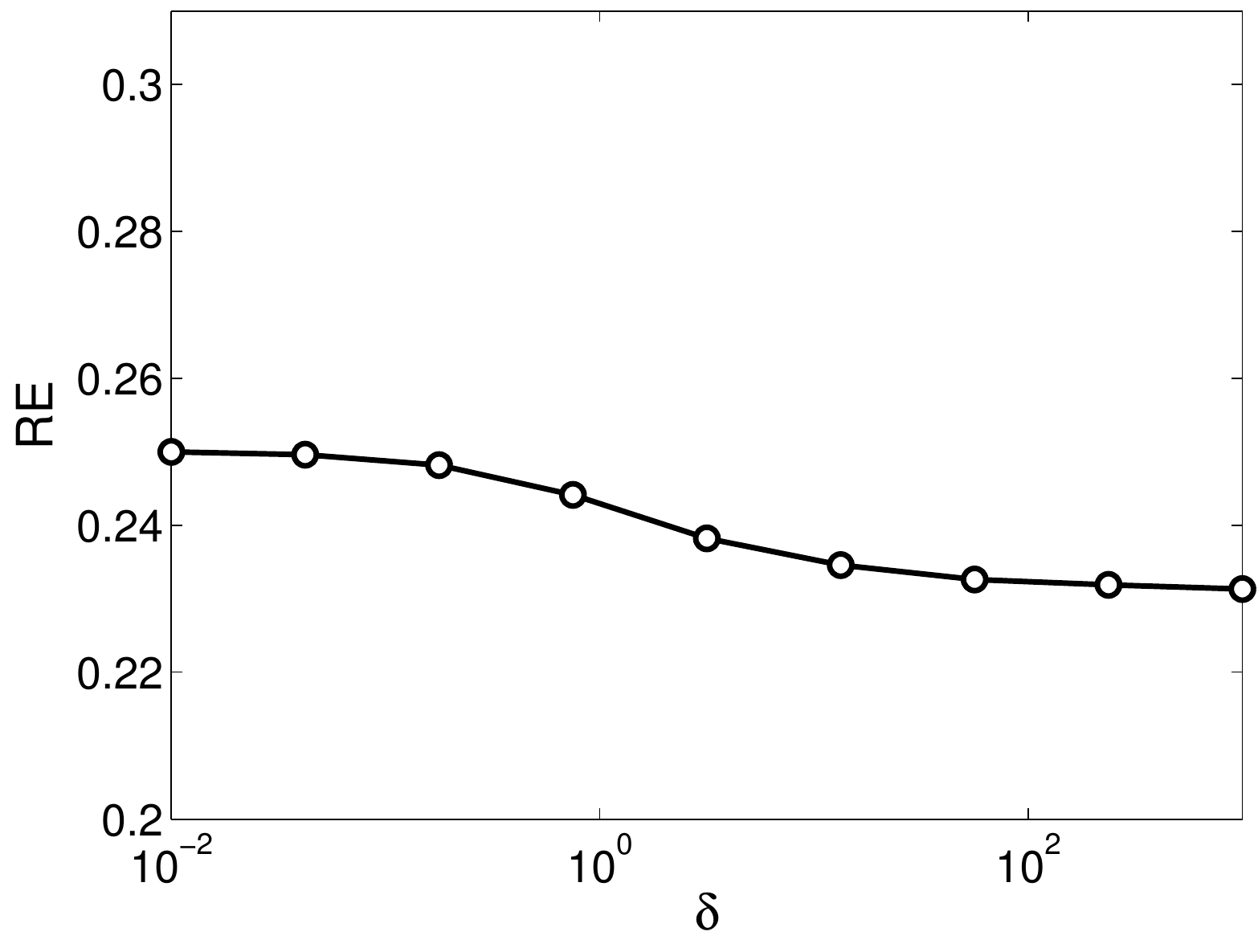}} \hspace{2mm}
\subfigure{\includegraphics[scale=0.34]{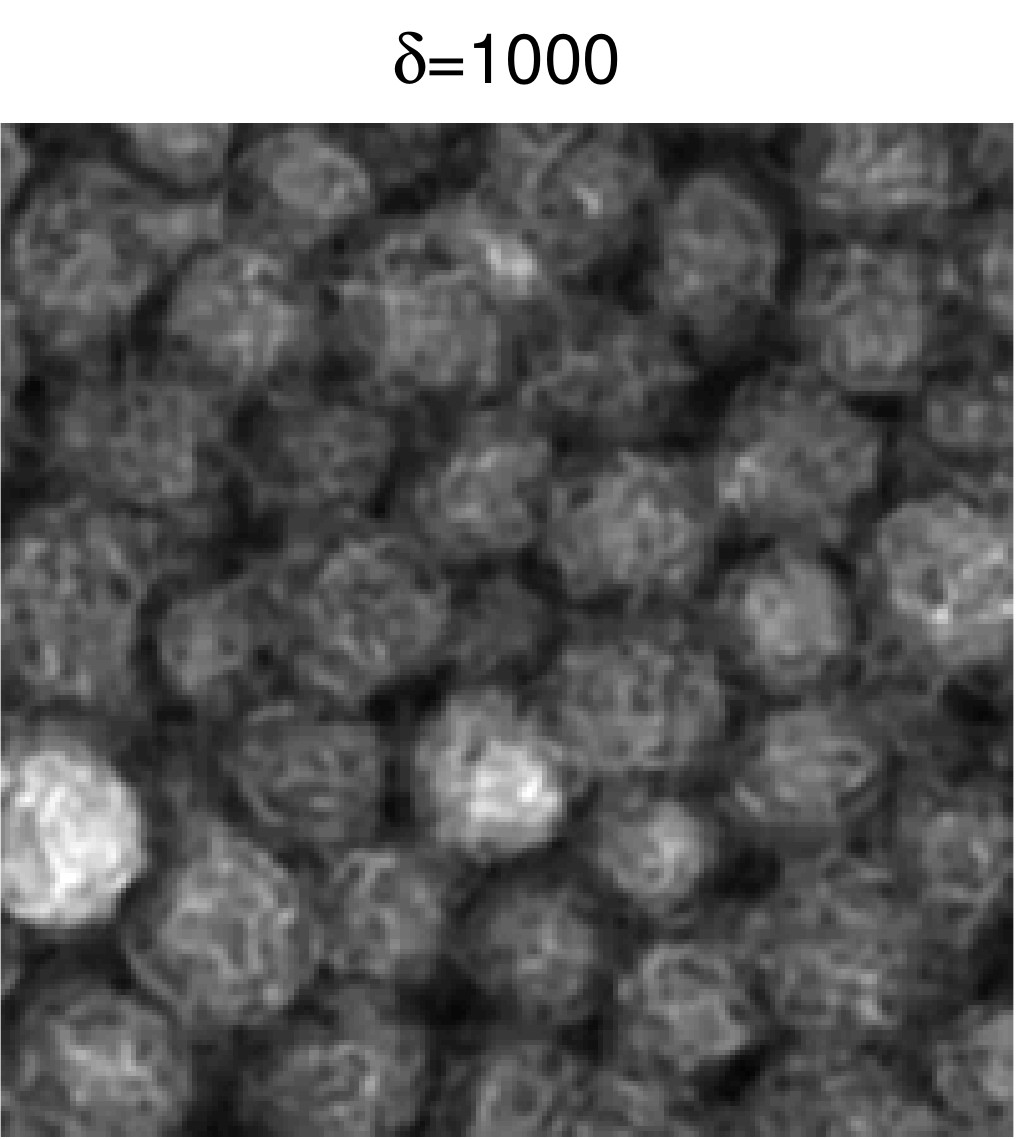}}
\caption{Plots of reconstruction error versus $\delta$ for problem \eqref{eq:NNLSQ},
  using fixed $\lambda=3.16$ and $\mu =0$, together with the best reconstructions
  with $\mathrm{RE} = 0.242$ and $\mathrm{RE} = 0.231$.
  Top and bottom correspond to patch sizes
  $10 \times 10$ and $20 \times 20$, respectively.}
\label{fig:NNLSQ}
\end{figure}

We use the same test problem with $25$ projections and relative noise level
0.01 as in Section \ref{sec:RecsTests}.
We solve problem \eqref{eq:Lasso} for $D^{(10)} \in \mathcal{D}_2 $,
which resulted in the smallest reconstruction error when solving \eqref{eq:mainRec}
(cf.\ Figure \ref{fig:Recs}).
Likewise we choose $10 \times 10$ and $20 \times 20$ patch sizes and
$D^{(10)},D^{(20)} \in \mathcal{D}_2$ to solve \eqref{eq:NNLSQ}.
Figures \ref{fig:lasso} and \ref{fig:NNLSQ} show the respective reconstructions.

There are two difficulties with the reconstructions computed via \eqref{eq:Lasso}.
The lack of a nonnegativity constraint on $\alpha$ can lead to
negative pixel values in the reconstruction, and this is undesired because it is
nonphysical and it leads to a larger reconstruction error .
Also, as can be seen in Figure~\ref{fig:lasso}, the reconstruction
is very sensitive to the choose of the regularization parameter~$\gamma$;
it must be sufficiently large to allow the solution
to be represented with a sufficient number of dictionary elements,
and it should be carefully chosen to provide an acceptable reconstruction.

The solution to problem \eqref{eq:NNLSQ} for a $20 \times 20$ patch size,
compared to the solution shown in Figure \ref{fig:Recs},
is not significantly worse both visually and in terms of reconstruction error.
This suggests that using the dictionary obtained from \eqref{e-dictionary-learn}
with a proper choice of $\lambda$ and patch size and a nonnegatively constraint
may be sufficient for the reconstruction problem,
i.e., we can let $\mu=0$.
While this seems to simplify the problem -- going from \eqref{eq:mainRec} to
\eqref{eq:NNLSQ} -- it does not significantly simplify the computational
optimization problem, since the 1-norm constraint is handled by
simple thresholding in the software;
but it help us to get rid of a parameter in the reconstruction process.
However, when the 1-norm constraint is omitted, additional care is
necessary when choosing $\lambda$ and the patch sizes to
avoid introducing artifacts or noise.

\subsection{Studies of Robustness} 
\label{sec:MoreTests}

\begin{figure}[ht!]%
 \centering
\subfigure[$D^{(10)}$, RE = 0.247]{ \includegraphics[width=0.3\linewidth]{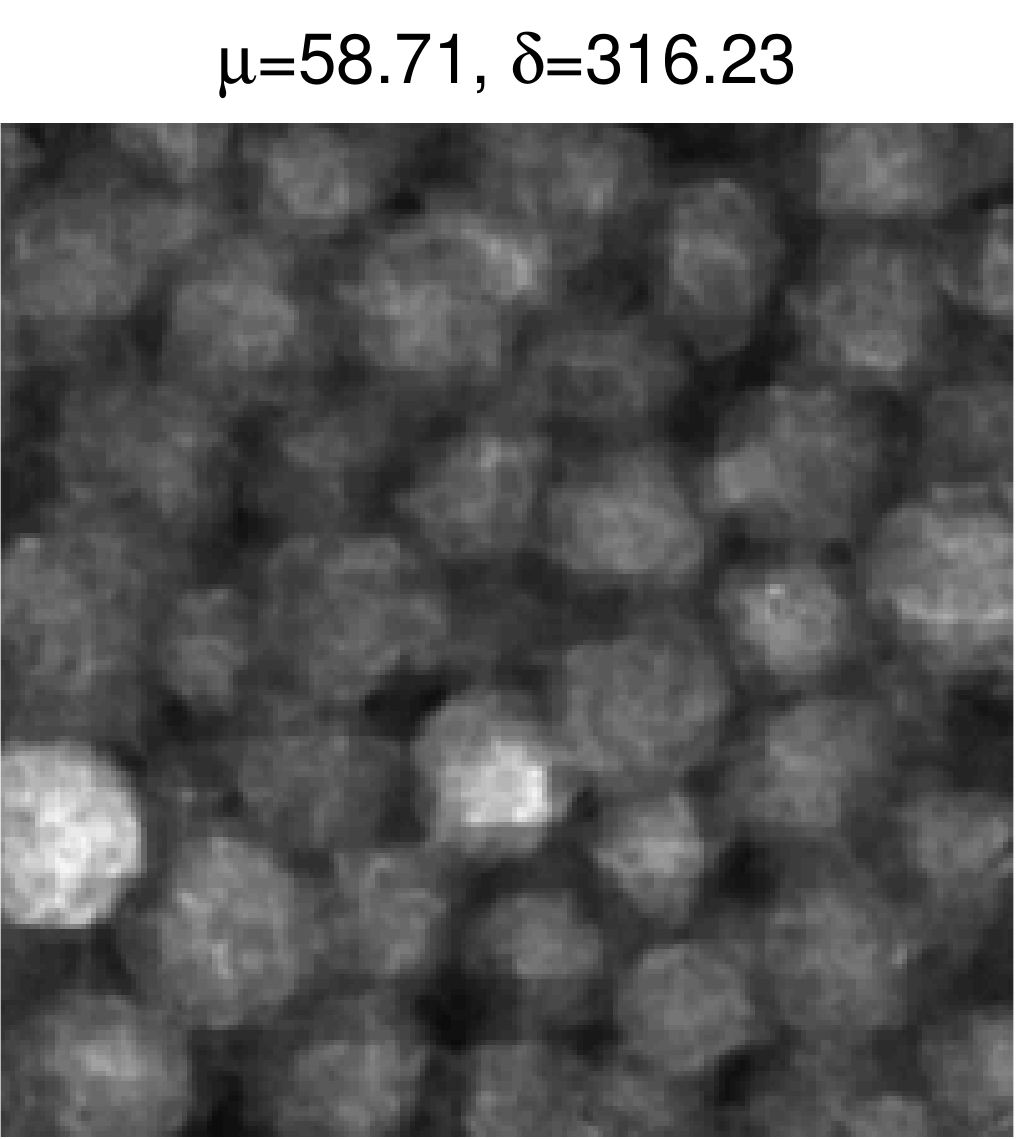}}
\subfigure[$D^{(20)}$, RE = 0.262]{ \includegraphics[width=0.3\linewidth]{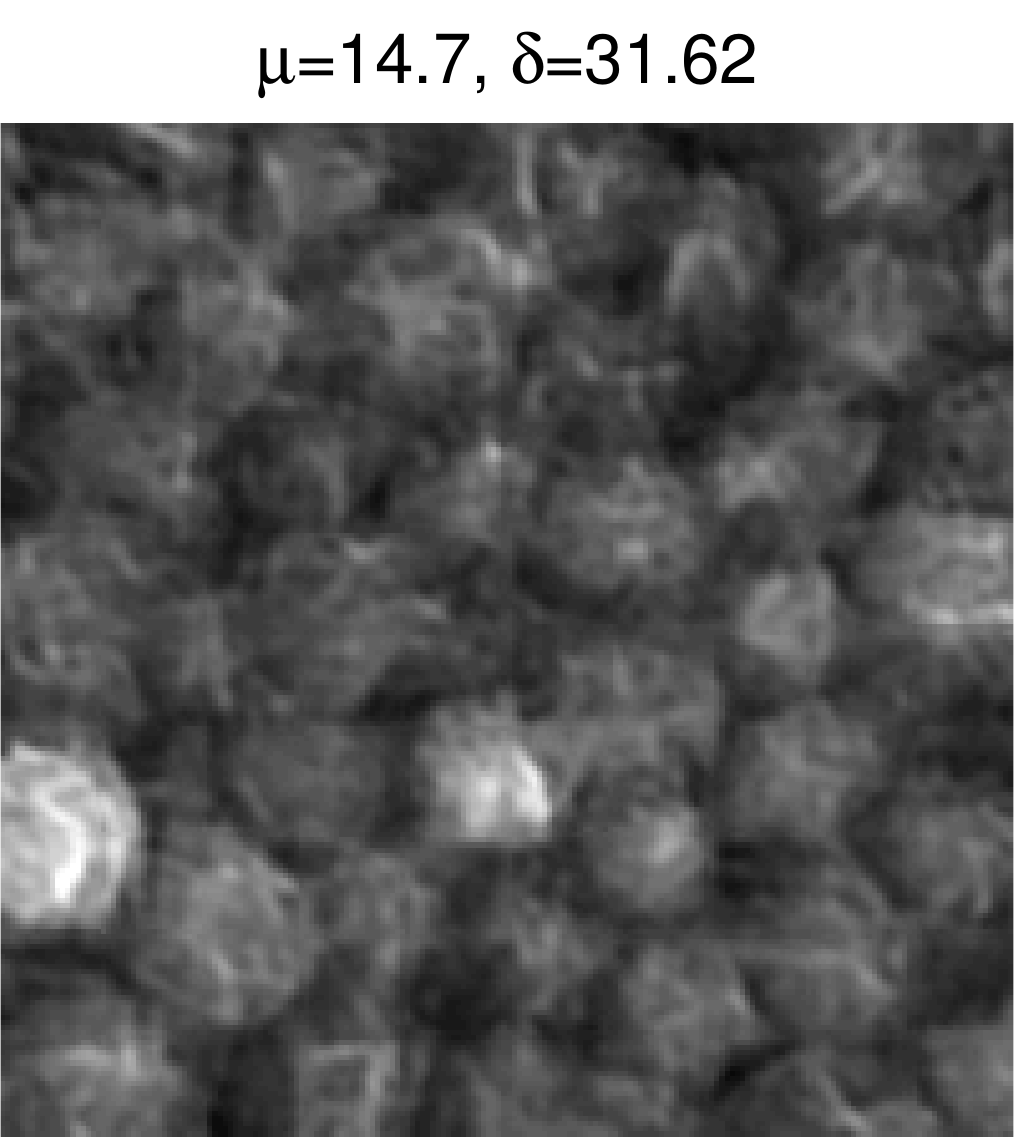}}
\subfigure[TV, RE = 0.245]{ \includegraphics[width=0.3\linewidth]{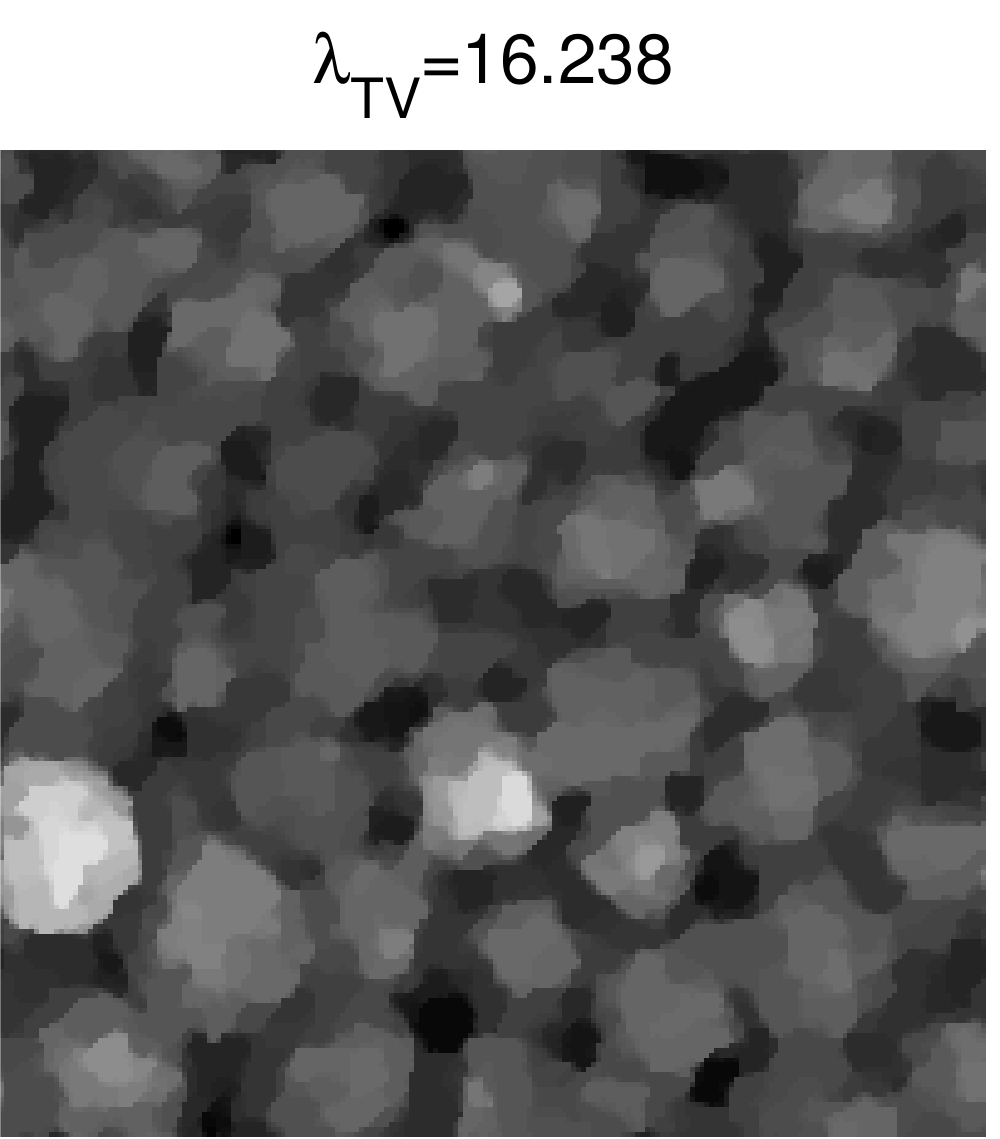}} \\
\subfigure[$D^{(10)}$, RE = 0.220]{ \includegraphics[width=0.3\linewidth]{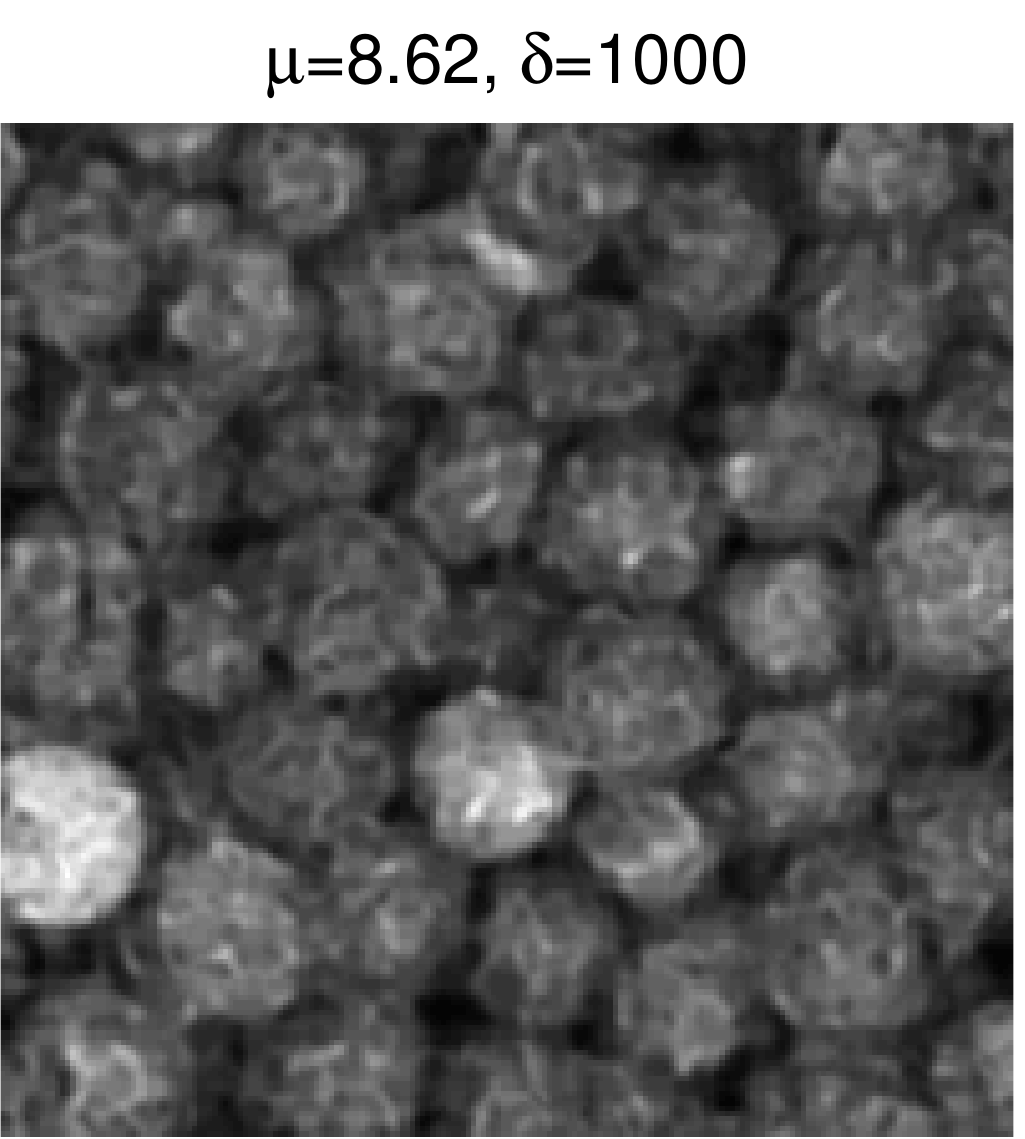}}
\subfigure[$D^{(20)}$, RE = 0.222]{ \includegraphics[width=0.3\linewidth]{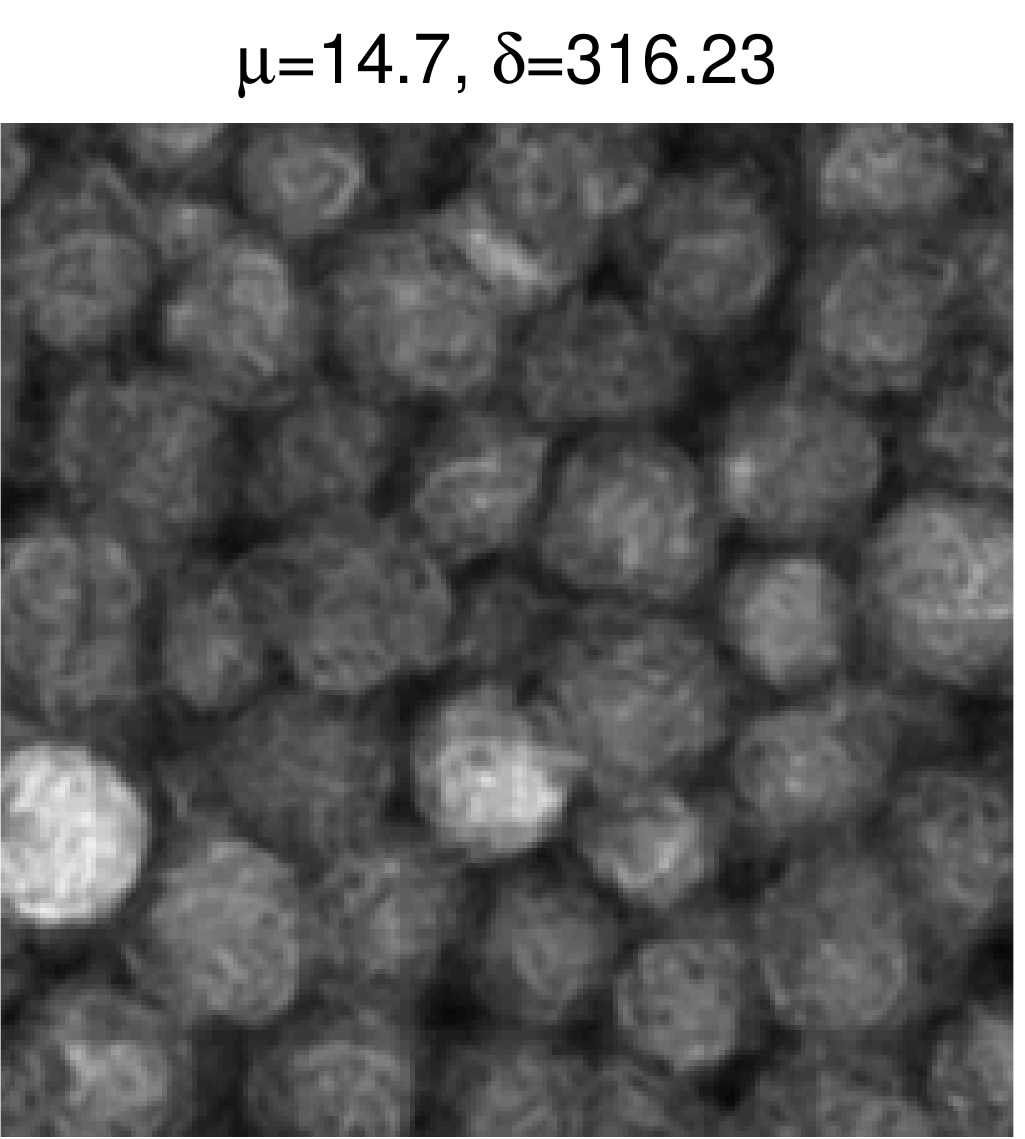}}
\subfigure[TV, RE = 0.215]{ \includegraphics[width=0.3\linewidth]{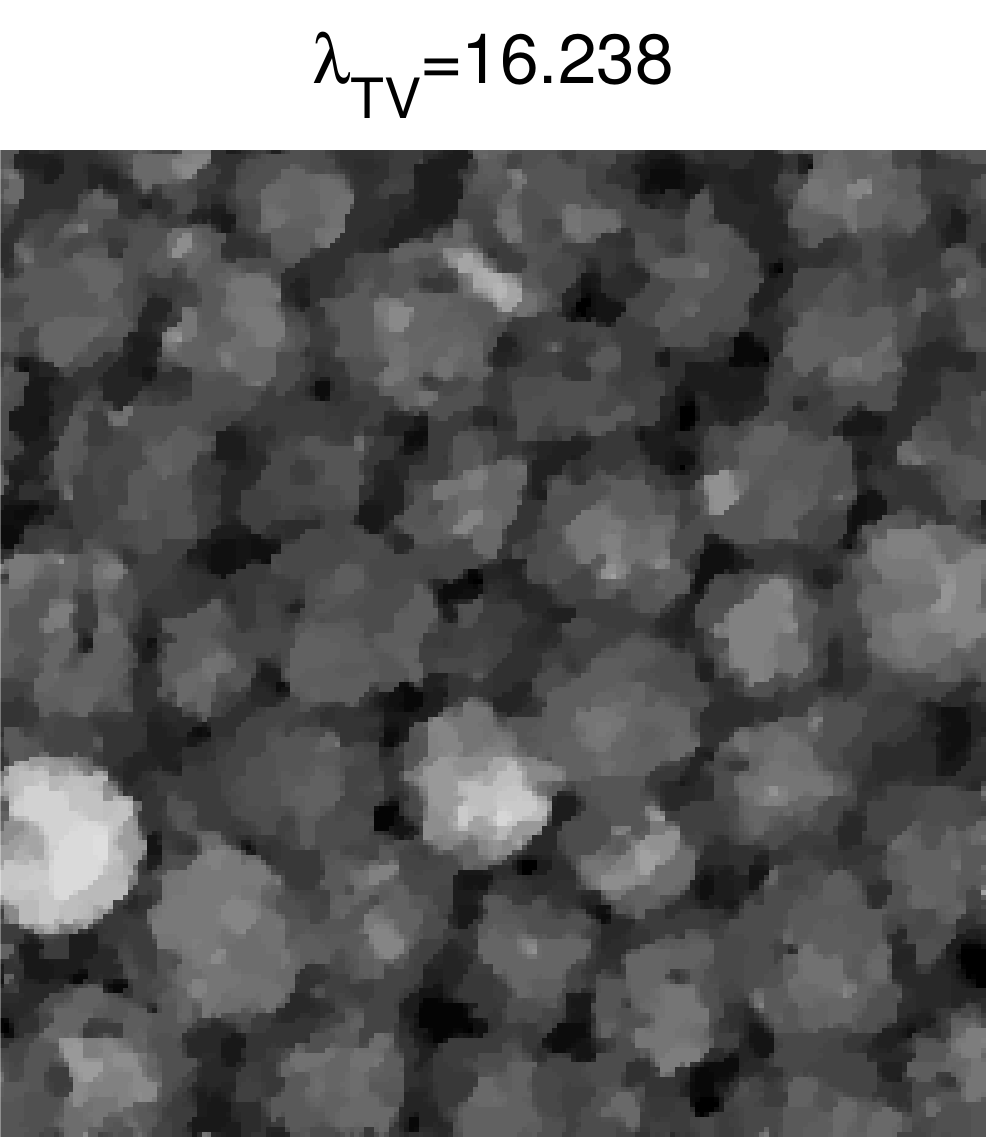}} \\
\subfigure[$D^{(10)}$, RE = 0.255]{ \includegraphics[width=0.3\linewidth]{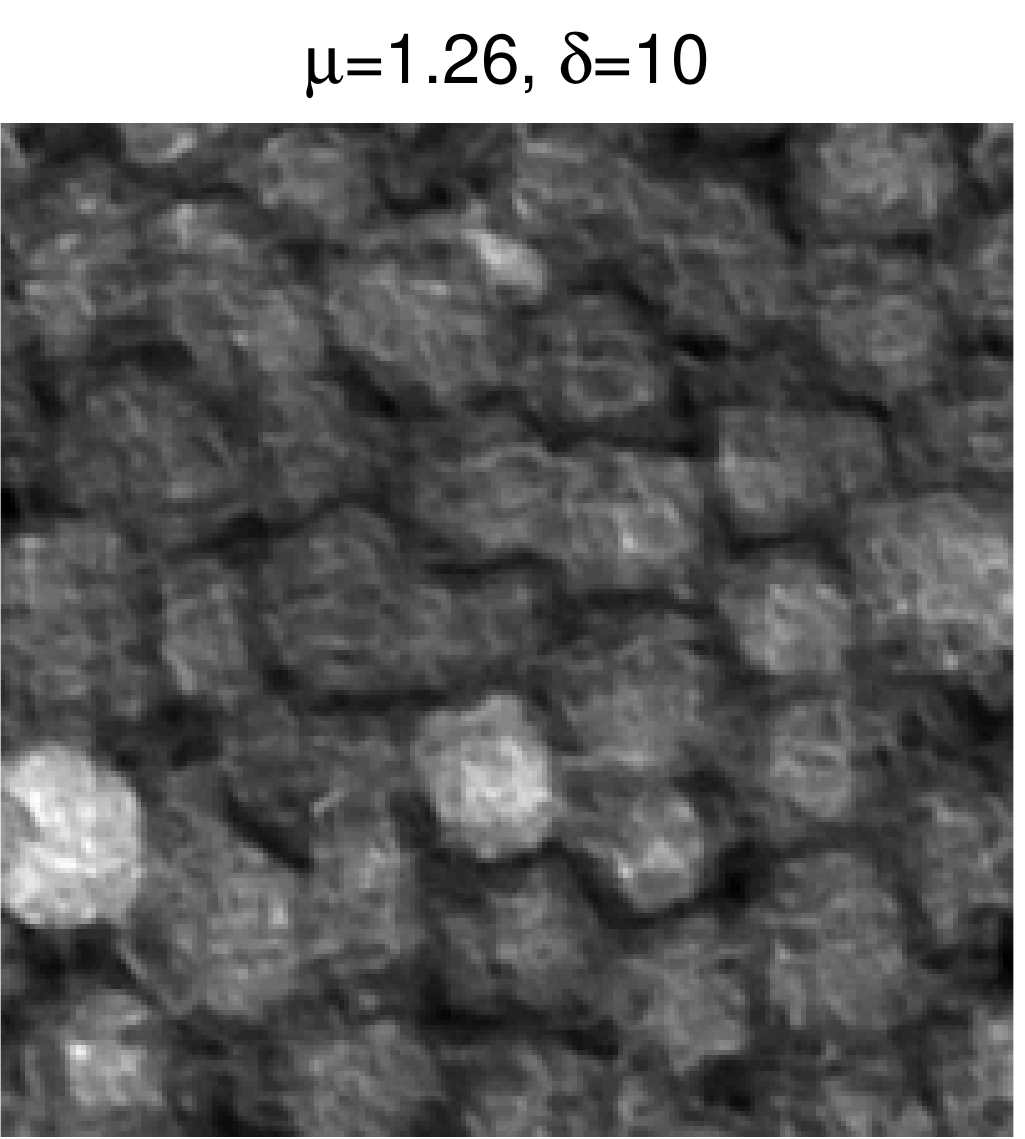}}
\subfigure[$D^{(20)}$, RE = 0.261]{ \includegraphics[width=0.3\linewidth]{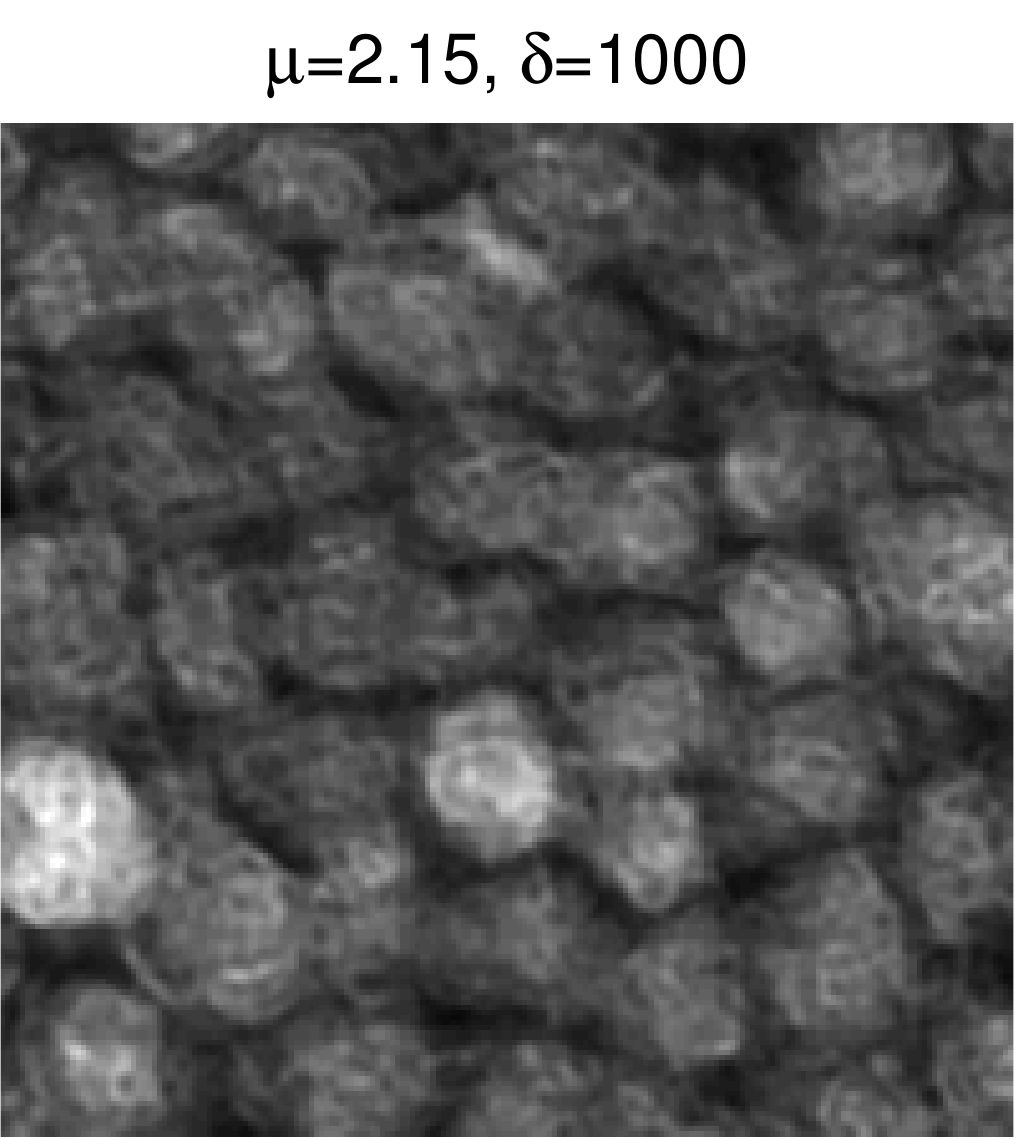}}
\subfigure[TV, RE = 0.246]{ \includegraphics[width=0.3\linewidth]{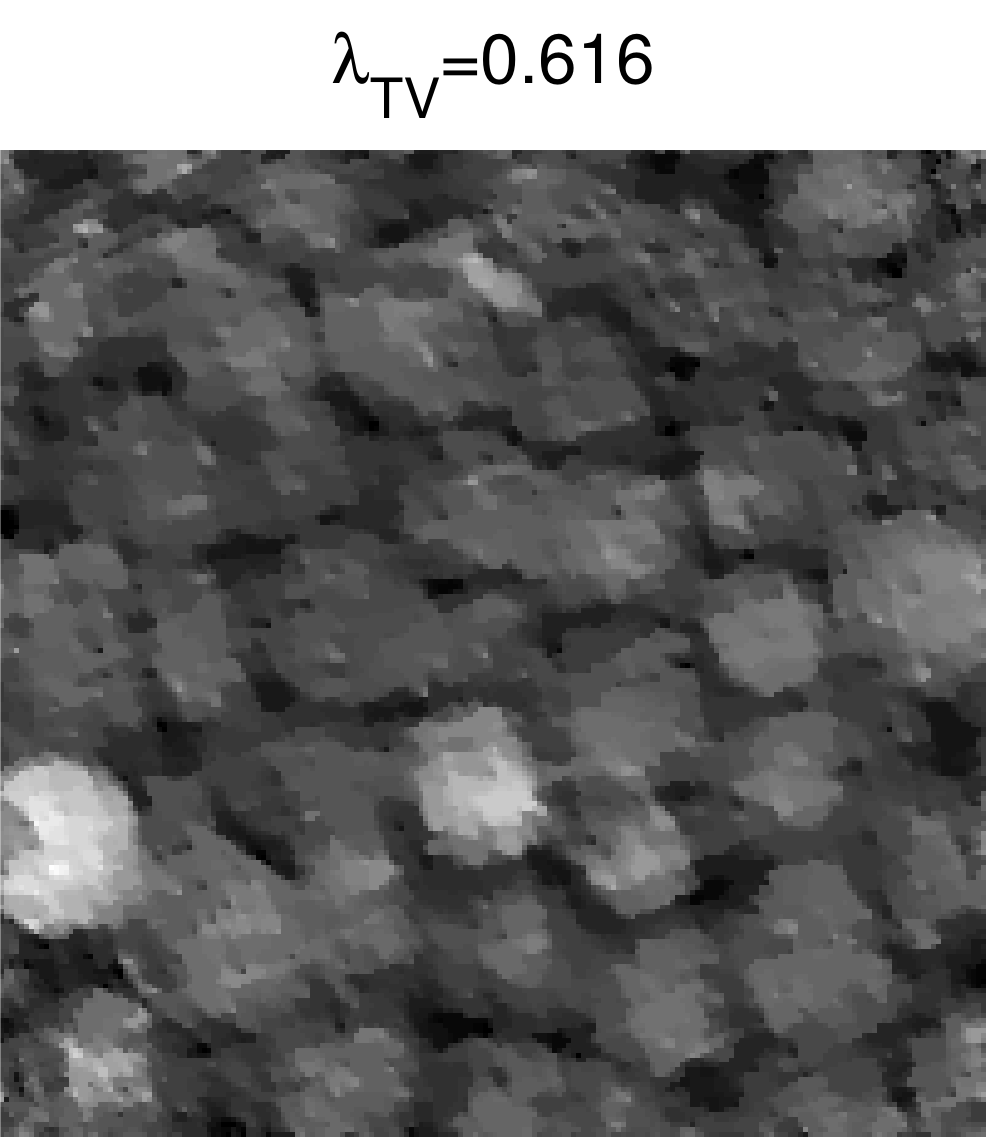}} \\
\caption{The left and middle columns show our reconstructions with $\lambda=3.16$
using $D^{(10)}$ and $D^{(20)}$, respectively;
the right column shows the TV reconstructions.
Top and middle rows:\ $N_p=25$ and $N_p=50$ projections in $[0^{\circ},180^{\circ}]$
and relative noise level 0.05.
Bottom row:\ $N_p=25$ projections in $[0^{\circ},120^{\circ}]$ and
relative noise level 0.01.}
 \label{fig:RecsSims}
\end{figure}

To further study the performance of our algorithm,
in this section we consider reconstructions based on (\ref{eq:mainRec})
with more noise in the data, and with projections within a limited range.
The first two tests use $25$ and $50$ projections
with uniform angular sampling in $[0^{\circ},180^{\circ}]$
and with relative noise level = 0.05, i.e., a higher noise level than above.
For our highly underdetermined problems we know that both filtered back projection
and algebraic iterative techniques give unsatisfactory solutions, and therefore we
only compare our method with TV\@.
As before the regularization parameters $\lambda$ and $\mu$ are chosen from
numerical experiments such that a solution with the smallest error is obtained.

The reconstructions are shown in the top and middle rows of Figure~\ref{fig:RecsSims}.
The reconstruction errors are still similar across the methods.
Again, the TV reconstructions have the characteristic ``cartoonish'' appearance
while the dictionary-based reconstructions retain more the structure and
texture but have other artifacts -- especially for $N_{\mathrm{p}} = 25$.
We also note that these artifacts are different for the two different dictionaries.

The third set uses $25$ projections uniformly distributed in the
limited range $[0^{\circ},120^{\circ}]$ and with relative noise level 0.01.
In this case the TV reconstructions display additional artifacts related to the
limited-angle situation, while such artifacts are somewhat less pronounced in the
reconstructions by our algorithm.

In the numerical studies performed in this paper there is an underlying assumption
that the scale and orientation of the training images are consistent with the unknown
image.
While this assumption is convenient for the studies performed here,
it may not be entirely realistic.
In a separate work \cite{Soltani} we therefore investigated the sensitivity and
robustness of the reconstruction to variations of the scale and orientation
in the training images, and we discuss algorithms to estimate the correct
relative scale and orientation from the data (scale being the more
difficult parameter to estimate).

\subsection{A Large Test Case}
\label{sec:large}

\begin{figure}[ht!]%
 \centering
\subfigure{ \includegraphics[height=0.3\linewidth]{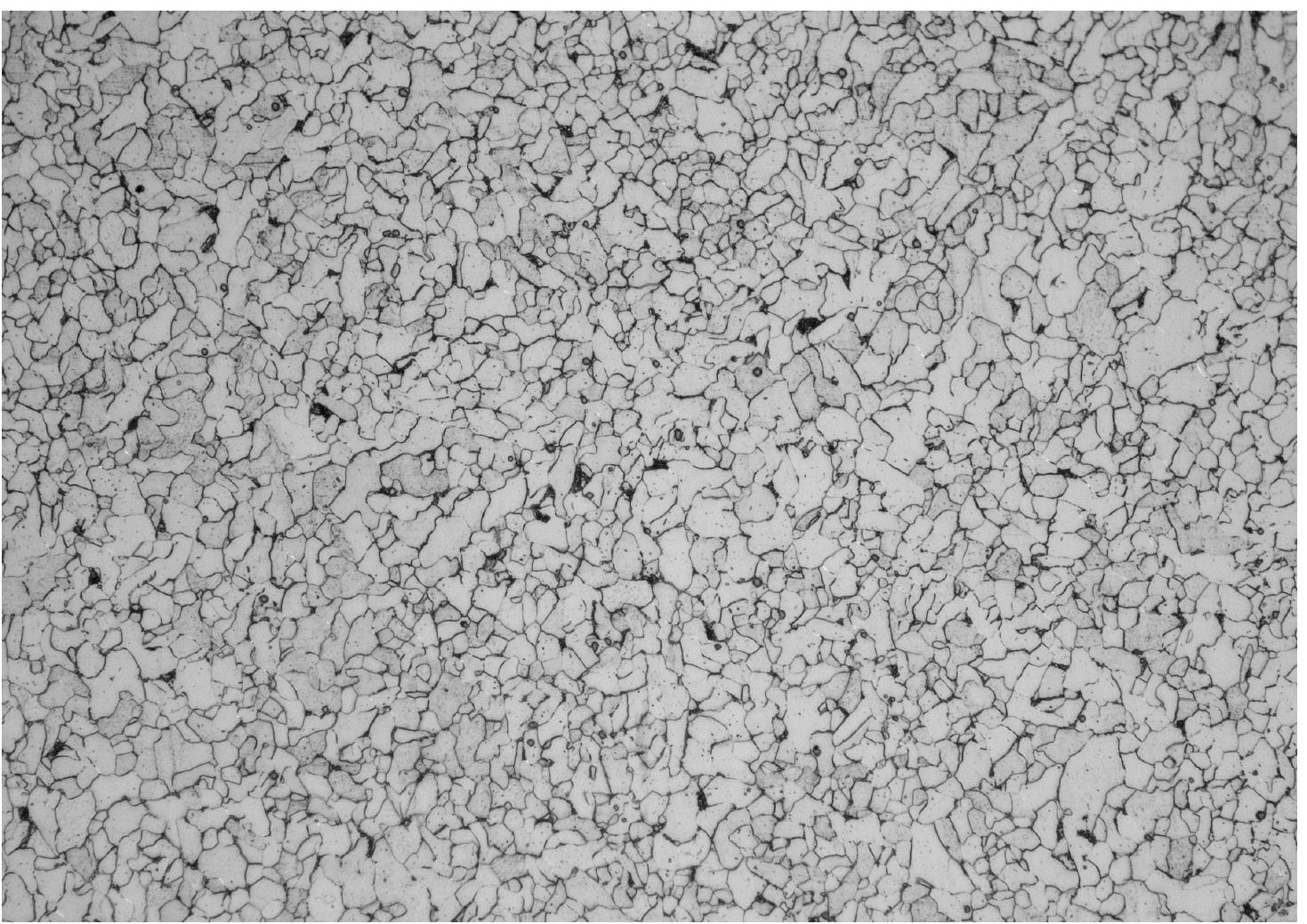}}~
\subfigure{ \includegraphics[height=0.3\linewidth]{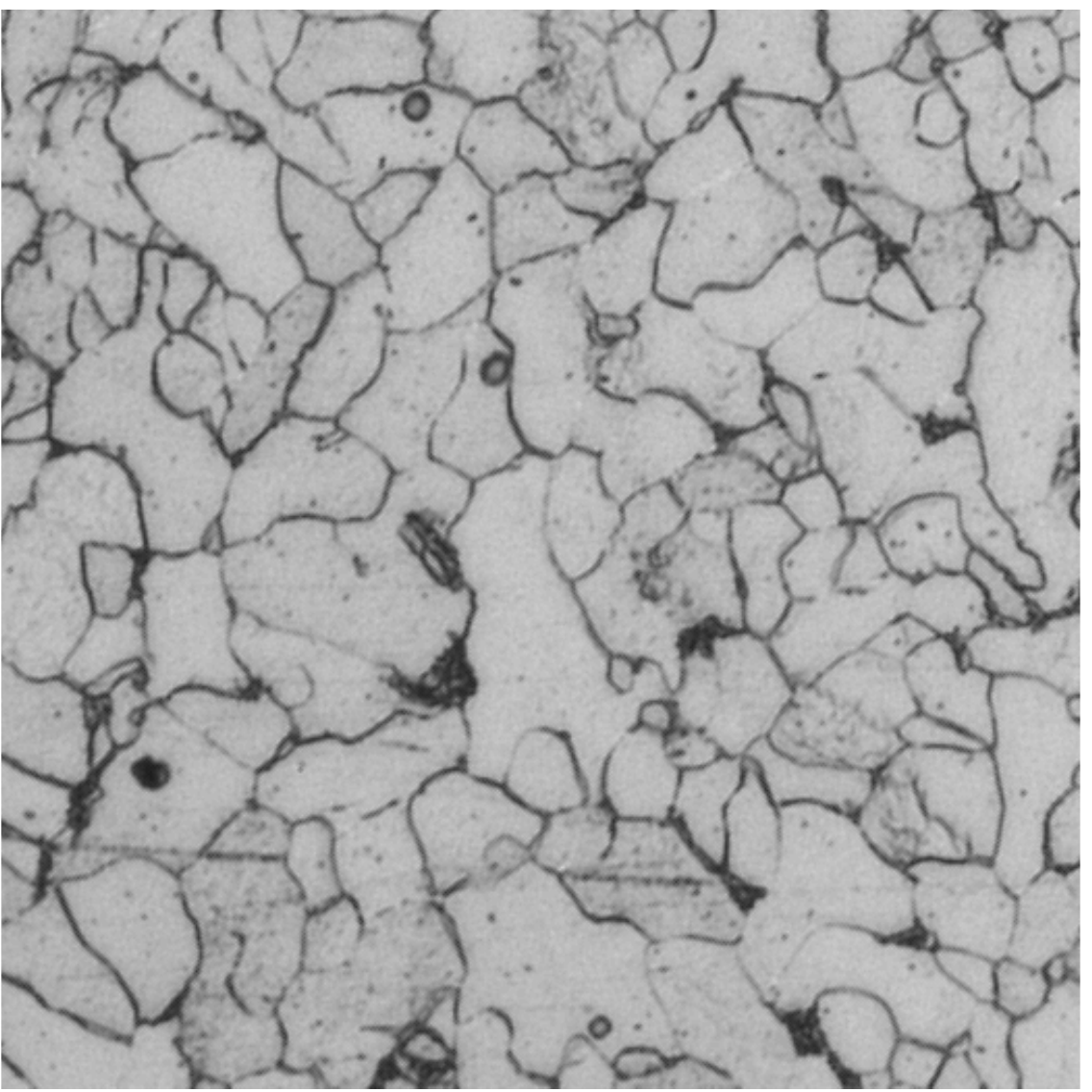}} \\
\subfigure{ \includegraphics[height=0.3\linewidth]{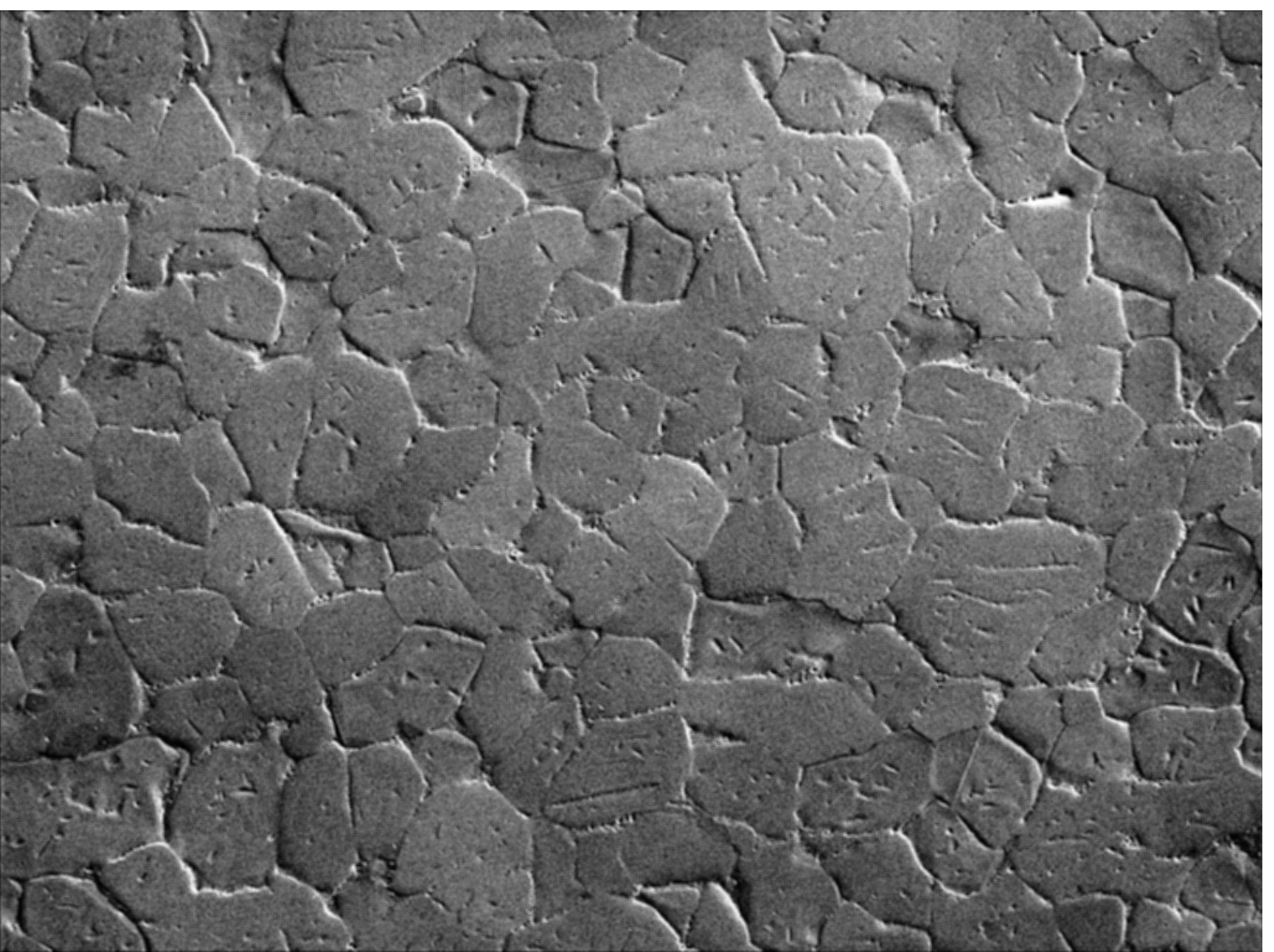}}~
\subfigure{ \includegraphics[height=0.3\linewidth]{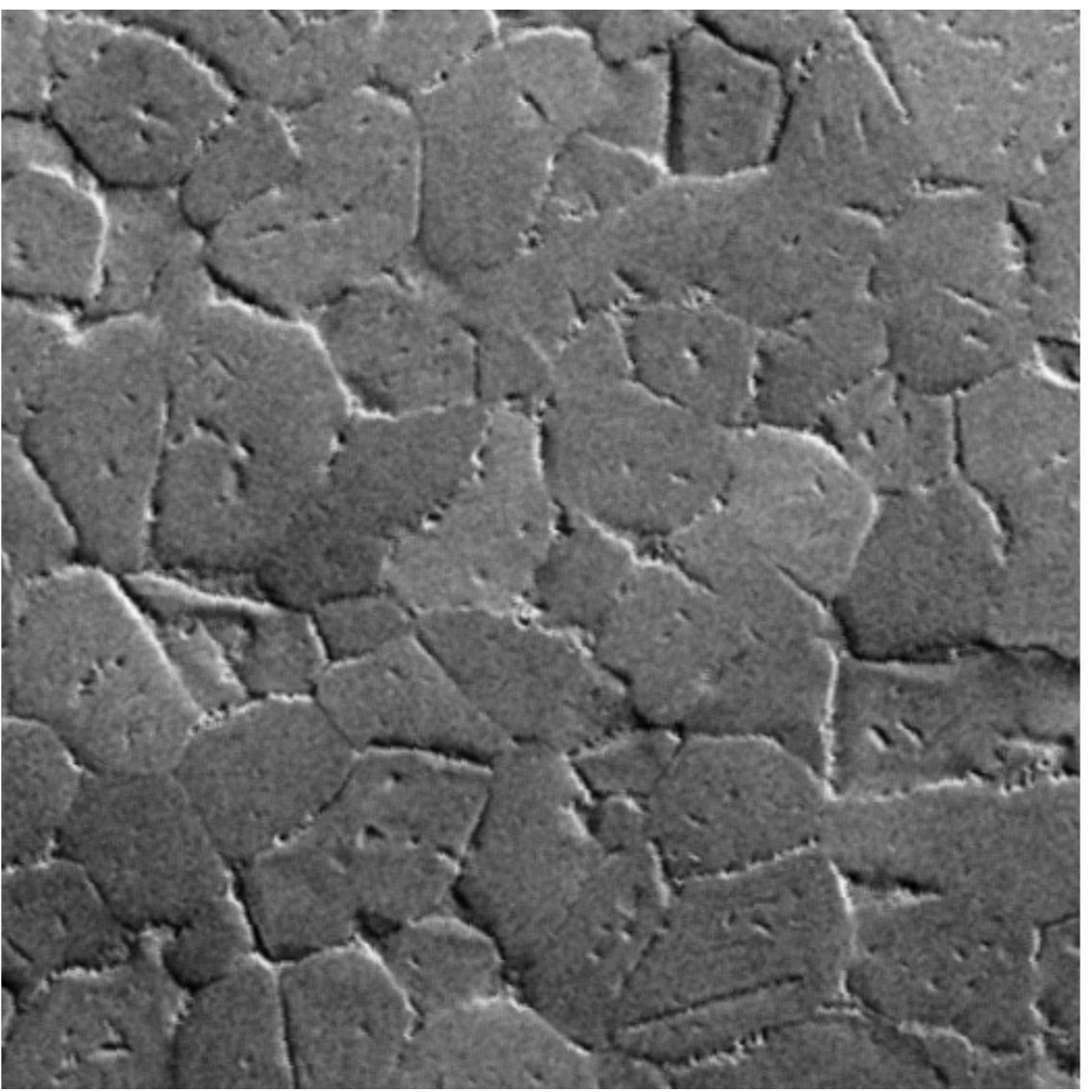}} \\
\caption{Left:\ high-resolution images of steel micro-structure \cite{steel} (top)
 and zirconium grains (bottom) used to generate the training images.
Right:\ the corresponding exact images of size $520 \times 520$.}
 \label{fig:TITrILarge}
\end{figure}

We finish the numerical experiments with a verification of our method
on two larger test problems that simulate the analysis of
microstructure in materials science.
Almost all common metals and many ceramics are polycrystalline, i.e.,
they are composed of many small crystals or grains of varying size and orientation,
and the variations in orientation can be random.
It is of particular interest to study how the grain boundaries\,---\,the
interfaces between grains\,---\,respond external stimuli such heat, stress or strain.
Here we assume that priors of the grain structure are available in the form
of training images.

The simulated data was computed using images of steel and zirconium grains.
The steel microstructure image from \cite{steel} is of dimensions $900 \times 1280$
and the zirconium grain image (produced by a scanning electron microscope)
is $760 \times 1020$.
More than $50,000$ patches are extracted from these images to learn dictionaries
$D^{(20)} \in \mathcal{D}_2,\mathcal{D}_{\infty}$ of size $400 \times 800$.
To avoid doing inverse crime, we obtain the
exact images of dimensions $520 \times 520$ by first rotating the high-resolution image
and then extracting the exact image.
The high-resolution images and the exact images are shown in Figure \ref{fig:TITrILarge}.

\begin{figure}[ht!]%
 \centering
\subfigure[$D^{20} \in \mathcal{D}_2$, RE = 0.095]{ \includegraphics[width=0.3\linewidth]{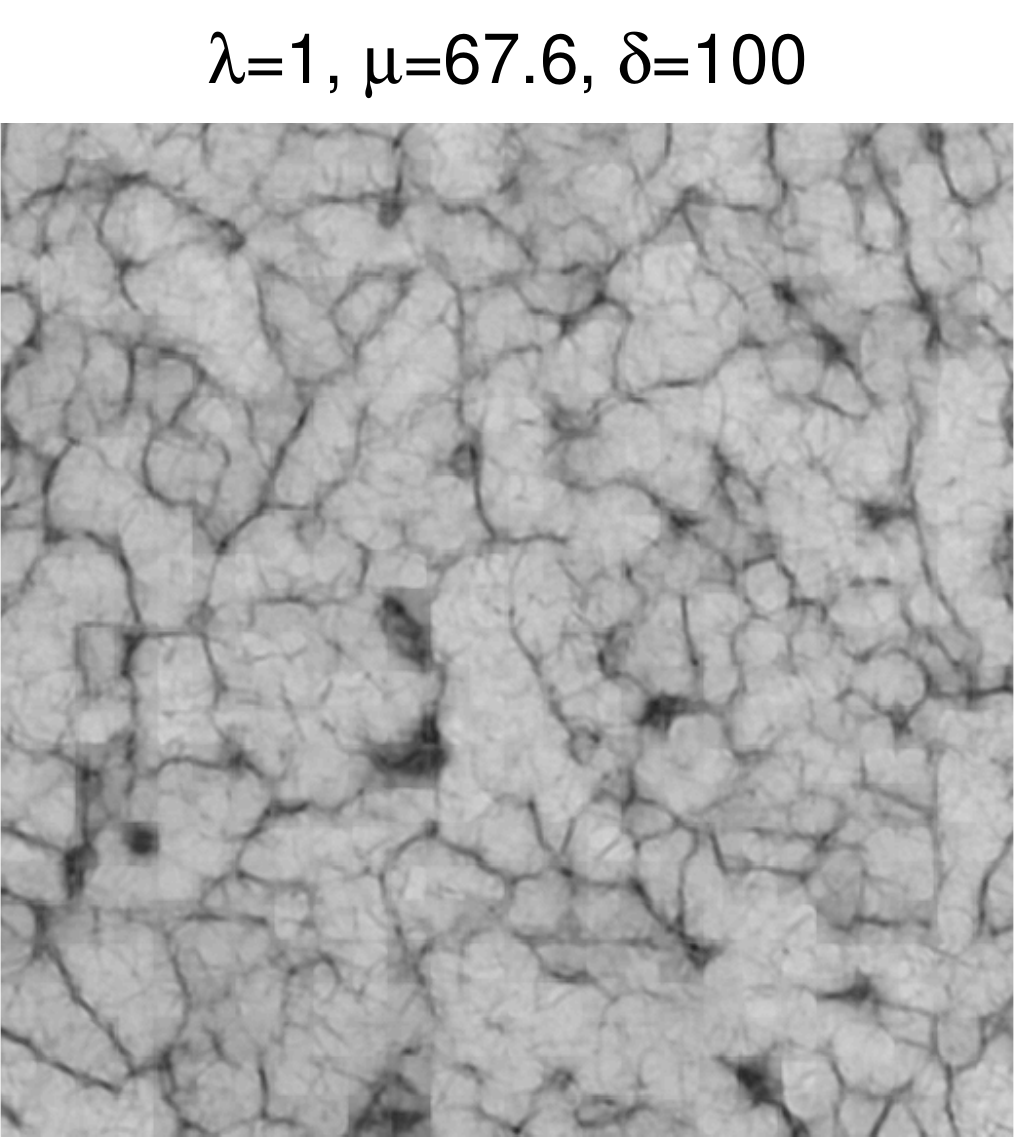}}
\subfigure[$D^{20} \in \mathcal{D}_{\infty}$, RE = 0.096]{ \includegraphics[width=0.3\linewidth]{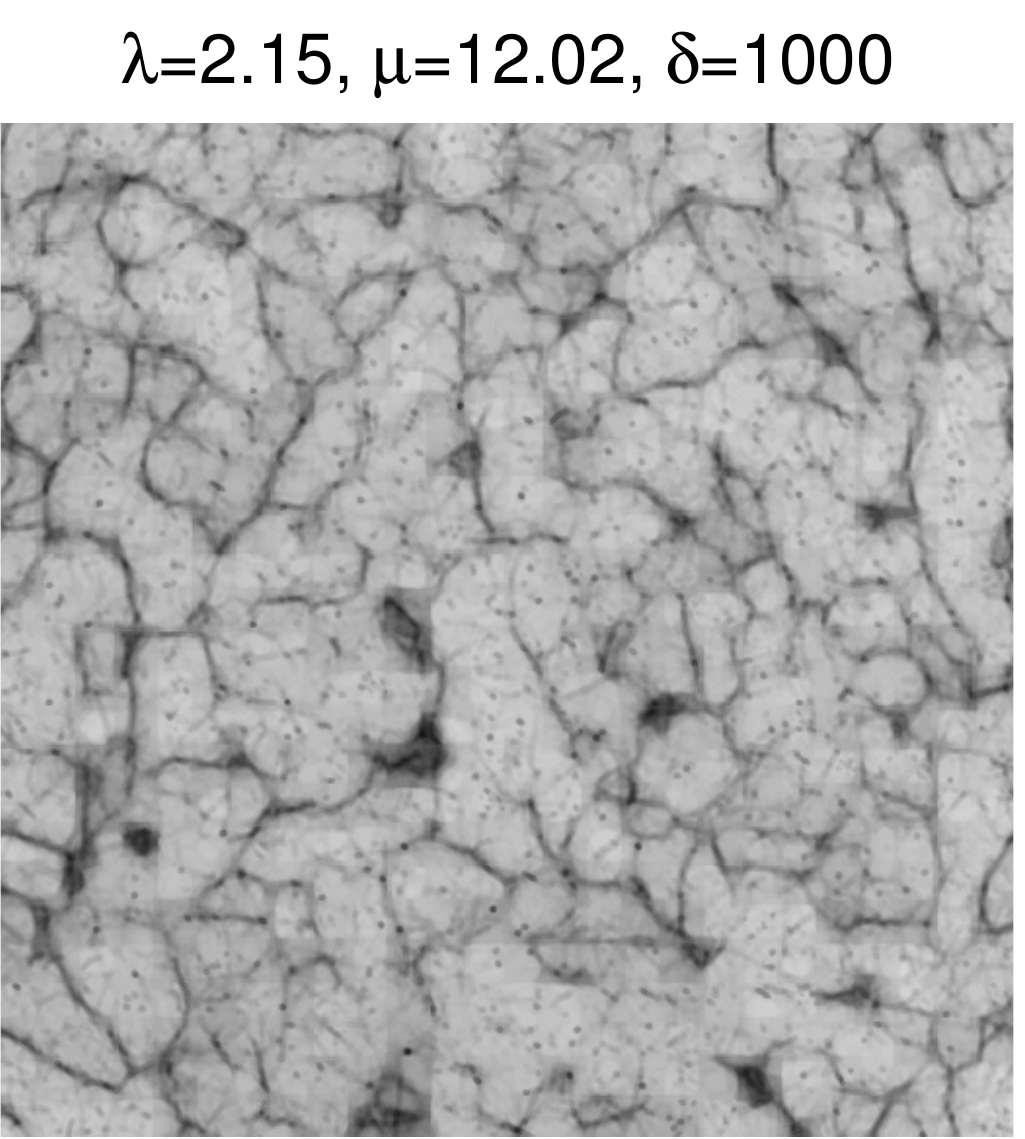}}
\subfigure[TV, RE = 0.099]{ \includegraphics[width=0.3\linewidth]{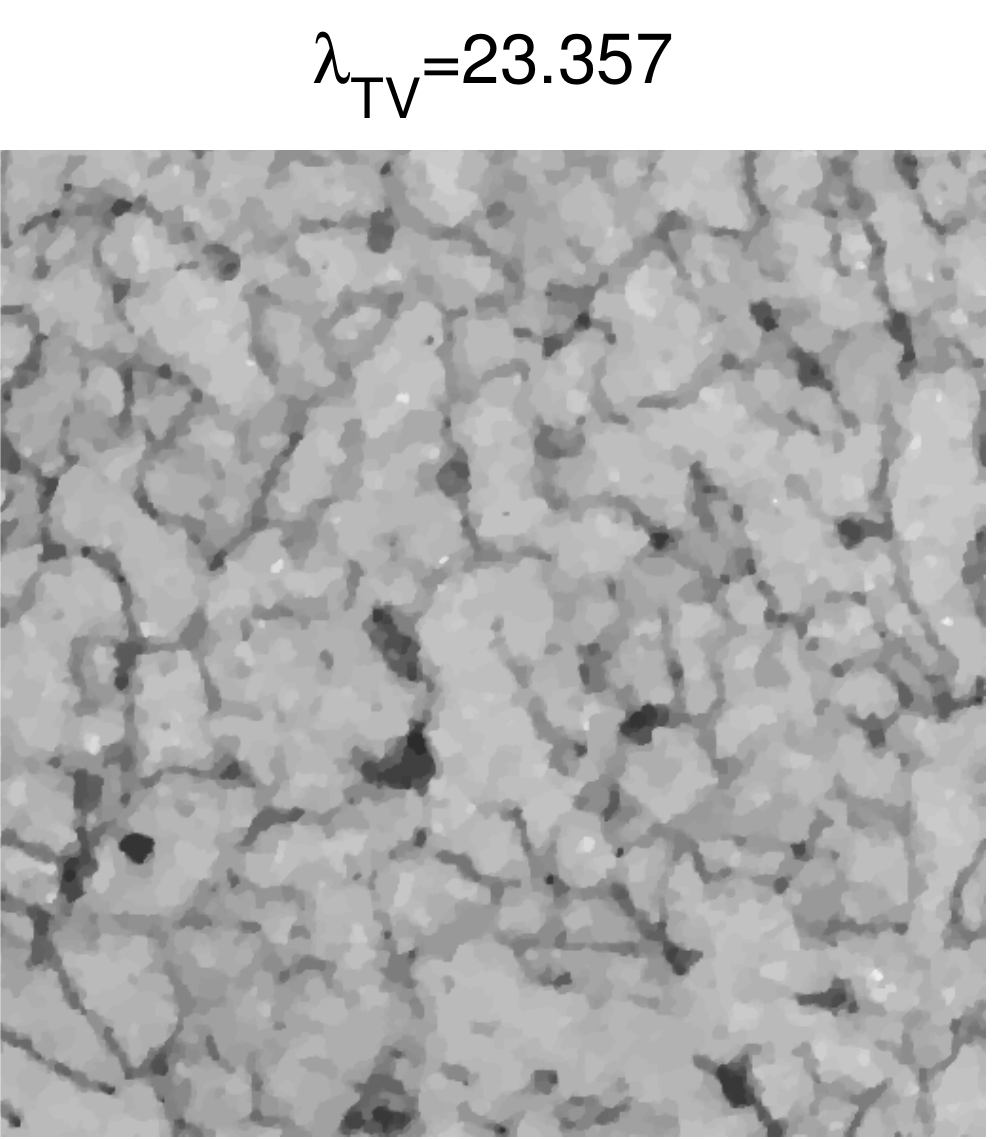}} \\
\subfigure[$D^{20} \in \mathcal{D}_2$, RE = 0.146]{ \includegraphics[width=0.3\linewidth]{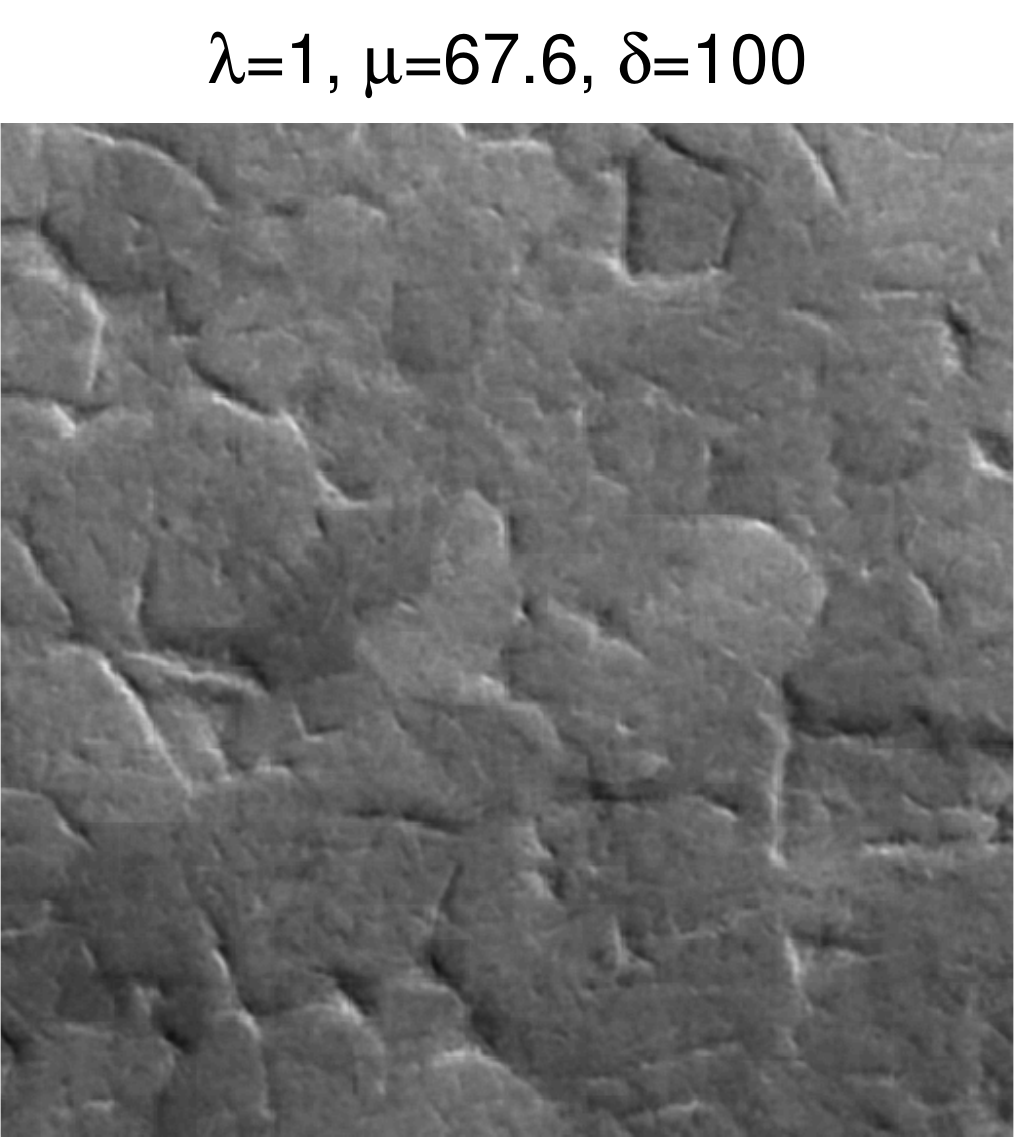}}
\subfigure[$D^{20} \in \mathcal{D}_{\infty}$, RE = 0.158]{ \includegraphics[width=0.3\linewidth]{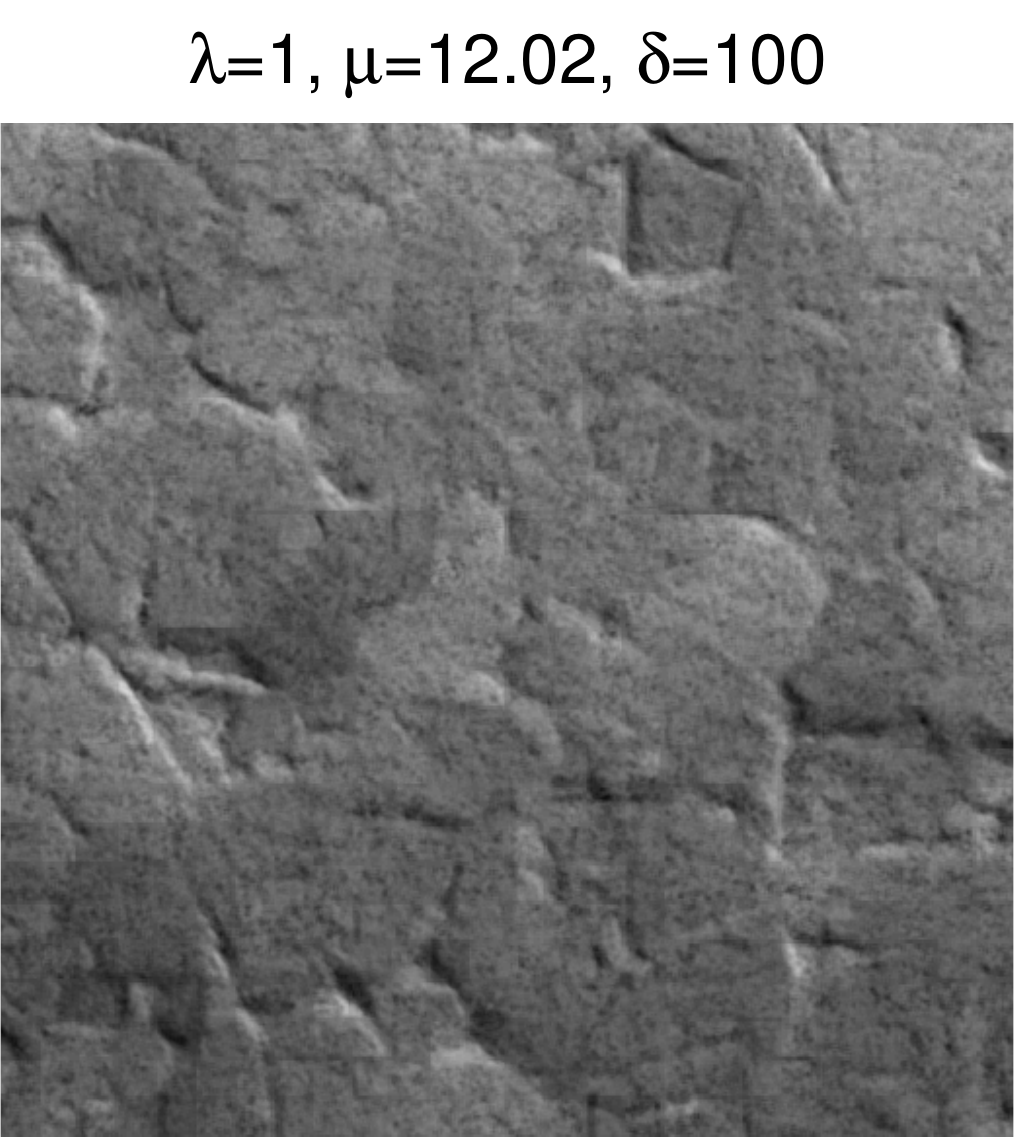}}
\subfigure[TV, RE = 0.137]{ \includegraphics[width=0.3\linewidth]{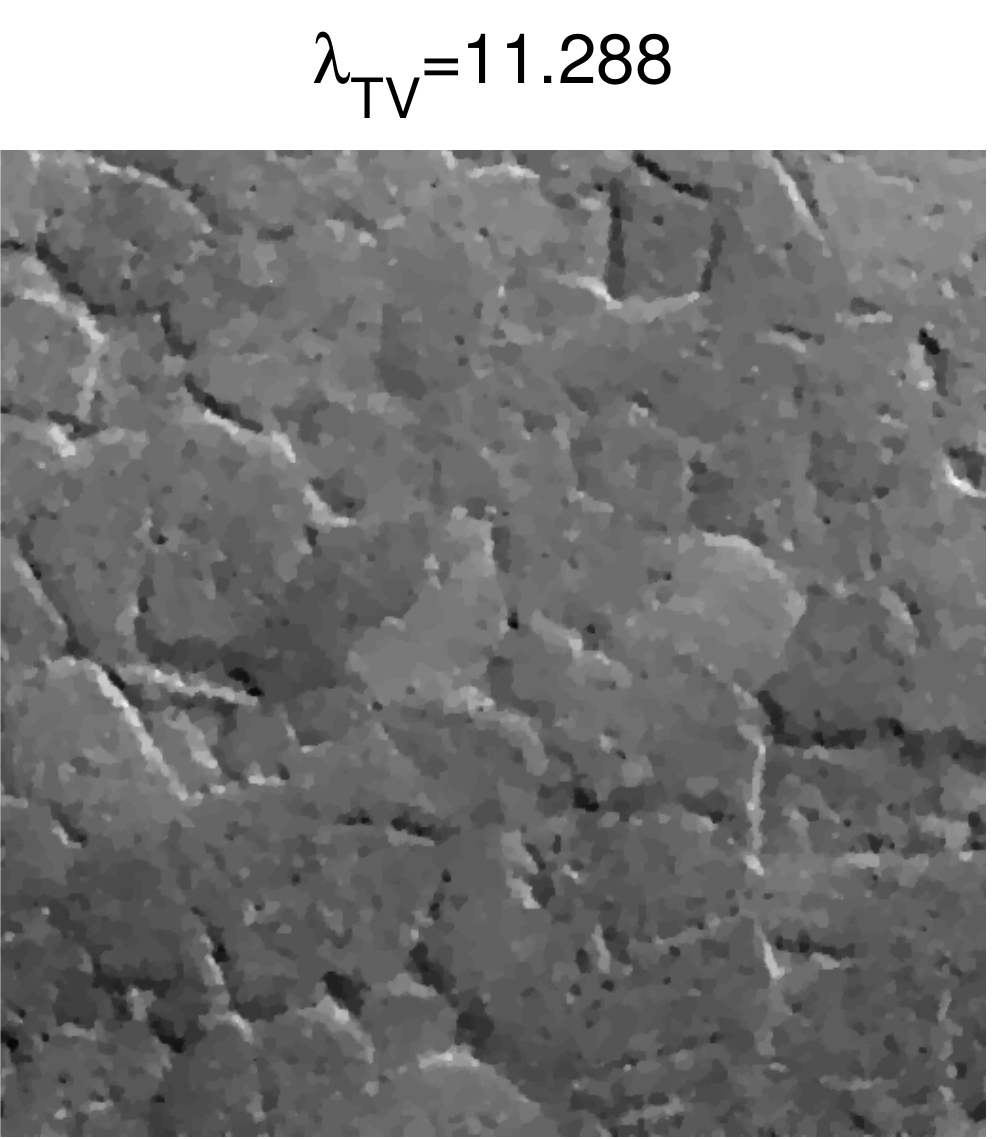}} \\
\caption{Reconstructions of the $520 \times 520$ images by our method (left and middle)
and by the TV method (right).  Top:\ steel microstructure.  Bottom:\ zirconium grains.}
 \label{fig:RecLarger}
\end{figure}

We consider a parallel-beam tomographic scenario with $N_{\mathrm{p}} = 50$ projections
corresponding to $50$ uniformly distributed projections in $[0^\circ,180^\circ]$,
leading to $m=36,750$ measurements.
We add Gaussian white noise with relative noise level 0.01 and compute
reconstructions by our method as well as the TV method;
these reconstruction are shown in Figure \ref{fig:RecLarger}.
All regularization parameters were chosen to give the best reconstruction
as measured by the~RE, and we note that the reconstruction errors are
dominated by the error coming from the regularization of the noisy data;
the approximation errors
$\| P_{\mathcal{C}} (x\ex) - x\ex \|_2/\| x\ex \|_2$ are of the order
0.03 and 0.05 for the steel and zirconium images, respectively.

As expected, the TV reconstructions exhibit ``cartoonish'' artifacts, and for
the steel grains the black interfaces tend to be too thick and they are
not so well resolved.
Our method, for both $\mathcal{D}_2$ and $\mathcal{D}_{\infty}$,
recovers better the grain interfaces that are of interest here.
We obtain the sharpest interfaces for $\mathcal{D}_{\infty}$
but some small black ``dots'' have appeared which are not
present for $\mathcal{D}_2$; in both cases the images are suited for
postprocessing via image analysis.

\section{Conclusions}
\label{sec:Final}
We describe and examine an algorithm that incorporates training images as priors
in computed tomography (CT) reconstruction problems.
This type of priors can be useful in low-dose CT where we are faced with
underdetermined systems of equations, and our numerical experiments focus
on such problems.

Our algorithm has two stages.
In the first stage we compute a learned dictionary from a set of training images
using a regularized nonnegative matrix factorization (NMF)\@.
In the second stage, via a regularized least squares fit we compute a
nonnegative reconstruction lying in the cone defined by the dictionary elements;
the reconstruction is sparse with respect to the dictionary.
Hence, regularization is obtained by enforcing that the reconstruction is
within the range of the dictionary elements and by the sparsity constraint.

Our algorithm works with non-overlapping image patches; the same dictionary is used
for all patches, and we are able to minimize blocking artifacts by an
additional regularization term.
This reduces the computational complexity, compared to all other proposed
algorithms that apply a dictionary-based regularization based on
overlapping patches around every pixel in the image.

Our algorithm includes several regularization parameters.
In the first stage a parameter is used to control the sparsity in the NMF,
and in the second stage we use one parameter to control the sparsity of the
representation in the dictionary, and another parameter to avoid blocking artifacts.
We perform a series of numerical experiments with noisy data and without
committing inverse crime, where we demonstrate the interplay between
these parameters and the computed reconstructions, and we show that
the reconstructions are not very sensitive to these parameters.
Further work is needed to develop automatic parameter choice algorithms.

We conclude that training images can be useful as a strong prior
for regularization of low-dose CT problems, through a sparse representation in
a nonnegative dictionary learned from the training images.
Our reconstructions are (not surprisingly) superior to those computed by
classical methods such as filtered back projection and algebraic iterative
methods, and they are competitive with total variation (TV) reconstructions.
Specifically, in our test problems our algorithm tends to be able to include more texture
and also produces edges whose location is more correct.

\section*{Acknowledgments}
The authors would like to thank Prof.\ Samuli Siltanen from Univ.\ of Helsinki
for providing the high-resolution image of the peppers,
and Dr.\ Hamidreza Abdolvand from Univ.\ of Oxford for providing
the zirconium image.

\appendix
\section{The Dictionary Learning Algorithm}
\label{app:Algorithm}
Recall that the dictionary learning problem \eqref{e-dictionary-learn}
is non-convex, and hence it is too costly to solve it globally. We
will therefore optimize locally by applying the Alternating Direction Method
of Multipliers (ADMM) method \cite{Boyd}to the following reformulation
of \eqref{e-dictionary-learn}
\begin{align} \label{eq:ALNNSC}
  \begin{array}{ll}
    \mbox{minimize}_{D,H}
    &  \frac{1}{2}\, \|Y-UV\|_{\mathrm{F}}^2 + \lambda \, \| H \|_{\mathrm{sum}} +
    I_{\mathbb{R}_+^{s \times t}}(H) +
    I_{\mathcal D}(D) \\
    \mbox{subject to}
    & D = U,\ H = V,
  \end{array}
\end{align}
where $U \in {\mathbb{R}}^{p \times s}$ and $V \in \mathbb{R}^{s \times
  t}$ are auxiliary variables that are introduced in order to make the
ADMM-updates separable and hence cheap. The augmented Lagrangian
associated with (\ref{eq:ALNNSC}) can be expressed as
\begin{align}
\label{eq:Lag}
\begin{split}
 L_{\rho}(D,H,U,V,\varLambda,\varPi) &= \frac{1}{2}\| Y- UV \|_{\mathrm{F}}^2+ \lambda
 \, \| H \|_{\mathrm{sum}}+ I_{\mathbb{R}_+^{s \times t}}(H)  +  I_{\mathcal D}(D) \\
 & \qquad + \mathrm{Tr}(\varLambda^T(D-U))+\mathrm{Tr}(\varPi^T(H-V)) \\
 & \qquad+ \frac{\rho}{2}\| D-U \|_{\mathrm{F}}^2 + \frac{\rho}{2}\| H-V \|_{\mathrm{F}}^2
\end{split}
\end{align}
where $\varLambda \in {\mathbb{R}}^{p \times s}$ and $\varPi \in
{\mathbb{R}}^{s \times l}$ are Lagrange multipliers, and
$\rho$ is a fixed positive penalty parameter which
can be chosen prior to the learning process. If we partition the variables
into two blocks $(D,V)$ and $(H,U)$ and apply ADMM to
\eqref{eq:ALNNSC}, we obtain an algorithm where each iteration
involves the following three steps: (i) minimize  $L_\rho$
jointly over $D$ and $V$; (ii) minimize  $L_\rho$ jointly over $H$ and
$U$; and (iii) update the dual variables $\varLambda$ and $\varPi$ by taking a
gradient-ascent step. Since $L_\rho$ is separable in $D$ and $V$, step
(i) can be expressed as two separate updates
\begin{subequations}\label{e-admm-updates}
 \begin{align}
  D_{k+1} &=  \min_{D\in \mathcal D} L_\rho
  (D,H_k,U_k,V_k,{\varLambda}_k,{\varPi}_k) =
  P_{\mathcal  D}( U_k - \rho^{-1}\varLambda_k) \\
  V_{k+1} &=  \min_{V}
  L_\rho(D_{k},H_{k},U_{k},V,{\varLambda}_k,{\varPi}_k) \\
 \notag &=(U_k^TU_k+\rho I)^{-1}(U_k^TY+\varPi_k + \rho H_k)
\end{align}
where $P_{\mathcal  D}( \cdot)$ is the projection onto the
set $\mathcal D$. Similarly, $L_\rho$ is also separable in $H$ and $U$, so step (ii) can be
written as \begin{align}
  H_{k+1} &= \min_{H \in \mathbb{R}_+^{s \times t}} L_\rho
  (D_{k+1},H,U_k,V_{k+1},{\varLambda}_k,{\varPi}_k) \\
 \notag & = P_{\mathbb{R}_+^{s \times t}} (\mathcal S_{\lambda/\rho} (V_{k+1}-\rho^{-1} \varPi_k ) ) \\
 U_{k+1} &= \min_{U} L_\rho
 (D_{k+1},H_k,U,V_{k+1},{\varLambda}_k,{\varPi}_k) \\
 \notag &= (YV_{k+1}^T+\varLambda_k + \rho D_{k+1}) (V_{k+1}V_{k+1}^T + \rho I)^{-1}
\end{align}
where ${\mathcal{S}}_{\lambda/\rho}$ denotes an entrywise
soft-thresholding operator, and $P_{\mathbb{R}_+^{s \times t}}(\cdot)$ is the projection onto the non-negative orthant. 
Finally, the dual variable updates in step (iii) are given by
\begin{align}
  \varLambda_{k+1}  &= \varLambda_{n}+ \rho (D_{k+1}-U_{k+1}) \\
  \varPi_{k+1} &= \varPi_{k}+ \rho (H_{k+1}-V_{k+1}).
\end{align}
\end{subequations}

The projection onto the set $\mathcal D_{\infty}$ is an element-wise
projection onto the interval $[0,1]$ and hence easy to
compute. However, the projection onto $\mathcal D_2$ does not have a
closed form solution, so we compute it iteratively using
Dykstra's alternating projection algorithm.

The convergence properties of ADMM when applied to non-convex problems
of the form \eqref{eq:ALNNSC} have been studied by e.g.\
\cite{XuY}. They show that whenever the sequence of iterates produced
by \eqref{e-admm-updates} converges, the limit
satisfies the the KKT-conditions (i.e., the first-order
necessary conditions for optimality) which can be expressed as
   \[
     D = U, \quad H = V,
   \]
   \[
     \varLambda = -(Y-DH)H^T, \quad \varPi = -D^T(Y-DH),
   \]
   \[
      - \varLambda \in \partial \Phi_{\mathrm{dic}}(D), \quad
      -\varPi  \in \partial \Phi_{\mathrm{rep}}(H),
  \]
where $\partial$ denotes the subdifferential operator.
The convergence result is somewhat weak, but empirical evidence suggests that
applying ADMM to non-convex problems often works well in practice
\cite{Boyd}. It is interesting to note that the point $D=U =0$ and
$H=V=0$ satisfies the KKT-conditions, and although it is a stationary
point, it is clearly not a local minima. For this reason, we
avoid initializing with zeros.
We initialize $U$ with some of the images from the training set,
and we set $V=[ I \ 0]$ (i.e., the leading $s$ columns of $V$ is the
identity matrix).

The KKT-conditions can be used to formulate stopping criteria.
We use the following conditions
\begin{subequations}
 \begin{align}
\frac{\|D-U\|_{\max}}{\max(1,\|D\|_{\max})} \leq \epsilon  & \quad \wedge \quad
\frac{\|H-V\|_{\max}}{\max(1,\|H\|_{\max})}
\leq  \epsilon \\
\frac{\|\varPi-D^T(DH-Y)\|_{\max}}{\max(1,\|\varPi\|_{\max})} \leq \epsilon & \quad \wedge \quad
\frac{\|\varLambda-(DH-Y)H^T\|_{\infty}}{\max(1,\|\varLambda\|_{\max})}
\leq  \epsilon 
\end{align}
\end{subequations}
where $\epsilon >0 $ is a given tolerance.

The KKT-conditions can also be used to derive an upper bound $\bar{\lambda}$
for the regularization parameter $\lambda$.
It follows from the optimality conditions that for $H=0_{s \times t}$,
${\varPi}=-D^\mathrm{T}Y$  and hence
for some $\bar{\lambda}$ and all $D \in \mathcal{D}$ we have
\[ D^{\mathrm{T}}Y \in \bar{\lambda} \,\partial \|0_{s \times t}\|_{\mathrm{sum}}, \]
i.e., $H =0$ satisfies the KKT-conditions for all $\lambda \geq \bar{\lambda} $.
Thus, if $Y$ is scaled such that all entries in $Y$ are between $0$ and $1$, then the upper
bound $\bar{\lambda} = p$ can be used for both dictionaries since
  \[
    \sup_{D \in {\mathcal{D}}_2} \|D^{\mathrm{T}}Y\|_{\max} = \max_{j=1,\ldots,t} \sqrt{p}
    \|Ye_j\|_2 \leq p  \]
    and
    \[
    \sup_{D \in {\mathcal{D}}_{\infty}} \|D^\mathrm{T}Y\|_{\max} = \max_{j=1,\ldots,t}
    \|Ye_j\|_1 \leq p
  \]
which implies that
$D^{\mathrm{T}}Y \in \bar{\lambda} \,\partial \Phi_{\mathrm{rep}}(0_{s \times t})$ for all
$D \in \mathcal{D}$.

\end{document}